\documentclass[acmtog]{acmart}

\AtBeginDocument{%
  \providecommand\BibTeX{{%
    \normalfont B\kern-0.5em{\scshape i\kern-0.25em b}\kern-0.8em\TeX}}}

\usepackage{kotex}
\usepackage{multirow}
\usepackage{subcaption} 

\begin{document}


\setcopyright{acmlicensed}
\acmJournal{TOG}
\acmYear{2021} \acmVolume{99} \acmNumber{99} \acmArticle{99} \acmMonth{99} \acmPrice{15.00}\acmDOI{nn.nnnn/nnnnnnn}

\newcommand{\Eq}[1]  {Eq.\ (\ref{eq:#1})}
\newcommand{\Eqs}[1] {Eqs.\ (\ref{eq:#1})}
\newcommand{\Fig}[1] {Fig.\ \ref{fig:#1}}
\newcommand{\Figs}[1]{Figs.\ \ref{fig:#1}}
\newcommand{\Tbl}[1]  {Table \ref{tbl:#1}}
\newcommand{\Tbls}[1] {Tables \ \ref{tbl:#1}}
\newcommand{\Sec}[1] {Sec.\ \ref{sec:#1}}
\newcommand{\SSec}[1] {Sec.\ \ref{ssec:#1}}
\newcommand{\Secs}[1] {Secs.\ \ref{sec:#1}}
\newcommand{\Etal}   {{\textit{et al.}}}

\newcommand{\setone}[1] {\left\{ #1 \right\}} 
\newcommand{\settwo}[2] {\left\{ #1 \mid #2 \right\}} 

\newcommand{\todo}[1]{{\textcolor{red}{TODO: #1}}}
\newcommand{\son}[1]{{\textcolor{magenta}{hyeongseok: #1}}}
\newcommand{\jy}[1]{{\textcolor{blue}{Junyong: #1}}}
\newcommand{\sean}[1]{{\textcolor{green}{sean: #1}}}
\newcommand{\sunghyun}[1]{{\textcolor[rgb]{0.6,0.0,0.6}{sunghyun: #1}}}
\newcommand{\change}[1]{{\color{red}#1}}
\newcommand{\correct}[1]{{\color{black}#1}}

\renewcommand{\topfraction}{0.95}
\setcounter{bottomnumber}{1}
\renewcommand{\bottomfraction}{0.95}
\setcounter{totalnumber}{3}
\renewcommand{\textfraction}{0.05}
\renewcommand{\floatpagefraction}{0.95}
\setcounter{dbltopnumber}{2}
\renewcommand{\dbltopfraction}{0.95}
\renewcommand{\dblfloatpagefraction}{0.95}



\renewcommand{\textrightarrow}{$\rightarrow$}

\title{Recurrent Video Deblurring with Blur-Invariant Motion Estimation and Pixel Volumes}



\author{Hyeongseok Son}
\email{sonhs@postech.ac.kr}
\affiliation{%
  \institution{POSTECH}
}

\author{Junyong Lee}
\email{junyonglee@postech.ac.kr}
\affiliation{%
\institution{POSTECH}
}

\author{Jonghyeop Lee}
\email{ljh5644@postech.ac.kr}
\affiliation{%
\institution{POSTECH}
}

\author{Sunghyun Cho}
\email{s.cho@postech.ac.kr}
\affiliation{%
\institution{POSTECH}
}
\author{Seungyong Lee}
\email{leesy@postech.ac.kr}
\affiliation{%
\institution{POSTECH}
}

\renewcommand{\shortauthors}{Son et al.}

\begin{abstract}
For the success of video deblurring, it is essential to utilize information from neighboring frames. Most state-of-the-art video deblurring methods adopt motion compensation between video frames to aggregate information from multiple frames that can help deblur a target frame. However, the motion compensation methods adopted by previous deblurring methods are not blur-invariant, and consequently, their accuracy is limited for blurry frames with different blur amounts. To alleviate this problem, we propose two novel approaches to deblur videos by effectively aggregating information from multiple video frames. First, we present {\em blur-invariant motion estimation learning} to improve motion estimation accuracy between blurry frames. Second, for motion compensation, instead of aligning frames by warping with estimated motions, we use a {\em pixel volume} that contains candidate sharp pixels to resolve motion estimation errors. We combine these two processes to propose an effective recurrent video deblurring network that fully exploits deblurred previous frames. Experiments show that our method achieves the state-of-the-art performance both quantitatively and qualitatively compared to recent methods that use deep learning.
\end{abstract}



\begin{teaserfigure}
  \includegraphics[width=\textwidth]{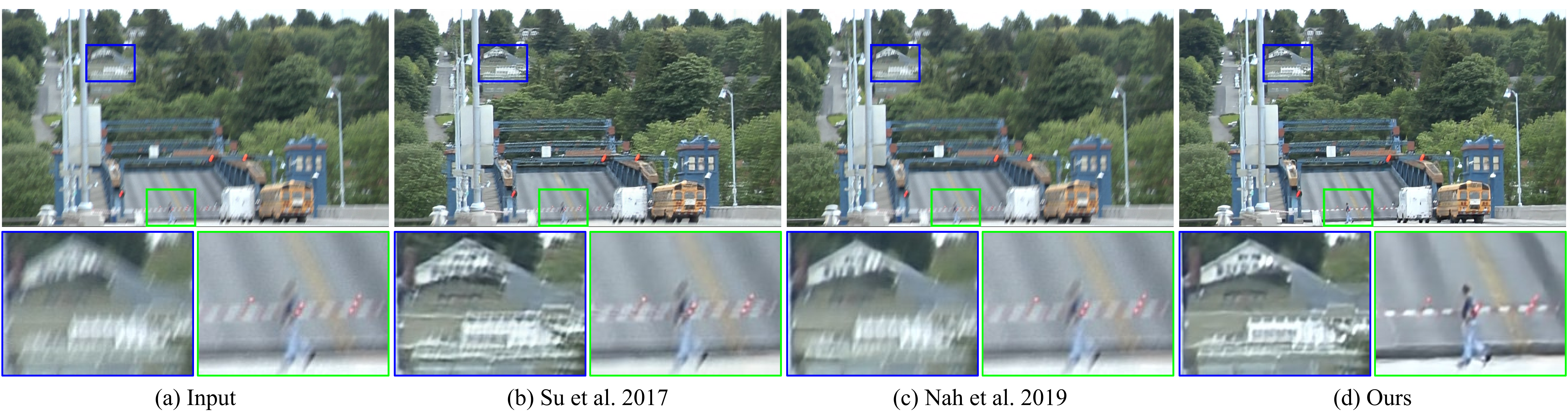}
  \vspace{-12pt}
  \caption{Deblurring results of a real-world video frame with large blur.
  Our method successfully restores the challenging video frame compared to other state-of-the-art methods.
  The input video frame is from Su \Etal~\shortcite{su2017deep}'s test set.}
  \label{fig:teaser}
\end{teaserfigure}

\maketitle

\section{Introduction}
Videos captured by hand-held devices are vulnerable to motion blur due to camera shakes and object motion.
Video deblurring increases the visual quality of a video, and can also improve the accuracy of other video processing tasks, such as object recognition~\cite{kupyn2018}, tracking~\cite{seibold2011}, and 3D reconstruction~\cite{lee2013dense}.
Numerous video deblurring methods have been proposed, including deconvolution-based~\cite{kim2015general,ren2017video}, aggregation-based~\cite{cho2012video,delbracio2015}, and deep learning-based approaches~\cite{su2017deep,kim2017online,kim2018spatio,Nah2019recurrent,Wang2019edvr}.

Video deblurring methods typically rely on information about motion between neighboring frames,
because it can be used for roughly estimating the blur
and aligning frames to provide different captures of the same scene.
Deconvolution-based methods~\cite{kim2015general,ren2017video} exploit motion between neighboring frames to estimate spatially-varying blur kernels, which are then used to deconvolve blurry frames and restore sharp frames.
Aggregation-based methods~\cite{cho2012video,delbracio2015} use motion information to directly aggregate sharp pixels from neighboring frames.
Recent deep learning-based methods~\cite{su2017deep,kim2017online,kim2018spatio,Nah2019recurrent,Wang2019edvr} train deep convolutional neural networks (CNNs)
to directly produce deblurred frames from multiple input frames.
While some deep learning-based methods do not explicitly assume aligned input frames, Su~\Etal~\shortcite{su2017deep} show that rough alignment
can increase deblurring quality.

However, accurate estimation of motion is challenging in the presence of blur, and
incorrectly estimated motion may cause structural deformations
during the motion compensation process, and thereby eventually degrade the deblurring quality.
To resolve this problem, we introduce two novel approaches: {\em blur-invariant motion estimation learning} and {\em pixel volume-based motion compensation}.
First, to estimate motion accurately, we adopt \emph{LiteFlowNet}~\cite{LiteFlowNet} and train it with a blur-invariant loss so that the trained network can estimate blur-invariant optical flow between frames.
We refer the resulting network as a blur-invariant motion estimation network (\emph{BIMNet}).
Second, for effective motion compensation, instead of aligning a frame by warping, we construct a pixel volume that consists of multiple matching candidates for each pixel.
Compared to a warped frame, our pixel volume provides additional information for a deblurring network to robustly restore a sharp frame, even under slight motion estimation errors.

By leveraging pixel volume-based motion compensation with blur-invariant motion estimation,
we propose an effective video deblurring framework that is based on a recurrent CNN structure. 
To produce a deblurred result of the current frame, our framework takes four video frames as input: the previous, current, and next blurry input frames, and the deblurred result of the previous frame.
In this way, we can exploit restored information of the previous frame as well as other information in neighboring input frames.
Our framework then estimates motion between the current and previous frames in a blur-invariant way by using \emph{BIMNet}, then uses the estimated motion to construct a pixel volume from the deblurred previous frame.
Finally, our deblurring network restores a sharp image for the current frame using the pixel volume with input blurry frames.

We refer to our deblurring network as a pixel volume-based deblurring network (\emph{PVDNet}).
Due to \emph{BIMNet} and pixel volume, \emph{PVDNet} can produce visually pleasing deblurring results (\Fig{teaser}).
Experimental results show that our framework achieves the state-of-the-art performance both quantitatively and qualitatively.

Our main contributions are summarized as follows.
\begin{itemize}
\item{We present blur-invariant learning of a motion estimation network for blurry video frames, which is essential for accurate alignment of neighboring frames.}
\item{We propose a novel pixel volume of matching candidates for motion compensation, which provides additional information for robust reconstruction of sharp frames.}
\item{We propose an effective recurrent CNN framework based on \emph{BIMNet} and pixel volume for video deblurring.}
\end{itemize}

\section{Related Work}

\subsection{Video deblurring}

\paragraph{Deconvolution-based approaches}
Typical single image deblurring approaches~\cite{cho2009fast,xu2010two,hirsch2011fast,whyte12non,kim2013dynamic,pan2016soft,xu2012depth} estimate blur kernels and restore latent sharp images by applying deconvolution with the estimated kernels.
However, estimating spatially-varying blur, which is common in the real world, is a severely ill-posed problem.
Therefore, many approaches assume specific blur models and focus on limited cases such as camera shakes~\cite{hirsch2011fast,whyte12non},
moving objects~\cite{kim2013dynamic,pan2016soft} and depth variation~\cite{xu2012depth}.

Video deblurring based on deconvolution also requires estimation of spatially-varying blur kernels.
Early approaches~\cite{bar2007,wulff2014} used segmentation maps to reduce the ill-posedness and to effectively process blur caused by moving objects.
Later methods \cite{kim2015general,ren2017video} proposed video deblurring frameworks based on optical flow to model inter-frame motion.
These methods alternatingly perform optical flow estimation and image deblurring.
However, they still assume relatively simple blur models, so can fail when applied to real-world videos that include complex blur.

\paragraph{Multi-frame aggregation-based approaches}
A few methods~\cite{matsushita2006,cho2012video,delbracio2015} assume that input frames are differently blurred, and have different partial information about latent sharp frames.
Then, by aggregating such partial information on the current frame, deblurring can be done without deconvolution.
These methods are usually performed in combination with frame alignment.
Matsushita \Etal~\shortcite{matsushita2006} align video frames by using a homography, then blend them to remove blur.
Cho \Etal~\shortcite{cho2012video} introduced patch-based local search to increase the accuracy of pixel-wise correspondence between video frames.
The patch-based local search yields visually pleasing results by directly blending sharp patches from neighboring frames.
To achieve video deblurring, Delbracio and Sapiro~\shortcite{delbracio2015} combined a multi-frame deblurring method for burst shot~\cite{delbracio2015burst} with optical flow-based frame alignment.
As these methods do not use deconvolution, they could be computationally more efficient and produce less artifacts.
However, they cannot restore sharp frames when all input frames are blurry.

\paragraph{Deep learning-based approaches}
Several approaches use neural networks that are trained to automatically aggregate information from neighboring frames and reconstruct deblurred frames.
Su \Etal~\shortcite{su2017deep} proposed a CNN that receives multiple frames concatenated together as input.
They also showed rough motion compensation between input frames helps to deblur difficult scenes with large motions.
Kim \Etal~\shortcite{kim2017online} and Nah \Etal~\shortcite{Nah2019recurrent} presented recurrent CNNs for video deblurring, and showed that utilizing previous results can improve deblurring quality.
However, these methods do not work well for videos with large motion, as they do not use motion compensation.
Recently, a spatio-temporal transformer network~\cite{kim2018spatio} was proposed to improve motion compensation performance for video restoration tasks, such as video super-resolution and video deblurring.
The network estimates optical flow from multiple neighboring frames together to effectively handle occlusions.
More recently, Wang \Etal~\shortcite{Wang2019edvr} proposed an alignment module based on deformable convolutions for video restoration tasks that allows more effective utilization of information from other video frames.

\subsection{Blur-invariant optical flow}
There have been a small number of approaches to computing optical flow between frames with spatially-varying blurs, such as Portz \Etal~\shortcite{portz2012optical} and Daraei \shortcite{Daraei2014OpticalFC}.
To take account of blur in the searching process, both methods blur each input frame with the blur kernel of the other frame.
While they can increase the accuracy of the estimation of optical flow, they are computationally heavy because of iterative update of optical flow and blur kernels.
Furthermore, they use simple parametric blur models that combine two linear motions by considering forward and backward optical flows, and therefore do not generalize well to real-world blurry videos.

\begin{figure}[t]
\centering
	\begin{tabular}{c}
		\includegraphics[width=0.95\columnwidth]{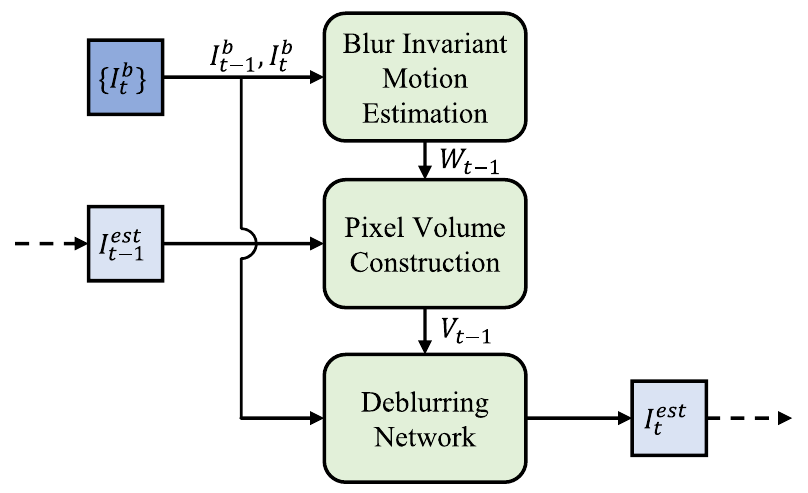}
	\end{tabular}
\vspace{-0.2cm}	
	\caption{Our video deblurring framework.}
\label{fig:framework}
\vspace{-0.2cm}
\end{figure}

\section{Our Video Deblurring Framework}
\label{sec:overall}

Our video deblurring framework consists of three modules: a blur-invariant motion estimation network (\emph{BIMNet}), a pixel volume generator, and a pixel volume-based deblurring network (\emph{PVDNet}) (see \Fig{framework}).
We first train \emph{BIMNet}; after it has converged, we combine the two networks with the pixel volume generator.
We then fix the parameters of \emph{BIMNet} and train \emph{PVDNet} by training the entire network.

The framework takes three consecutive input blurry frames $I_{t-1}^b$, $I_{t}^b$, and $I_{t+1}^b$, and the deblurred result of the previous frame $I_{t-1}^{est}$ as input where $I_{t}^b$ is the $t$-th input blurry frame.
Our \emph{BIMNet} first estimates optical flow $W_{t-1}$ from $I_{t}^{b}$ to $I_{t-1}^{b}$ (\SSec{BIMNet}).
The pixel volume generator then takes $W_{t-1}$ and $I_{t-1}^{est}$, and constructs a pixel volume by using pixel values of $I_{t-1}^{est}$ according to $W_{t-1}$ (\SSec{PV}).
Finally, our \emph{PVDNet} takes the pixel volume $V_{t-1}$ and the input blurry frames $\{I_{t-1}^b,I_{t}^b,I_{t+1}^b\}$, and produces a deblurring result $I_{t}^{est}$ for the current frame $I_{t}^{b}$ (\SSec{RDN}).

We feed a pixel volume $V_{t-1}$ obtained from the restored frame $I_{t-1}^{est}$ into \emph{PVDNet}, then the use of $I_{t-1}^b$ as input of \emph{PVDNet} may seem unnecessary.
However, $I_{t-1}^b$ is occasionally sharper than $I_{t-1}^{est}$ especially when camera motion is not significant, and \emph{PVDNet} can take advantage of $I_{t-1}^b$ in such cases.
In addition, we apply motion compensation not to $I_{t-1}^b$ and $I_{t+1}^b$ but only to $I_{t-1}^{est}$.
This choice may seem counterintuitive, but preliminary experiments demonstrated that it maximizes the deblurring performance of \emph{PVDNet} (Appendix \ref{ssec:preliminary_test}).

\subsection{Blur-invariant motion estimation learning}
\label{ssec:BIMNet}

Our \emph{BIMNet} adopts the network architecture of \emph{LiteFlowNet}~\cite{LiteFlowNet}, which takes two images and generates optical flow between them.
\emph{FlowNet}~\cite{FlowNet} and its follow-up works,
\emph{FlowNet2}~\cite{ilg17flow} and \emph{LiteFlowNet}~\cite{LiteFlowNet}, have a Siamese network structure for the encoder, and we found that the structure enables effective blur-invariant feature learning (Appendix \ref{ssec:siamese_structure}).
We adopt \emph{LiteFlowNet} for the network structure of \emph{BIMNet}, because \emph{LiteFlowNet} shows comparable accuracy to \emph{FlowNet2} with far fewer parameters (\SSec{mc_performance}).
Originally, \emph{LiteFlowNet} was trained using blur-free datasets that provide ground truth optical flow maps, such as
Scene Flow~\cite{MIFDB16}, Sintel~\cite{Butler:ECCV:2012:sintel}, KITTI~\cite{Geiger2012CVPR:KITTI}, and Middlebury~\cite{Scharstein:2014:Middlebury}.
However, no available dataset has ground truth optical flow maps for {\em blurry} images.
Thus, we train our \emph{BIMNet} in a {\em self-supervised} way by using a blurred video dataset~\cite{su2017deep} that contains pairs of sharp and blurred videos.

A successfully trained \emph{BIMNet} should be able to produce accurate optical flow from a pair of video frames regardless of the amounts of blur that the frames include.
To train the network to acquire this property, we generate four pairs of frames from each pair of consecutive frames in sharp and blurred videos:
$(I_{t-1}^s,I_{t}^s)$, $(I_{t-1}^b,I_{t}^b)$, $(I_{t-1}^b,I_{t}^s)$, and $(I_{t-1}^s,I_{t}^b)$,
where $I_{t}^s$ and $I_{t}^b$ are the $t$-th frames in sharp and blurred videos, respectively.
For each image pair $(I_{t-1}^\alpha,I_{t}^\beta)$, where $\alpha, \beta \in \{s,b\}$, our \emph{BIMNet} estimates optical flow $W^{\alpha\beta}_t$ from $I_{t}^\beta$ to $I_{t-1}^\alpha$. 

Then, we train \emph{BIMNet} with a {\em blur-invariant} loss $L_{BIM}^{\alpha\beta}$ for every pair ($\alpha$, $\beta$):
\begin{equation}
L_{BIM}^{\alpha\beta}=MSE(Warp(I_{t-1}^s,W^{\alpha\beta}_t), I_t^s),
\label{eq:L_BIM}
\end{equation}
where $MSE$ computes the mean squared error between two images and $Warp(I_{t-1}^s,W^{\alpha\beta}_t)$ is a warping of $I_{t-1}^s$ by $W^{\alpha\beta}_t$.
$Warp$ can be easily implemented using the sampling layer of the spatial transformer network~\cite{NIPS2015_5854}.
The key idea of $L_{BIM}^{\alpha\beta}$ is that it induces \emph{BIMNet} to learn optical flow between sharp frames, $I_{t-1}^s$ and $I_t^s$, as the ground truth regardless of the blur types of input images, $I_{t-1}^\alpha$ and $I_{t}^\beta$.

A straightforward idea for blur-invariant learning would be to estimate optical flow maps between sharp video frames and use them as ground truth labels for training optical flow estimation between blurry frames.
However, in preliminary experiments, this approach did not improve accuracy as anticipated.
One possible reason for the failure is that optical flow estimated from sharp video frames had the low quality.
Specifically, there may be a domain gap between our training dataset and the optical flow datasets that have been used by recent learning-based optical flow methods, such as, \emph{FlowNet}~\cite{FlowNet}, \emph{FlowNet2}~\cite{ilg17flow}, and \emph{LiteFlowNet}~\cite{LiteFlowNet}.
Thus, optical flow maps estimated from sharp frames may have errors that can hinder using them 
as ground-truth labels.
In contrast, our proposed approach can effectively train \emph{BIMNet} without the need for accurate ground-truth labels.

\subsection{Pixel volume for motion compensation }
\label{ssec:PV}

\begin{figure}[t]
\centering
	\begin{tabular}{c}
		\includegraphics[width=0.9\columnwidth]{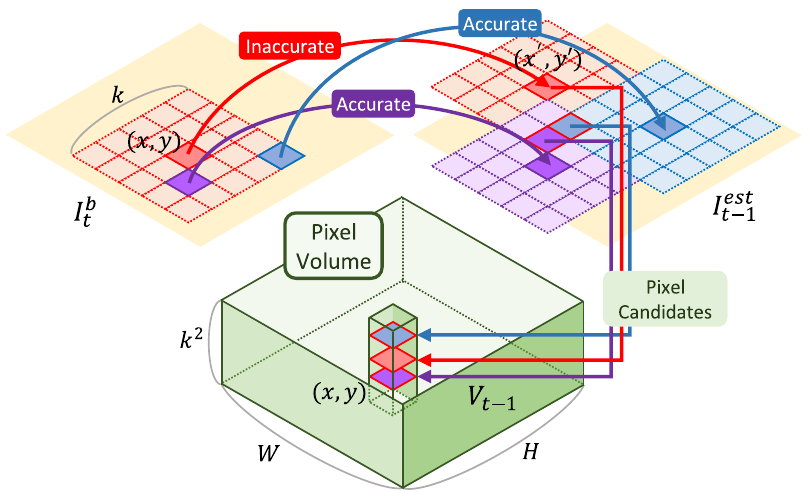}
	\end{tabular}
	\vspace{-0.2cm}
	\caption{Pixel volume construction. A pixel $(x, y)$ in $I_{t}^{b}$ has $k^2$ matching candidates in $I_{t-1}^{est}$, where each pixel in the $k \times k$ window centered at $(x, y)$ gives one candidate.
Pixel volume $V_{t-1}$ is a 3D volume containing the $k^2$ candidates for each pixel of $I_{t}^{b}$. For a pixel $(x, y)$, even when the matching candidate determined by \emph{BIMNet} is inaccurate, the pixel pile of $V_{t-1}$ at $(x, y)$ may contain a correct match.}
	\vspace{-0.2cm}
\label{fig:pv_const}
\end{figure}

Previous video deblurring methods~\cite{su2017deep,kim2018spatio} first align video frames by warping them before deblurring.
However, during the warping process, structural artifacts may occur due to inaccurately estimated motion, which can eventually degrade the final deblurring results.
To relieve this problem, Liao \Etal~\shortcite{Liao2015vsr} proposed the draft-ensemble approach that computes multiple optical flow maps by using different regularization strengths so that correct motion for a local region can be estimated in some of the maps.
However, this approach is computationally inefficient because it requires multiple optical flow estimation.
To tackle the problem effectively and efficiently, we propose a novel pixel volume that does not require multiple optical flow estimation but can still provide multiple candidates for matching pixels between images.

For a pixel $(x,y)$ of $I_t^b$, we consider a spatial window of size $k\times k$ centered at $(x,y)$, where we fix $k=5$.
For each pixel $(x+\Delta_x, y+\Delta_y)$ in the window, where $\Delta_x,\Delta_y\in\{-\lfloor k/2 \rfloor, ..., \lfloor k/2 \rfloor\}$, we have a matched pixel $(x',y')$ in $I_{t-1}^{est}$ determined by optical flow $W_{t-1}$.
Then, its neighboring pixel $(x'-\Delta_x,y'-\Delta_y)$ in $I_{t-1}^{est}$ is a matching candidate for pixel $(x,y)$.
Consequently, we have $k^2$ matching candidates in $I_{t-1}^{est}$ for $(x,y)$.
We collect such matching candidates for all pixels in $I_t^b$ and stack them to obtain a pixel volume $V_{t-1}$ of size $W\times H\times k^2$, where $W$ and $H$ are the width and height of a video frame, respectively.
The resulting pixel volume is in the form of a pile of $k^2$ images, where each image consists of pixels sampled from $I_{t-1}^{est}$, and the image for $(\Delta_x,\Delta_y)=(0,0)$ is the warping of $I_{t-1}^{est}$ by $W_{t-1}$.
\Fig{pv_const} illustrates the construction of a pixel volume.

The occurrence of multiple matching candidates in a pixel volume provide additional information that can reduce the deblurring artifacts caused by inaccurate motion estimation.
Suppose that the pixel matching for $(x, y)$ determined by $W_{t-1}$ is incorrect due to small errors in motion estimation.
Even in that case, if one of the $k^2$ neighbors of $(x, y)$ has a correct match, the pixel pile of $V_{t-1}$ at $(x, y)$ would contain a correct match for $(x,y)$.
Moreover, a pixel volume provides an additional cue for motion compensation based on the majority.
When the neighbors of a pixel have correct matches, the pixel volume can provide multiple duplicates of the correct match in the matching candidates for the pixel. Then, the majority cue can increase the reliability of motion compensation in deblurring the current frame $I_t^b$.
In \SSec{pv_statistic}, we provide detailed statistics of candidate pixels in a pixel volume.

In short, our pixel volume approach leads to the performance improvement of video deblurring by utilizing the multiple candidates in a pixel volume in two aspects: 1) in most cases, the majority cue for the correct match would help as the statistics in \SSec{pv_statistic} shows, and 2) in other cases, \emph{PVDNet} would exploit multiple candidates to estimate the correct match referring to nearby pixels that have majority cues.
This is a clear advantage over warping based alignment, because both majority cue and exploitation of multiple values cannot be provided by a single warped image that has only one candidate for each pixel.

\begin{figure}[t]
\centering
	\begin{tabular}{c}
		\includegraphics[width=\columnwidth]{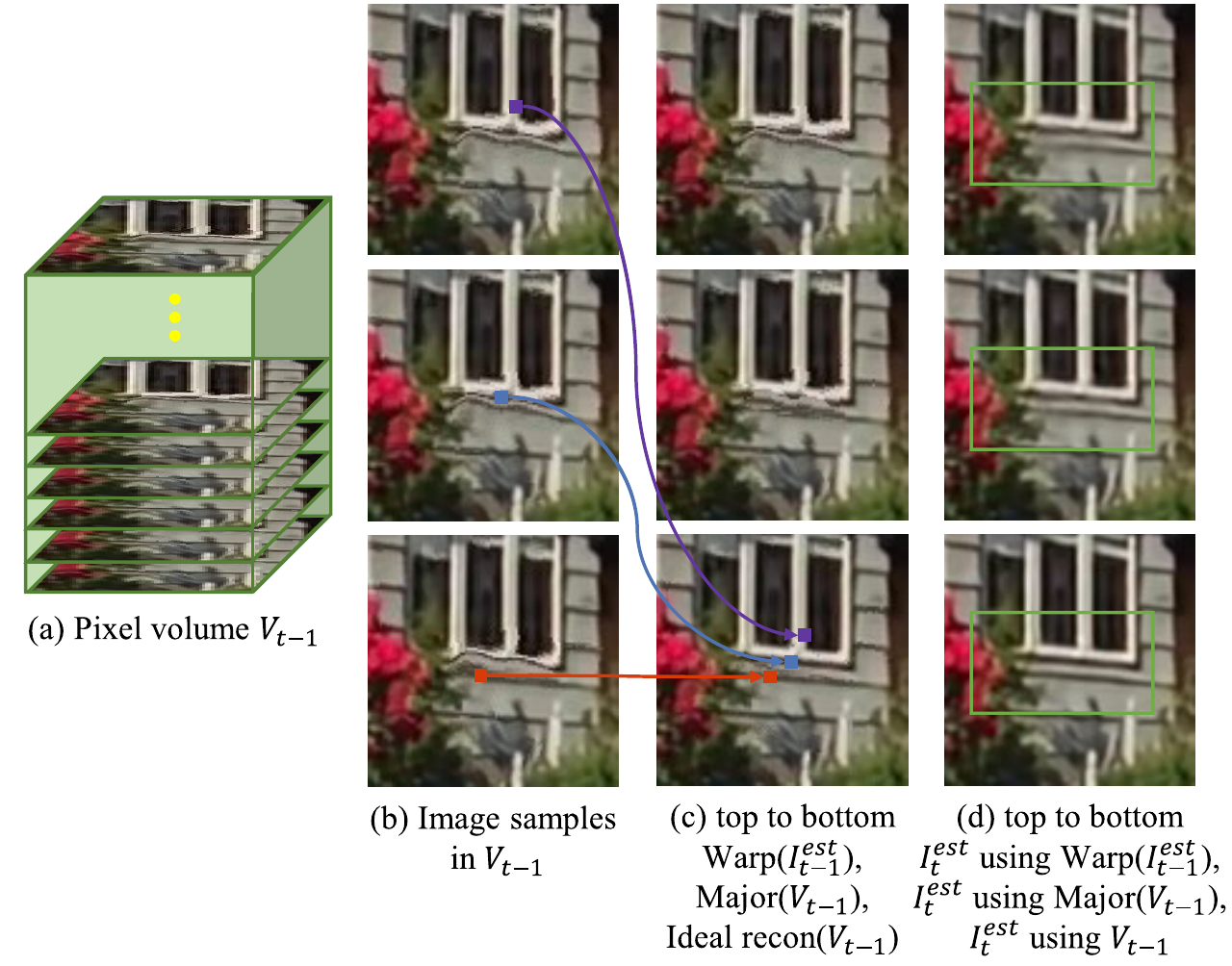}
	\end{tabular}
	\vspace{-0.3cm}
	\caption{Robust deblurring with a pixel volume.
    Due to motion estimation errors, both na\"ive warping and majority-based warping of the deblurred previous frame $I_{t-1}^{est}$ contain artifacts ((c) top and middle). As a result, the variants of \emph{PVDNet} using such warped frames produce deblurred results with remaining blur ((d) top and middle). In contrast, despite the motion estimation errors, our pixel volume contains enough information to reconstruct the ideal warping of $I_{t-1}^{est}$ ((c) bottom). Our \emph{PVDNet} utilizes the information in the pixel volume to produce a high quality deblurred frame ((d) bottom).}
	\vspace{-0.3cm}
\label{fig:pv_work}
\end{figure}

\begin{figure*}[t]
\centering
	\begin{tabular}{c}
		\includegraphics[width=0.98\textwidth]{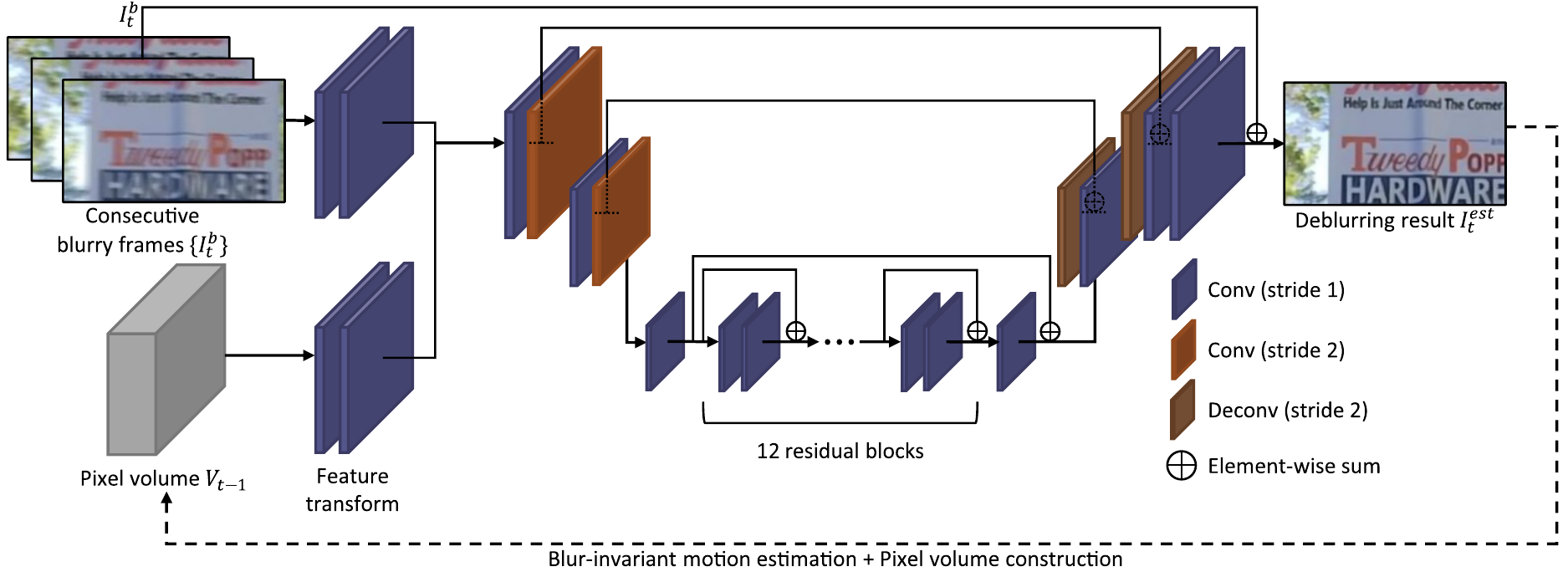}
	\end{tabular}
	\vspace{-0.4cm}
	\caption{Architecture of \emph{PVDNet}.} 
\label{fig:network_architecture}
	\vspace{-0.2cm}
\end{figure*}

\Fig{pv_work} compares motion compensation using warping and a pixel volume.
The warped result of $I_{t-1}^{est}$ may contain deformation artifacts caused by motion estimation errors (\Fig{pv_work}c top).
We also consider the majority-based warped result of $I_{t-1}^{est}$ (\Fig{pv_work}c middle).
In majority-based warping, 
for each pixel, we choose the most frequent pixel value (the mode) among the $k^2$ candidates in a pixel volume.
The majority-based warped frame contains similar artifacts to the simple warping result.
In regions where the motion estimation results are inaccurate, the estimated flows are not consistent among neighboring pixels. 
Then, the most frequent values may not always be the best for motion compensation.
As a result, the deblurring results obtained using such warped frames still contain a small amount of blur (\Fig{pv_work}d top and middle).
In contrast, even in regions where motion estimation results are inaccurate, 
probably one or a few of the candidates in the pixel volume are correct.
To demonstrate this phenomenon, we reconstructed an ideally warped image (\Fig{pv_work}c bottom) by choosing the most similar pixel to the ground truth sharp frame $I_{t}^s$ among the $k^2$ candidates for each pixel.
The ideally warped image contains successfully reconstructed structures with less distortion, which shows the benefit of using a pixel volume. 
Consequently, the deblurring result obtained using the pixel volume (\Fig{pv_work}a) shows successfully restored sharp structures (\Fig{pv_work}d bottom).
In \SSec{pv_analysis}, we provide a quantitative analysis on the effect of a pixel volume.

\paragraph{Differences from relevant approaches}
Our pixel volume construction scheme shares the same basis with the propagation step of the PatchMatch algorithm~\cite{Barnes2009}: both methods exploit the spatial coherency of neighbor matches.
However, our pixel volume scheme is distinct from PatchMatch in the following points.
The PatchMatch propagation is designed to accelerate correspondence search, and therefore does not construct an explicit volume.
Moreover, it is not straightforward to adopt PatchMatch to deep learning models.
In contrast, our pixel volume scheme is specifically designed to improve the robustness of motion compensation and easily integrated into deep learning networks. 

There have been other approaches relevant to our pixel volume scheme. Cho \Etal~\shortcite{cho2012video} also considered multiple candidates for video deblurring by using a window-based patch search. However, their method does not exploit the spatial coherency of neighbor matches, and is not designed for neural networks. Conventional stereo matching algorithms also construct a cost volume, but such cost volumes contain not pixel values but matching costs, and are therefore not directly applicable to video deblurring. Cost volumes do not exploit the spatial coherency of neighbor matches either.

\subsection{Pixel volume based video deblurring network}
\label{ssec:RDN}

\emph{PVDNet} employs the U-Net~\cite{Ronneberger2015} architecture that has shown to be effective for deblurring~\cite{nah2017deep,su2017deep,tao2018scale}.
We use symmetric skip-connections between the encoder and decoder to reconstruct a result image from encoded features while preserving image structures (\Fig{network_architecture}).
We add 12 residual blocks between the encoder and decoder to effectively refine blurry feature maps in low-resolution.
We apply ReLU after all convolution and deconvolution layers except the layers that are connected by skip-connections; in these layers, the ReLU is applied after the summation of skip-connected features.
We refer the readers to the appendix for a detailed architecture of \emph{PVDNet}.

For early fusion of input images and a pixel volume, we add initial feature transform layers to the network.
As aforementioned, our \emph{PVDNet} takes four inputs (the previous, current, and next input frames, and pixel volume of the deblurred previous frame).
Instead of directly concatenating the inputs, we transform and concatenate them using feature transform layers.
We divide the inputs into two groups: the concatenated consecutive frames, and the pixel volume.
We feed each group into the feature transform layers and fuse their features by concatenation.
We use a grayscale pixel volume to reduce training and computational complexity.
Our preliminary test shows that the model using a grayscale pixel volume has comparable performance to the model using a color pixel volume (Appendix \ref{ssec:preliminary_test}).
The network has a long skip-connection from the input frame $I_t^b$ to the network output to make the network predict only residuals while preserving overall contents of the input frame.

To train \emph{PVDNet}, we use mean absolute difference between deblurred frames and their corresponding ground truth sharp frames as the loss function.
Although the network does not have any specific constraints or loss functions for temporal coherence, it generally produces temporally smooth results as our recurrent network utilizes the deblurred previous frame to process the current frame.

\section{Experiments}

\subsection{Implementation and training details}
\label{ssec:training}

To train both \emph{BIMNet} and \emph{PVDNet}, we used a dataset of synthetically blurred videos \cite{su2017deep},
which was generated by capturing sharp videos at high frame rate, and averaging adjacent frames.
The dataset consists of 71 pairs of a blurry video and its corresponding sharp video, which provide 6,708 pairs of 1280$\times$720 blurred and sharp frames in total.
The videos in the dataset have blur caused by camera shake and object motion.
We used 61 videos in the dataset as our training set, and 10 videos as our test set as Su \Etal~\shortcite{su2017deep}~did. The training and test sets consist of 5,708 and 1,000 pairs of images, respectively.

Our training process consists of two steps. We first train \emph{BIMNet}, and then \emph{PVDNet} while fixing \emph{BIMNet}.
To train \emph{BIMNet}, we fine-tuned LiteFlowNet~\cite{LiteFlowNet} over the pre-trained weights provided by the authors.
For fine-tuning, we randomly selected two consecutive frames from the training video set.
We then randomly cropped the same areas of size $256\times 256$ from the selected frames, and fed them to the network.
We used normalized pixel values from -1 to 1, and set the batch size to eight.
We used Adam optimizer~\cite{kingma2015adam} with $\beta_{1}=0.9$ and $\beta_{2}=0.999$.
We initially set the learning rate to 0.0001, and decayed it with a decaying rate 0.1 for every 100,000 iterations.
The network converged after $\sim$200,000 iterations.
The training took about two days on a PC with an Nvidia GeForce TITAN-Xp (12GB).

To train \emph{PVDNet}, we follow a conventional training strategy such as \cite{kim2017online} to generate mini-batches for recurrent video processing. 
Specifically, we randomly sampled sequences of 13 consecutive frames from the training set.
We sampled eight sequences, i.e., we used batch size of eight.
Then, we randomly cropped the same areas of $256\times256$ from the frames in each sequence.
For each sequence, we iterate from the first to last cropped frame. For each iteration, we set the current cropped frame as the target blurry frame $I_t^b$ and compute the gradient using the result of the previous frame $I_{t-1}^{est}$ as well as the current target blurry frame and its neighboring blurry frames.
Then, we update the network parameters by using backpropagation.
The backpropagation is performed only for the current iteration, i.e., the gradient does not propagate to the network at the previous state through $I_{t-1}^{est}$.
After iterating to the last frames, we repeat the process from the random sampling of video frame sequences.
For the initial frame of each sequence, we set $I_{t-1}^{est}=I_{t}^{b}$ and $I_{t-1}^{b}=I_{t}^{b}$.
We used Adam optimizer with $\beta_{1}=0.9$ and $\beta_{2}=0.999$ as was done for \emph{BIMNet}.
We trained \emph{PVDNet} for 400,000 iterations with a learning rate of 0.0001. Then, we further trained the network for 200,000 iterations with the learning rate of 0.000025.
The training of \emph{PVDNet} took about four days on the aforementioned environment.

We do not train \emph{BIMNet} and \emph{PVDNet} jointly, because we use the explicit blur-invariant loss for motion estimation learning.
\emph{BIMNet} had been already trained with highly accurate supervision based on ground truth sharp images, so end-to-end training of \emph{BIMNet} and \emph{PVDNet} together did not improve the accuracy or quality of video deblurring.
Similarly, Gast and Roth~\shortcite{Gast:2019:DVD} reported that fixed pre-trained optical flow models \cite{FlowNet,Sun:2018:PWC} can be used to improve video deblurring accuracy quite well without sophisticated joint training.
It is also known that end-to-end training schemes do not always improve the model performances in various image processing tasks, such as object detection ~\cite{Ren2015,He2017mask}, image super-resolution~\cite{Weifeng2018}, and image inpainting~\cite{ren2019structureflow}. 
%

\begin{table*}[t]
\centering
\caption{Quantitative comparison between \emph{BIMNet} and other optical flow estimation methods on blurry video frames. We computed average errors between $I_{t}^s$ and $Warp(I_{t-1}^s, W^{bb})$ in PSNR and SSIM, where $W^{bb}$ is the optical flow estimated from $I_{t}^b$ to $I_{t-1}^b$.}
\vspace{-8pt}
\begin{tabular}{ | c || c c | c c | c c | }
\hline
 \multirow{2}{*}{Model} & \multicolumn{2}{|c|}{Entire test set} & \multicolumn{2}{|c|}{Sharpest 10\%} & \multicolumn{2}{|c|}{Blurriest 10\%}\\
 & PSNR (dB) & SSIM & PSNR (dB) & SSIM & PSNR (dB) & SSIM\\
\hline \hline
\emph{TVL1}~\cite{perez2013}           & 28.20           & 0.890          & \textbf{30.73}  & 0.924          & 25.08           & 0.784\\
\emph{FlowNet2}~\cite{ilg17flow}       & 27.85           & 0.903          & 29.75           & 0.923          & 25.79           & 0.864\\
\emph{LiteFlowNet}~\cite{LiteFlowNet}  & 27.64           & 0.895          & 29.99           & 0.923          & 24.64           & 0.833\\
\emph{FlowNet2$_{SS}$}                 & 28.21           & \textbf{0.909} & 30.57           & 0.925          & 25.06           & 0.811\\
\emph{LiteFlowNet$_{SS}$}              & 27.91           & 0.893          & 30.12           & 0.923          & 24.89           & 0.830\\
\emph{FlowNet2$_{**}$}     & 27.89           & 0.868          & 30.20           & 0.922          & 24.14           & 0.761\\
\emph{LiteFlowNet$_{**}$}  & 27.23           & 0.811          & 30.06           & 0.906          & 24.07           & 0.753\\
\emph{BIMNet$_{FN2}$}           & \textbf{28.79}  & 0.907          & 30.61           & \textbf{0.926} & \textbf{26.64}  & \textbf{0.873}\\
\emph{BIMNet$_{LFN}$}           & 28.43           & 0.900          & 30.29           & 0.923          & 26.03           & 0.853\\

\hline
\end{tabular}
\label{tbl:mc_table}
\end{table*}

\begin{figure*}[tp]
  \centering
  \begin{subfigure}[t]{0.19\textwidth}
    \includegraphics[width=\textwidth]{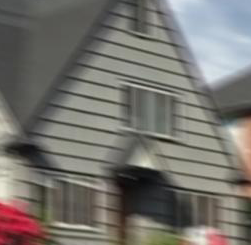}
    \caption{$I_{t-1}^b$}
  \end{subfigure}
  \hfill
  \begin{subfigure}[t]{0.19\textwidth}
    \includegraphics[width=\textwidth]{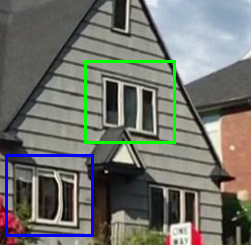}
    \caption{\emph{FlowNet2}}
  \end{subfigure}
  \hfill
  \begin{subfigure}[t]{0.19\textwidth}
    \includegraphics[width=\textwidth]{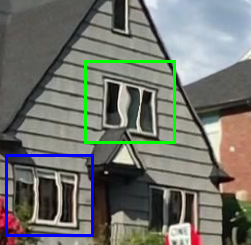}
    \caption{\emph{FlowNet2$_{SS}$}}
  \end{subfigure}
  \hfill
  \begin{subfigure}[t]{0.19\textwidth}
    \includegraphics[width=\textwidth]{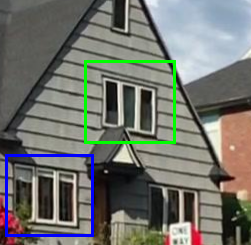}
    \caption{\emph{BIMNet$_{FN2}$}}
  \end{subfigure}
  \hfill
  \begin{subfigure}[t]{0.19\textwidth}
    \includegraphics[width=\textwidth]{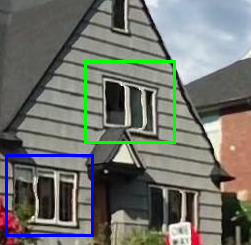}
    \caption{\emph{TVL1}}
  \end{subfigure}

  \vspace{0.3ex}

  \begin{subfigure}[t]{0.19\textwidth}
    \includegraphics[width=\textwidth]{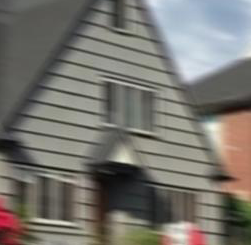}
    \caption{$I_{t}^b$}
  \end{subfigure}
  \hfill
  \begin{subfigure}[t]{0.19\textwidth}
    \includegraphics[width=\textwidth]{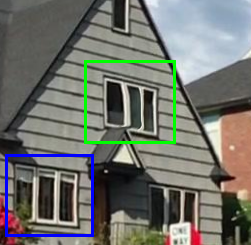}
    \caption{\emph{LiteFlowNet}}
  \end{subfigure}
  \hfill
  \begin{subfigure}[t]{0.19\textwidth}
    \includegraphics[width=\textwidth]{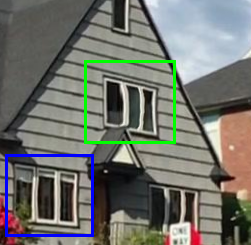}
    \caption{\emph{LiteFlowNet$_{SS}$}}
  \end{subfigure}
  \hfill
  \begin{subfigure}[t]{0.19\textwidth}
    \includegraphics[width=\textwidth]{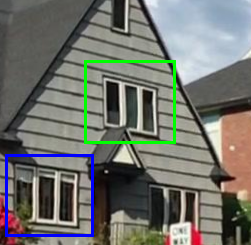}
    \caption{\emph{BIMNet$_{LFN}$}}
  \end{subfigure}
  \hfill
  \begin{subfigure}[t]{0.19\textwidth}
    \includegraphics[width=\textwidth]{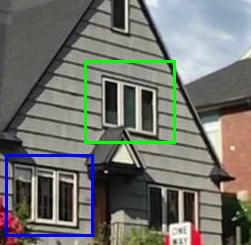}
    \caption{GT}
  \end{subfigure}

  \vspace{-0.2cm}
  \caption{Qualitative comparison between \emph{BIMNet} and other optical flow estimation methods. (a) and (f) are input blurry frames $I_{t-1}^b$ and $I_{t}^b$, respectively.
  From (b) to (i), we show the warped images $I_{t-1}^{'s} = Warp(I_{t-1}^s, W^{bb})$, where the optical flow maps $W^{bb}$ have been estimated from $I_{t-1}^b$ and $I_{t}^b$ by different methods.}
  \vspace{-0.2cm}

\label{fig:mc}
\end{figure*}

\subsection{Motion estimation performance}
\label{ssec:mc_performance}
We first verify the effectiveness of our blur-invariant motion estimation learning
by comparing our \emph{BIMNet} with state-of-the-art optical flow estimation methods: \emph{TVL1}~\cite{perez2013}, \emph{FlowNet2}~\cite{ilg17flow}, and \emph{LiteFlowNet}~\cite{LiteFlowNet}.
\emph{TVL1} is a traditional optimization-based approach.
\emph{FlowNet2} improves \emph{FlowNet} in terms of accuracy at the expense of an increased model size and slower computation.
\emph{LiteFlowNet} is a light-weight variant of \emph{FlowNet} with a much reduced model size and faster computation.
We considered two versions of \emph{BIMNet}: \emph{BIMNet$_{FN2}$, which uses \emph{FlowNet2} network architecture; and  \emph{BIMNet$_{LFN}$}, which uses \emph{LiteFlowNet} network architecture.}
\emph{BIMNet$_{LFN}$} is the model adopted in our final framework.
Our blur-invariant motion estimation learning uses blurry video frames and the blur-invariant loss in the training process.
To verify the effect of each component, we consider four versions of \emph{FlowNet2} and \emph{LiteFlowNet}: \emph{FlowNet2$_{SS}$}, \emph{LiteFlowNet$_{SS}$}, \emph{FlowNet2$_{**}$}, and \emph{LiteFlowNet$_{**}$}. 
\emph{FlowNet2$_{SS}$} and \emph{LiteFlowNet$_{SS}$} are fine-tuned versions of \emph{FlowNet2} and \emph{LiteFlowNet} using sharp video frames in our training set, respectively. 
\emph{FlowNet2$_{**}$} and \emph{LiteFlowNet$_{**}$} are fine-tuned using all combinations of sharp and blurry video frames in our training set as done for \emph{BIMNet$_{FN2}$} and \emph{BIMNet$_{LFN}$}. 
To fine-tune these variants, we did not use a blur-invariant loss in \Eq{L_BIM}. 
Instead, we used the loss function defined as:
\begin{equation}
L^{\alpha\beta} = MSE(Warp(I_{t-1}^\alpha,W^{\alpha\beta}),I_t^\beta) ,
\label{eq:blur_variant_loss}
\end{equation}
which is blur-variant, where $\alpha, \beta \in \{s,b\}$.

To measure the performance of each method, we measured the warping accuracy of motion estimation results.
Specifically, we first estimated the optical flow from $I_{t}^{b}$ to $I_{t-1}^{b}$.
We then obtained $I_{t}^{'s}$ by warping $I_{t-1}^{s}$ with the computed optical flow, and measured the difference between the warping result $I_{t}^{'s}$ and its ground truth sharp frame $I_{t}^{s}$, in terms of peak signal-to-noise ratio (PSNR) and structural similarity index (SSIM)~\cite{wang2004ssim}.
Following Su \Etal~\shortcite{su2017deep}, to handle the ambiguity in the pixel location caused by blur, we measure PSNRs and SSIMs using an approach \cite{kohler2012record} that first aligns two images using global translation.
To analyze the performance of different methods on blurry videos more clearly, we also measured the performance on the sharpest 10\% and the blurriest 10\% video frames in each video in the test set. 
To choose the sharpest and blurriest 10\% frames, we measured the blurriness of all test frames as PSNRs between blurry frames and their corresponding sharp ground truth frames. 
We assume that a test frame with a low PSNR is likely to have large blur, because blur is the primary factor of structural differences from its ground truth. 

\Tbl{mc_table} shows a summary of the quantitative comparison using our test set.
For the entire test set, \emph{BIMNet$_{FN2}$} and \emph{BIMNet$_{LFN}$} achieve the best and second-best PSNR and comparable SSIM.
The result shows that \emph{BIMNet$_{FN2}$} significantly outperforms the other methods on the severely blurred frames by a large margin, while it performs comparably on the relatively sharp frames.
\emph{BIMNet$_{LFN}$} was slightly inferior to \emph{BIMNet$_{FN2}$}, but still outperforms all the other methods.
Moreover, \emph{BIMNet$_{LFN}$} has only 5.4M parameters, whereas \emph{BIMNet$_{FN2}$} has 150M parameters.
Thus, we adopt \emph{BIMNet$_{LFN}$} for our framework.

To identify the source of the improvement of \emph{BIMNet},
we compared \emph{FlowNet2}, \emph{FlowNet2$_{SS}$}, \emph{FlowNet2$_{**}$}, and \emph{BIMNet$_{FN2}$}.
Compared to \emph{FlowNet2}, \emph{FlowNet2$_{SS}$} shows improvement for relatively sharp frames, but yields inferior results for blurry frames as a result of overfitting to sharp video frames.
\emph{FlowNet2$_{**}$} performs worse than \emph{FlowNet2$_{SS}$} for both sharp and blurry frames; this result shows that simply including blurry frames in training does not improve the accuracy of optical flow estimation on blurry frames.
Compared to these, \emph{BIMNet$_{FN2}$} that is trained using our blur-invariant loss shows improvements in all cases.
A similar trend can also be found from \emph{LiteFlowNet$_{SS}$}, \emph{LiteFlowNet$_{**}$}, and \emph{BIMNet$_{LFN}$}.
This result verifies that our blur-invariant loss is the source of the improvement in motion estimation accuracy in the presence of blur.

\Fig{mc} shows a qualitative comparison of different optical flow methods.
To visually compare the quality of different optical flow methods, we estimated optical flow maps between two consecutive blurred video frames, $I_{t-1}^b$ and $I_{t}^b$, using different optical flow methods,
and warped the sharp video frame $I_{t-1}^s$ corresponding to $I_{t-1}^b$ using the estimated optical flow maps.
As the figure shows, the results of the other methods contain severely distorted structures due to errors in their optical flow maps.
In contrast, the results of \emph{BIMNet$_{FN2}$} and \emph{BIMNet$_{LFN}$} show much less distortions.

\subsection{Ablation study}
\label{ssec:ablation}

To show the effectiveness of the \emph{BIMNet} and pixel volume in our video deblurring framework, we performed an ablation study.
We tested five models in this study.
The baseline is \emph{PVDNet} that receives consecutive blurry frames as well as a deblurred previous frame without any motion compensation.
Then, one by one, we add motion compensation (MC) using \emph{LiteFlowNet$_{SS}$}, blur-invariant motion compensation (BIMC) using \emph{BIMNet$_{LFN}$}, and pixel volume (PV) into the baseline model. 
Another straightforward idea to increase the performance is to jointly train motion estimation and deblurring. To verify the effectiveness of our BIMC against such a joint approach, we also include a jointly trained model (MC$_{e2e}$ + PV) in which the motion estimation module is first initialized by \emph{LiteFlowNet$_{SS}$} and then jointly trained with \emph{PVDNet} by using the deblurring loss in an end-to-end manner.

Our model that uses BIMC and PV shows the best PSNR and SSIM (\Tbl{ablation_table}).
For the sharpest 10\% of the frames of each video in the test set, all the models, even the baseline model without motion compensation, show high performances.
In contrast, for the blurriest 10\% of the frames, each of BIMC and PV achieves drastic improvement over the models with and without MC; these results indicate that BIMC and PV are effective for handling large blur.
The table also shows that joint training of motion estimation and \emph{PVDNet} (MC$_{e2e}$ + PV) does not increase but decreases the deblurring performance compared to MC + PV. This result occurs because the deblurring loss does not provide good guidance for accurate motion estimation, and the loss adversarially affects its training.

\begin{figure}[tp]
\centering
\setlength\tabcolsep{1 pt}
  \begin{tabular}{ccccc}
    \rotatebox[origin=l]{90}{  \,  Input}  &
  	\multicolumn{1}{c}{\includegraphics[width=0.31\linewidth]{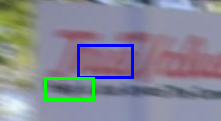}} &
  	\multicolumn{1}{c}{\includegraphics[width=0.31\linewidth]{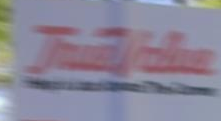}} &
  	\multicolumn{1}{c}{\includegraphics[width=0.31\linewidth]{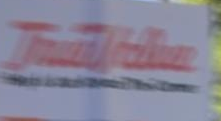}} \\

    \rotatebox[origin=l]{90}{  Baseline} &
  	\multicolumn{1}{c}{\includegraphics[width=0.31\linewidth]{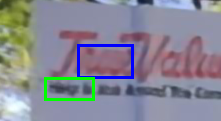}} &
  	\multicolumn{1}{c}{\includegraphics[width=0.31\linewidth]{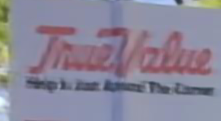}} &
  	\multicolumn{1}{c}{\includegraphics[width=0.31\linewidth]{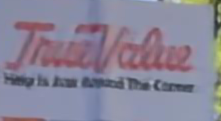}} \\

    \rotatebox[origin=l]{90}{  \,  \,  MC} &
  	\multicolumn{1}{c}{\includegraphics[width=0.31\linewidth]{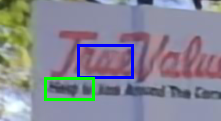}} &
  	\multicolumn{1}{c}{\includegraphics[width=0.31\linewidth]{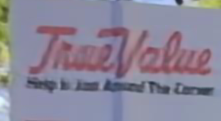}} &
  	\multicolumn{1}{c}{\includegraphics[width=0.31\linewidth]{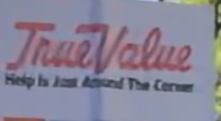}} \\

    \rotatebox[origin=l]{90}{\,  \,  BIMC} &
  	\multicolumn{1}{c}{\includegraphics[width=0.31\linewidth]{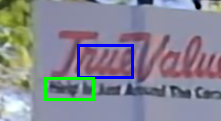}} &
  	\multicolumn{1}{c}{\includegraphics[width=0.31\linewidth]{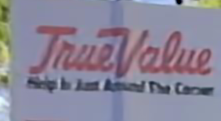}} &
  	\multicolumn{1}{c}{\includegraphics[width=0.31\linewidth]{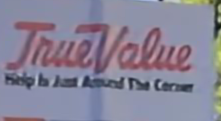}} \\

    \rotatebox[origin=l]{90}{  \,  MC+PV} &
  	\multicolumn{1}{c}{\includegraphics[width=0.31\linewidth]{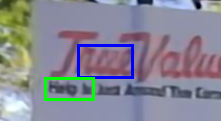}} &
  	\multicolumn{1}{c}{\includegraphics[width=0.31\linewidth]{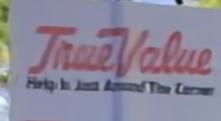}} &
  	\multicolumn{1}{c}{\includegraphics[width=0.31\linewidth]{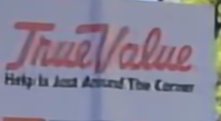}} \\

    \rotatebox[origin=l]{90}{BIMC+PV} &
  	\multicolumn{1}{c}{\includegraphics[width=0.31\linewidth]{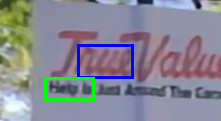}} &
  	\multicolumn{1}{c}{\includegraphics[width=0.31\linewidth]{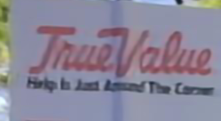}} &
  	\multicolumn{1}{c}{\includegraphics[width=0.31\linewidth]{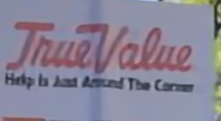}} \\

  \end{tabular}
  \vspace{-0.3cm}
  \caption{Qualitative comparison of the models in \Tbl{ablation_table} for three consecutive video frames.}
  \vspace{-0.3cm}
\label{fig:ablation}
\end{figure}

\Fig{ablation} shows a qualitative comparison.
The models without BIMC cannot compensate for large motions, and therefore did not well restore the big red characters in the blue box.
The models without PV cannot compensate for small distortions caused by motion estimation errors, and did not properly recover the small black characters in the green box.
Our full model with BIMC and PV could deblur successfully in both cases, and produced the best results.

\begin{table}[t]
\centering
\caption{Ablation study for video deblurring performance. PSNR are in dB.}
\vspace{-8pt}
\begin{tabular}{ | c || c c | c c | c c | }
\hline
\multirow{2}{*}{Model} & \multicolumn{2}{c}{Entire test set} & \multicolumn{2}{|c|}{Sharpest 10\%} & \multicolumn{2}{|c|}{Blurriest 10\%}\\
 & PSNR & SSIM & PSNR & SSIM & PSNR & SSIM\\
\hline \hline
input & 27.40 & 0.819 & 33.02 & 0.936 & 22.77 & 0.665\\
baseline & 30.66 & 0.891 & 34.76 & 0.954 & 26.15 & 0.780 \\
MC & 31.23 & 0.901 & 35.10 & \bf{0.957} & 26.69 & 0.798\\
BIMC & 31.46 & 0.912 & 35.15 & \bf{0.958} & 27.28 & 0.820\\
MC + PV & 31.40 & 0.910 & 35.13 & \bf{0.958} & 27.19 & 0.817\\
MC$_{e2e}$ + PV & 31.32 & 0.909 & 34.95 & 0.956 & 27.21 & 0.818 \\
BIMC + PV & \bf{31.63} & \bf{0.915} & \bf{35.18} & \bf{0.958} & \bf{27.46} & \bf{0.828}\\
\hline
\end{tabular}
\vspace{-0.3cm}
\label{tbl:ablation_table}
\end{table}

\subsection{Statistics of candidate pixels in pixel volume}
\label{ssec:pv_statistic}
From each video in the test set of \cite{su2017deep}, we sampled ten frames, and gathered 100 frames. We then constructed PVs for those frames, and investigated the distribution of pixel values in the volumes. In the PVs, 95.1\% of the pixels had majority values in the candidates, where a majority value means more than a half of the $k^2$ candidates for a pixel have the same value. We then investigated the accuracy of those majority values. As our PVs are constructed from deblurred previous frames, they may not have perfect matches with ground truth pixel values. Thus, to measure the accuracy, we counted the number of majority values that are close enough to the ground truth pixel values, i.e., the pixel value differences are within $\pm c$ in 0-255 scale. The accuracy of those majority values was 89.5\% when $c$ = 0 (perfect match) and increased to 92\% for $c$ = 2. 

While most of the pixels in the pixel volume have correct majority values, the pixel volume is still helpful even for deblurring regions that have no majority values or wrong ones.
To validate it, we measured the accuracy of deblurring results on those regions using two models: one that uses PV, and the other that uses na\"{i}ve warping.
The accuracy was measured in the same way as above with $c$ = 2. 
In regions where the majority values are correct,
the model that uses PV achieved accuracy of 97.42\%, and the model that uses na\"{i}ve warping achieved 96.91\%. 
For regions in which majority values are wrong,
the model that uses PV (57.35\%) shows a significantly higher accuracy than the model that uses na\"{i}ve warping (50.49\%).
Likewise, for the region that has no majority values, 
the model that uses PV (89.97\%) also shows higher accuracy than the model that uses na\"{i}ve warping (87.77\%).
This result confirms that our PV helps our deblurring network obtain the correct values even when the majority pixels in the PV are wrong.


\subsection{Motion compensation \textit{vs.} larger model}
\label{ssec:model_size}
Our method adopts an additional motion estimation network, \emph{BIMNet}, so the accuracy gain could be simply a result of increasing the total model size.
To investigate on this possibility, we prepared variants of our deblurring framework with different model sizes either with or without motion compensation.
Specifically, we prepared small, medium, and large versions of \emph{PVDNet}.
The medium model is the model described in \SSec{RDN}.
The small model has half the number of filters in every convolution layer, and half the number of residual blocks of the medium model.
The large model has twice the number of filters in every convolution layer.
Then, we prepared six variants of our framework either with or without our motion compensation scheme that combines \emph{BIMNet} and PV, and measured the models' deblurring accuracies.

\Tbl{model_size} shows that the small and medium models with our motion compensation scheme outperformed the large model without motion compensation despite their significantly smaller model sizes.
\Fig{model_size_vis} also shows that the small and medium models with motion compensation could recover small-scale structures and textures better than the large model without motion compensation.
Both quantitative and qualitative comparisons clearly show the superiority of our motion compensation scheme over simply increasing the size of the model.

\begin{table}[t]
\centering
\caption{Performance comparison of \emph{PVDNet}s with different model sizes either with or without MC. The unit of the number of parameters is million (M).
Values in the parentheses are the numbers of parameters of \emph{PVDNet} and \emph{BIMNet}.}
\vspace{-8pt}
\begin{tabular}{ |c || c | c | c | }
\hline

Model & PSNR (dB) & SSIM & Parameters (M)  \\
\hline \hline
Small & 30.03 & 0.880 & 0.7\\
Medium & 30.66 & 0.891 & 5.1\\
Large & 30.85 & 0.898 & 19.6\\
Small + MC & 30.95 & 0.904 & 6.1 \small(0.7 + 5.4)\\
Medium + MC & 31.63 & 0.915 & 10.5 \small(5.1 + 5.4)\\
Large + MC & 31.75 & 0.918 & 25.0 \small(19.6 + 5.4)\\
\hline

\hline
\end{tabular}
  \vspace{-0.2cm}
\label{tbl:model_size}
\end{table}

\begin{table*}[t]
\centering
\caption{Quantitative comparison of our video deblurring framework with recent state-of-the-art methods on the dataset of \cite{su2017deep}.
Values in the parentheses are the numbers of parameters or computations of \emph{PVDNet} and \emph{BIMNet}.
For DVD-MC, the numbers of parameters and computations for motion estimation are not included, as DVD-MC does not use a deep learning-based model for motion estimation.
CA denotes the cosine annealing scheduling method for learning rate~\cite{Loshchilov2017}, which EDVR adopts for its training.
}
\vspace{-8pt}
\setlength{\tabcolsep}{16pt}
\begin{tabular}{ |c || c c c c | }
\hline
 & PSNR \small{(dB)}  & SSIM & Parameters \small{(M)}  & Computations \small{(BMACs)} \\
\hline \hline
\cite{nah2017deep} & 29.85 & 0.880 & 75.7 & 2364\\
\cite{tao2018scale} & 30.53 & 0.894 & 3.8 & 1175\\
\hline
DVD-noMC \small{\cite{su2017deep}} & 30.29 & 0.888 & 15.3 & 557\\
DVD-MC \small{\cite{su2017deep}}  & 30.47 & 0.898 & 15.3 & 557\\
OVD \small{\cite{kim2017online}}    & 30.08 & 0.873 & 0.9 & 206\\
IFI-RNN \small{\cite{Nah2019recurrent}} & 31.04 & 0.903 & 1.6 & 200 \\
IFI-RNN-L \small{\cite{Nah2019recurrent}} & 31.67 & 0.916 & 12.2 & 1,425 \\
EDVR \small{\cite{Wang2019edvr}}& 31.82 & 0.916 & 23.6 & 2,739 \\
\hline
Ours-small     & 31.25 & 0.908 & 6.1 \small{(0.7 + 5.4)} & 424 \small{(105 + 319 )} \\
Ours     & 31.95 & 0.920 & 10.5 \small{(5.1 + 5.4)} & 936 \small{(617 + 319)}\\
Ours w/ CA & \bf{32.31} & \bf{0.926} & 10.5 \small{(5.1 + 5.4)} & 936 \small{(617 + 319)}\\
\hline
\end{tabular}
\label{tbl:result_table}
\end{table*}

\subsection{Comparison with other deblurring methods}
\label{ssec:comparison}
We compared our method with six recent state-of-the-art deep learning based deblurring methods, including two single image-based methods~\cite{nah2017deep,tao2018scale}, and four video deblurring methods~\cite{su2017deep,kim2017online,Nah2019recurrent,Wang2019edvr}.
For the single image deblurring methods of Nah~\Etal~\shortcite{nah2017deep} and Tao~\Etal~\shortcite{tao2018scale}, we applied them to each input frame independently.
The method of Su \Etal~\shortcite{su2017deep} takes consecutive blurry video frames as input, which can be either motion compensated using optical flow or not.
In our comparison, we included both versions with and without motion compensation, which are denoted by DVD-MC and DVD-noMC, respectively.
For motion compensation for DVD-MC, we used the method of S\'{a}nchez \Etal~\shortcite{perez2013} as done in \cite{su2017deep}.
The method of Wang \Etal~\shortcite{Wang2019edvr}, which we refer to as EDVR, also takes consecutive blurry video frames as input.
The methods of Kim \Etal~\shortcite{kim2017online} and Nah \Etal~\shortcite{Nah2019recurrent}, which we refer to as OVD and IFI-RNN, respectively, adopt recurrent network structures that take an intermediate feature map of the previous frame without motion compensation.

For fair comparison in terms of model size, we also included a variant of IFI-RNN with an increased model size, which we refer to as IFI-RNN-L,
and a smaller version of our framework (the small \emph{PVDNet} in \SSec{model_size}), which we refer to as Ours-small.
IFI-RNN-L has twice the number of residual blocks and also twice the number of channels in every convolution layer.
Finally, for fair comparison with EDVR, we also include another variant of our framework (Ours w/ CA), which is trained using cosine annealing (CA)~\cite{Loshchilov2017}, a sophisticated learning rate scheduling strategy adopted to EDVR.

For evaluation, we used three test sets in the comparison.
Specifically, in addition to Su \Etal~\shortcite{su2017deep}'s dataset, we used another popular video deblurring dataset \cite{nah2017deep}, synthetically generated in a similar way to Su \Etal~\shortcite{su2017deep}'s dataset. We also used real-world videos \cite{su2017deep} to check the generalization ability of our model.

\begin{figure}[tp]
\centering
\setlength\tabcolsep{1 pt}
  \begin{tabular}{cccccccc}
    \multicolumn{2}{c}{\includegraphics[width=0.24\linewidth]{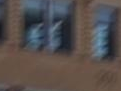}} &
    \multicolumn{2}{c}{\includegraphics[width=0.24\linewidth]{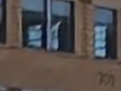}}&
    \multicolumn{2}{c}{\includegraphics[width=0.24\linewidth]{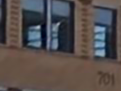}}&
    \multicolumn{2}{c}{\includegraphics[width=0.24\linewidth]{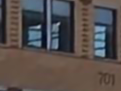}} \\

    \multicolumn{2}{c}{\includegraphics[width=0.24\linewidth]{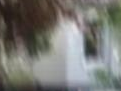}} &
    \multicolumn{2}{c}{\includegraphics[width=0.24\linewidth]{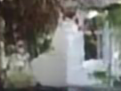}}&
    \multicolumn{2}{c}{\includegraphics[width=0.24\linewidth]{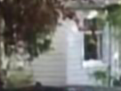}}&
    \multicolumn{2}{c}{\includegraphics[width=0.24\linewidth]{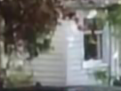}} \\

    \multicolumn{2}{c}{\small{(a) Input}}  & \multicolumn{2}{c}{\small{(b) Large}}
     & \multicolumn{2}{c}{\small{(c) Small + MC}} & \multicolumn{2}{c}{\small{(d) Medium + MC}} \\
  \end{tabular}
  \vspace{-0.2cm}
  \caption{Qualitative comparison of \emph{PVDNet}s with different model sizes either with or without MC.
  The small and medium \emph{PVDNet}s with MC restore image details better than the large \emph{PVDNet}, showing that MC is more effective than simply increasing the model size.}
  \vspace{-0.2cm}
\label{fig:model_size_vis}
\end{figure}

\begin{figure*}[tp]
\centering
\setlength\tabcolsep{1 pt}
  \begin{tabular}{cccccccc}
    \multicolumn{2}{c}{\includegraphics[width=0.24\linewidth]{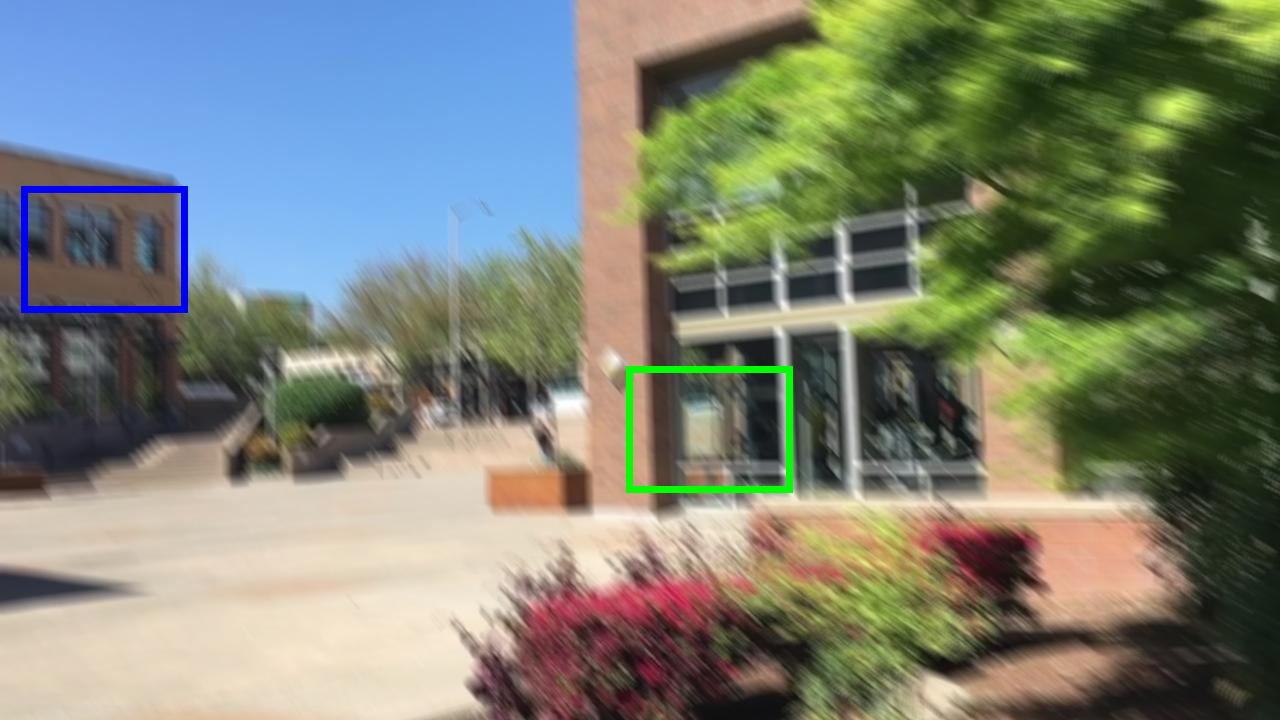}} &
    \multicolumn{2}{c}{\includegraphics[width=0.24\linewidth]{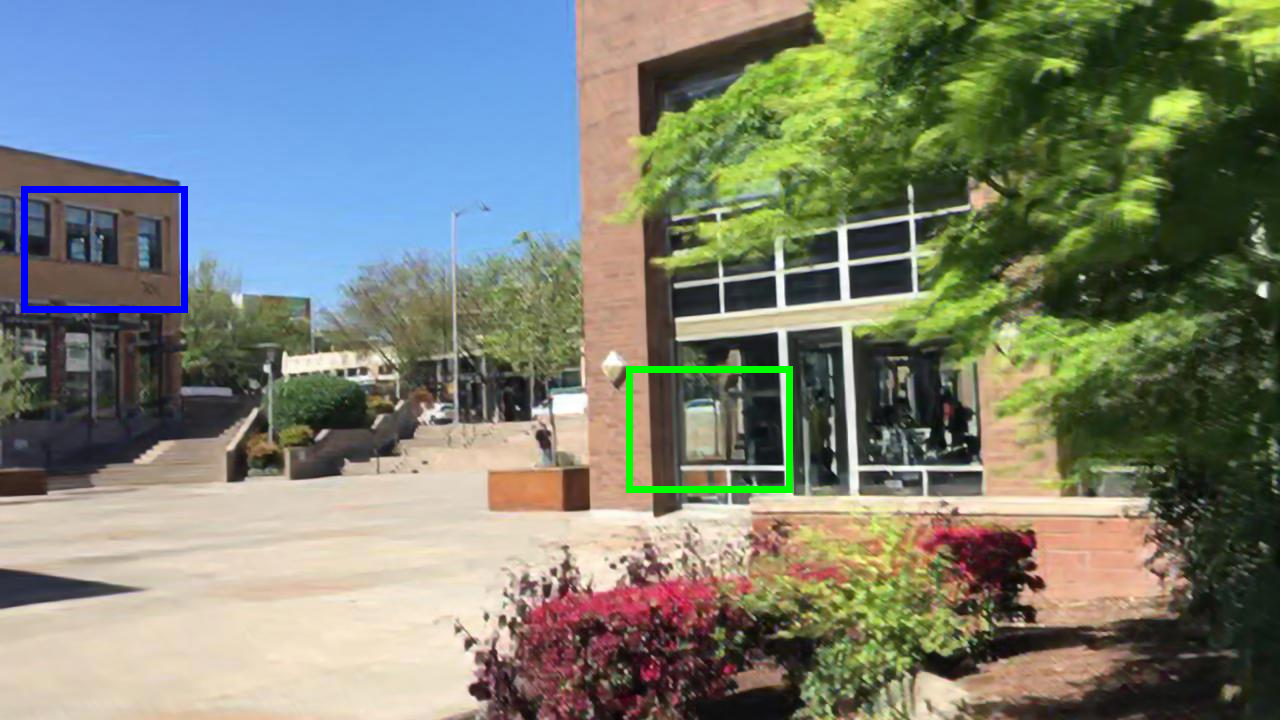}}&
    \multicolumn{2}{c}{\includegraphics[width=0.24\linewidth]{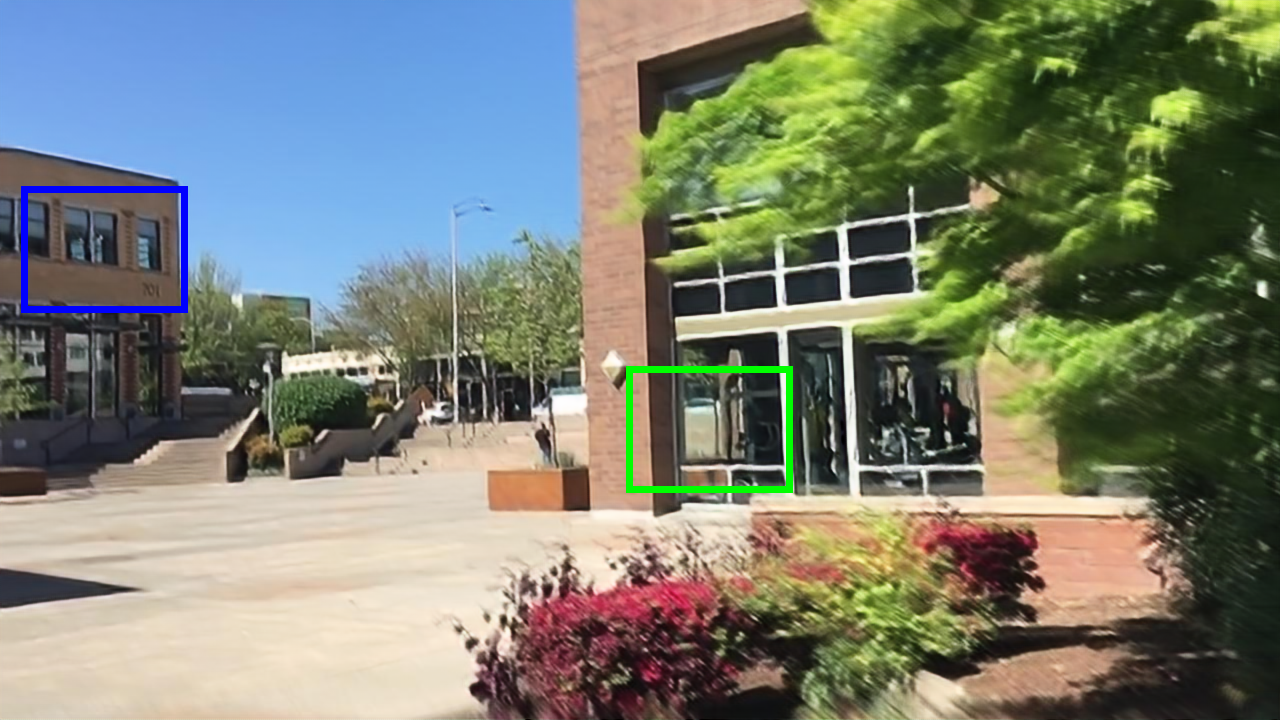}} &
    \multicolumn{2}{c}{\includegraphics[width=0.24\linewidth]{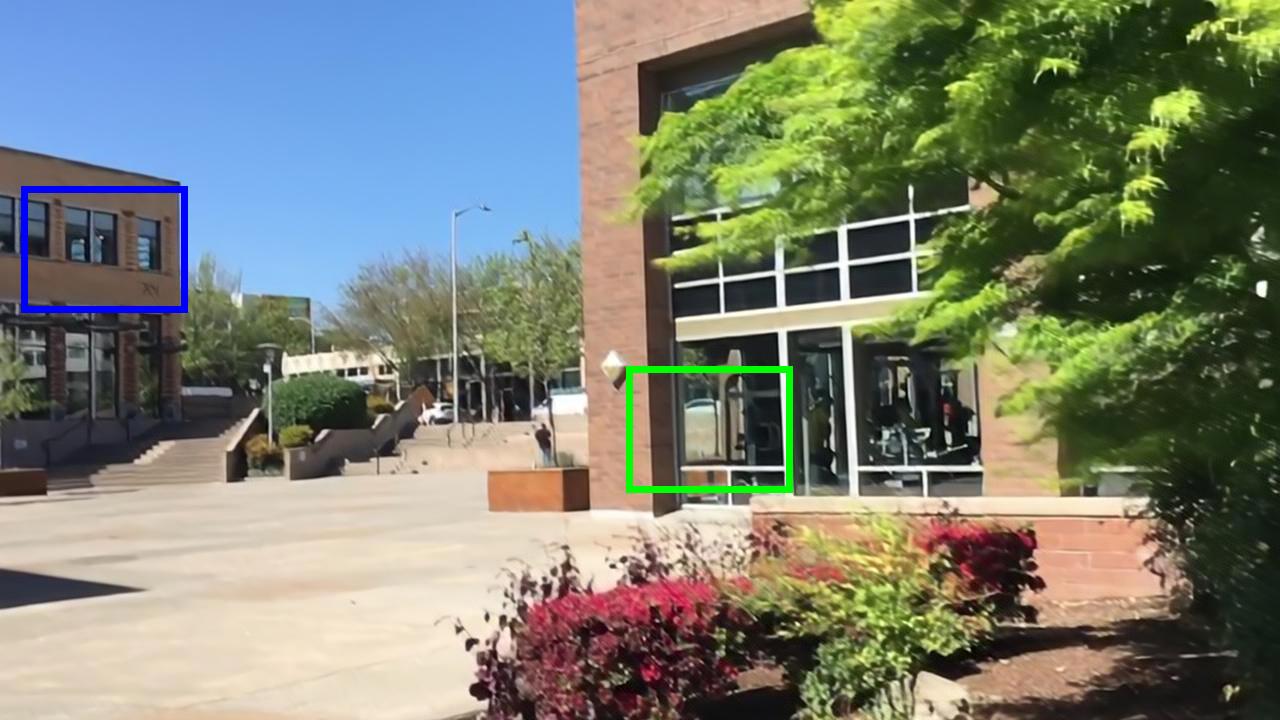}}\\

    \multicolumn{1}{c}{\includegraphics[width=0.118\linewidth]{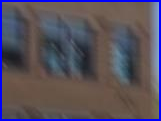}} &
    \multicolumn{1}{c}{\includegraphics[width=0.118\linewidth]{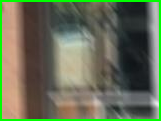}} &
    \multicolumn{1}{c}{\includegraphics[width=0.118\linewidth]{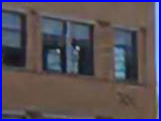}} &
    \multicolumn{1}{c}{\includegraphics[width=0.118\linewidth]{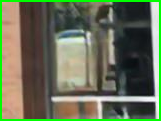}} &
    \multicolumn{1}{c}{\includegraphics[width=0.118\linewidth]{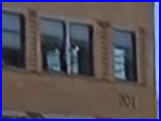}} &
    \multicolumn{1}{c}{\includegraphics[width=0.118\linewidth]{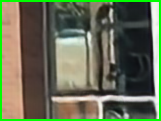}} &
    \multicolumn{1}{c}{\includegraphics[width=0.118\linewidth]{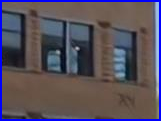}} &
    \multicolumn{1}{c}{\includegraphics[width=0.118\linewidth]{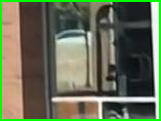}} \\

    \multicolumn{2}{c}{(a) Input}  & \multicolumn{2}{c}{(b) \cite{tao2018scale}}
     & \multicolumn{2}{c}{(c) DVD-MC \cite{su2017deep}} & \multicolumn{2}{c}{(d) EDVR \cite{Wang2019edvr}}\\

    \multicolumn{2}{c}{\includegraphics[width=0.24\linewidth]{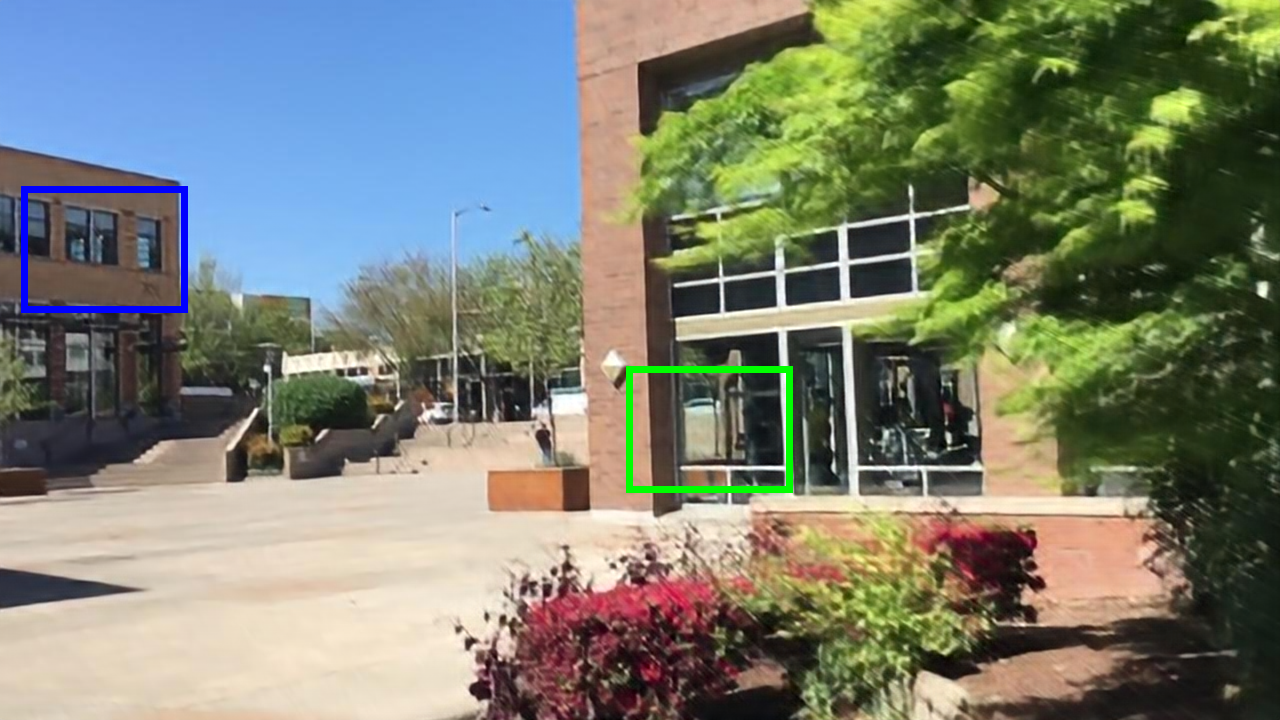}}&
    \multicolumn{2}{c}{\includegraphics[width=0.24\linewidth]{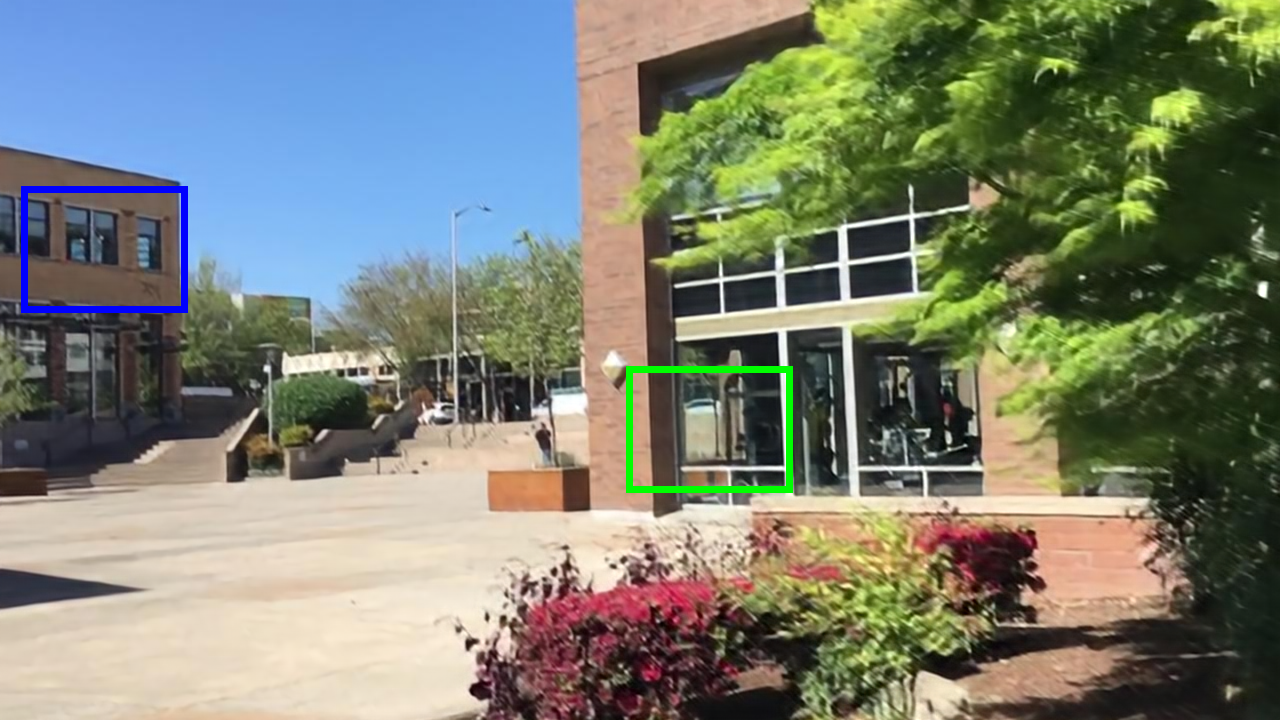}}&
    \multicolumn{2}{c}{\includegraphics[width=0.24\linewidth]{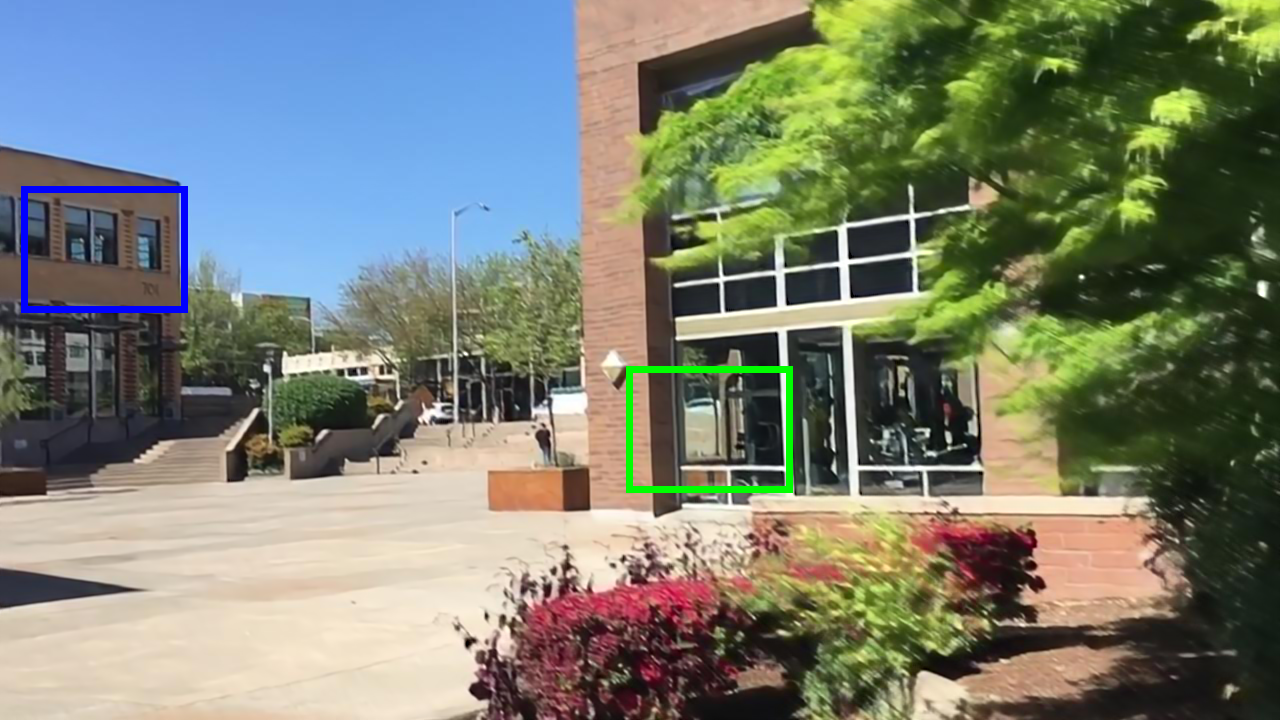}} &
    \multicolumn{2}{c}{\includegraphics[width=0.24\linewidth]{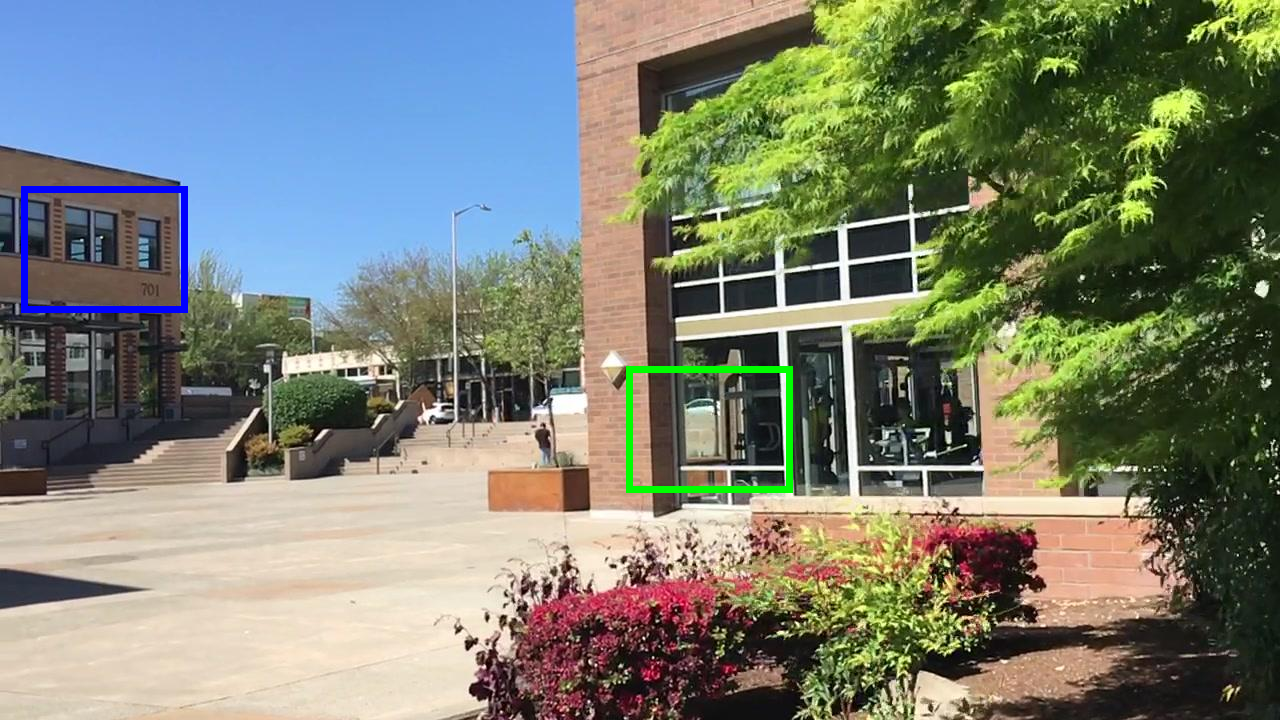}}\\

    \multicolumn{1}{c}{\includegraphics[width=0.118\linewidth]{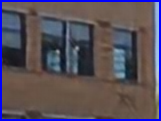}} &
    \multicolumn{1}{c}{\includegraphics[width=0.118\linewidth]{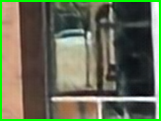}} &
    \multicolumn{1}{c}{\includegraphics[width=0.118\linewidth]{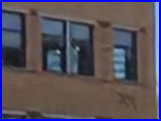}} &
    \multicolumn{1}{c}{\includegraphics[width=0.118\linewidth]{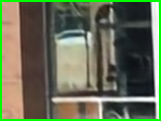}} &
    \multicolumn{1}{c}{\includegraphics[width=0.118\linewidth]{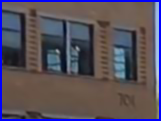}} &
    \multicolumn{1}{c}{\includegraphics[width=0.118\linewidth]{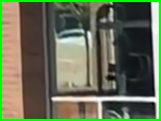}} &
    \multicolumn{1}{c}{\includegraphics[width=0.118\linewidth]{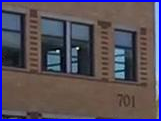}} &
    \multicolumn{1}{c}{\includegraphics[width=0.118\linewidth]{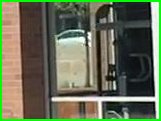}} \\

    \multicolumn{2}{c}{(e) IFI-RNN \cite{Nah2019recurrent}}  & \multicolumn{2}{c}{(f) IFI-RNN-L \cite{Nah2019recurrent}}
     & \multicolumn{2}{c}{(g) Ours} & \multicolumn{2}{c}{(h) GT}\\
  \end{tabular}
  \vspace{-0.3cm}
  \caption{Qualitative comparison on synthetically blurred video frames of Su \Etal~\shortcite{su2017deep}'s dataset.}
\label{fig:gopro}
\end{figure*}

\begin{figure*}[tp]
\centering
\setlength\tabcolsep{1 pt}
  \begin{tabular}{cccccccc}
    \multicolumn{2}{c}{\includegraphics[width=0.245\linewidth]{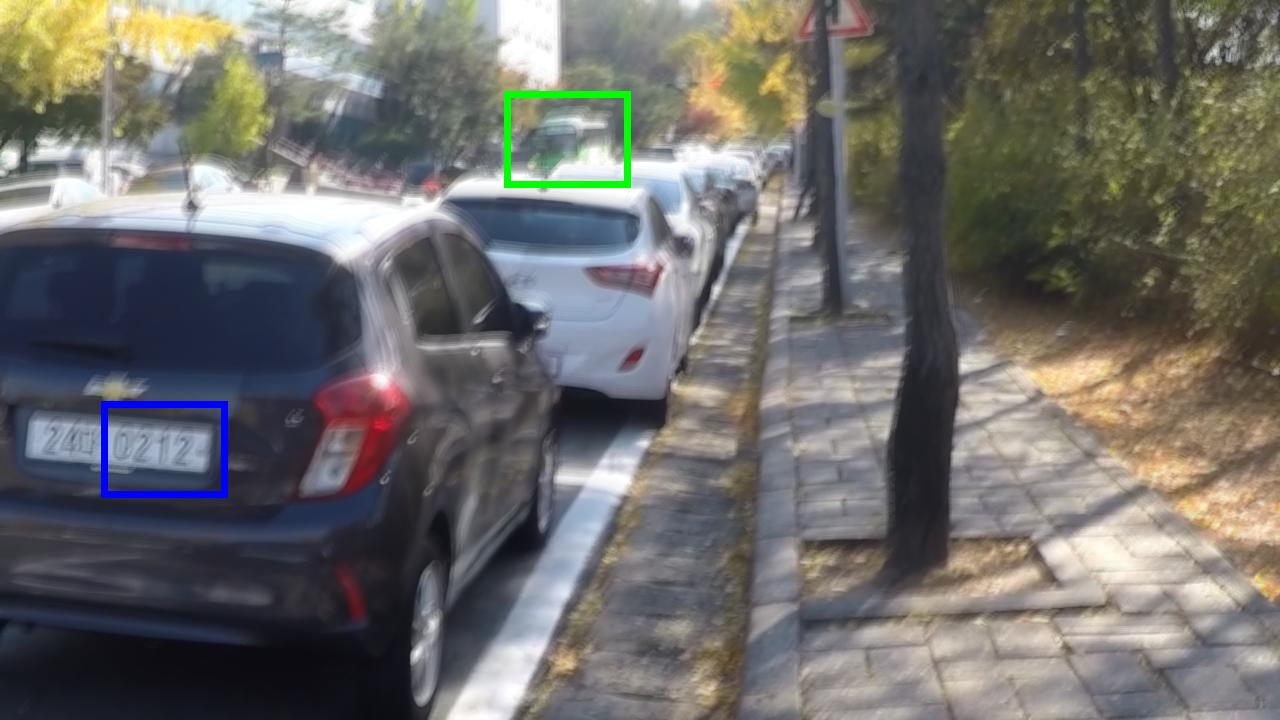}} &
    \multicolumn{2}{c}{\includegraphics[width=0.245\linewidth]{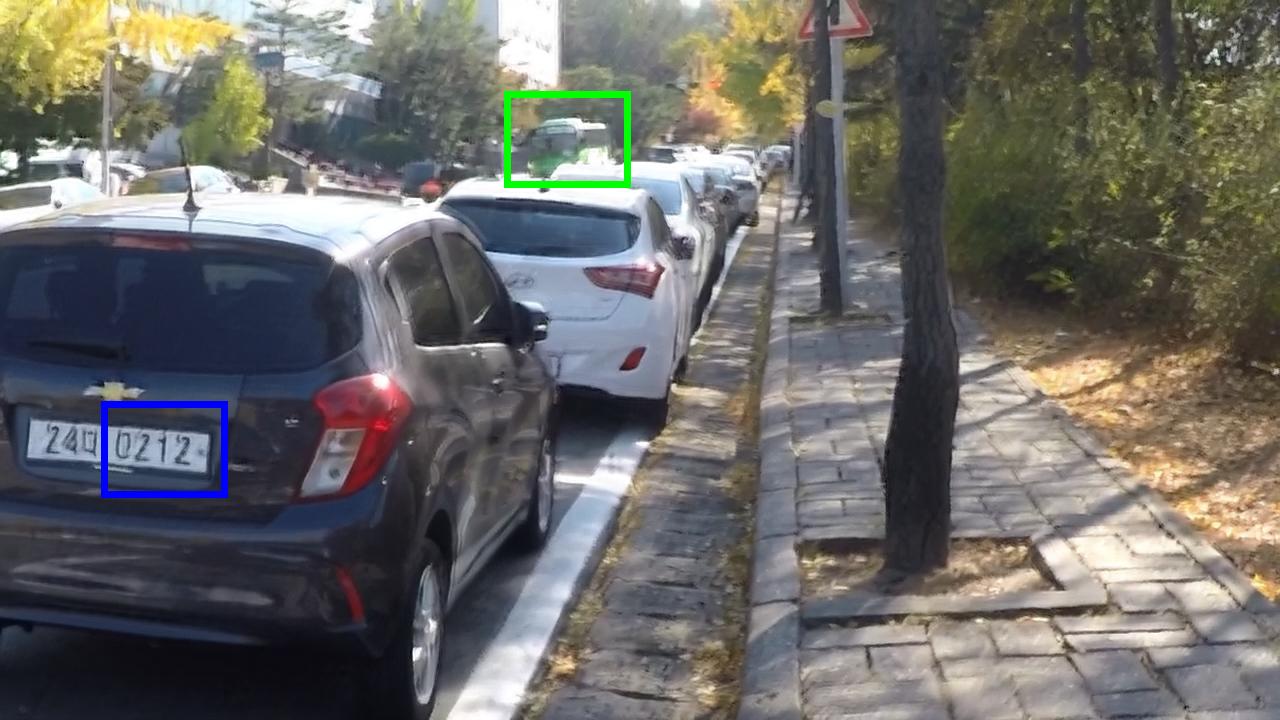}}&
    \multicolumn{2}{c}{\includegraphics[width=0.245\linewidth]{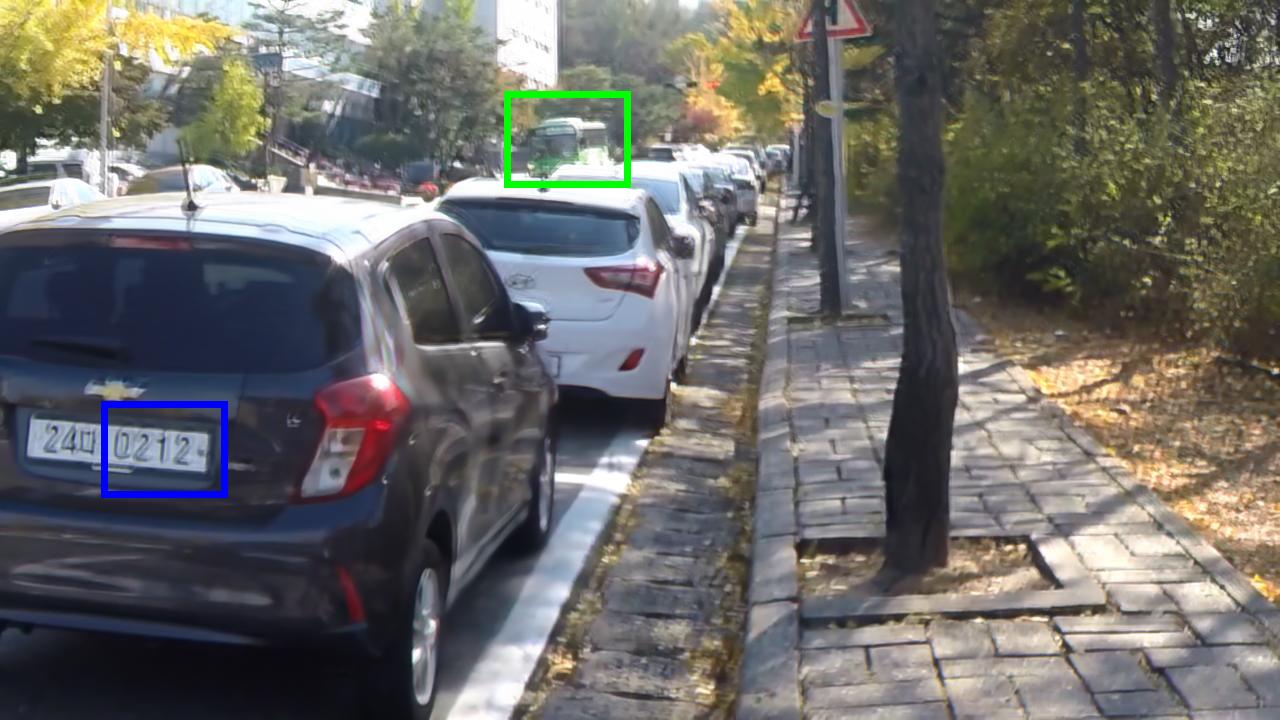}} &
    \multicolumn{2}{c}{\includegraphics[width=0.245\linewidth]{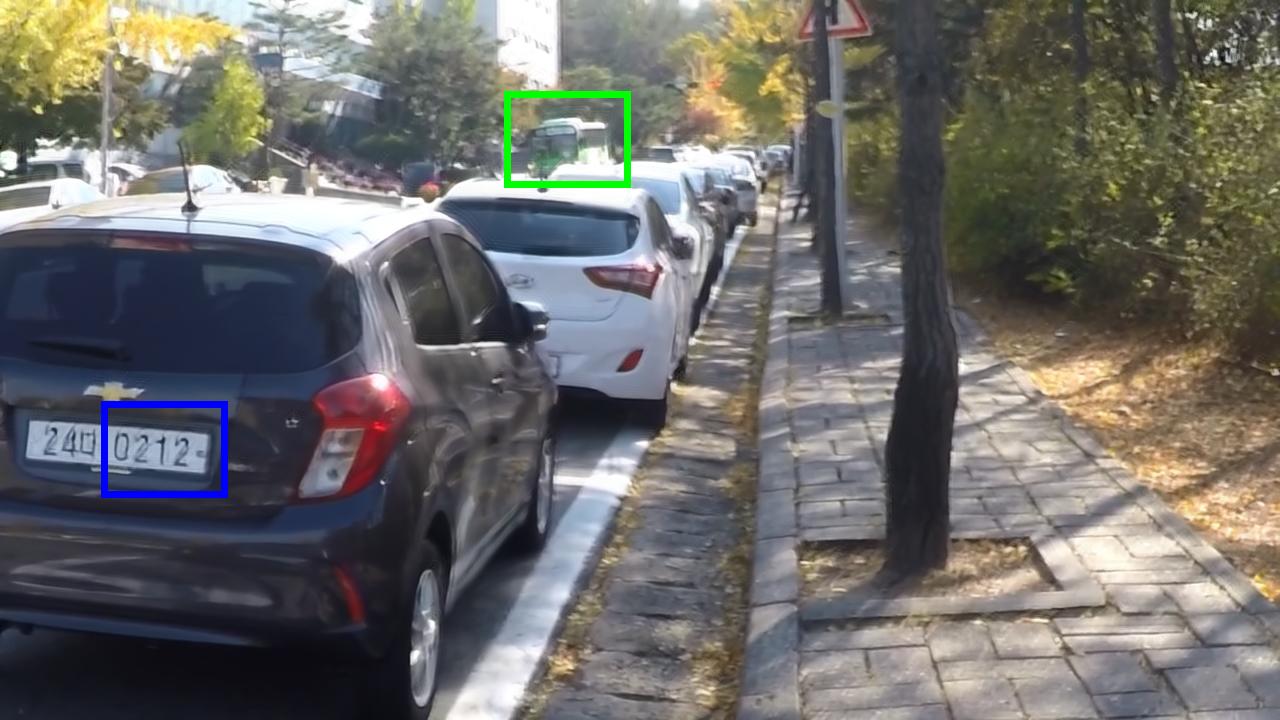}}\\

    \multicolumn{1}{l}{\includegraphics[width=0.12\linewidth]{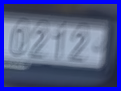}} &
    \multicolumn{1}{r}{\includegraphics[width=0.12\linewidth]{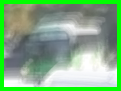}} &
    \multicolumn{1}{l}{\includegraphics[width=0.12\linewidth]{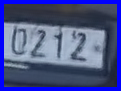}} &
    \multicolumn{1}{r}{\includegraphics[width=0.12\linewidth]{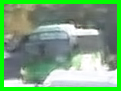}} &
    \multicolumn{1}{l}{\includegraphics[width=0.12\linewidth]{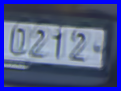}} &
    \multicolumn{1}{r}{\includegraphics[width=0.12\linewidth]{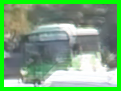}} &
    \multicolumn{1}{l}{\includegraphics[width=0.12\linewidth]{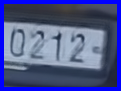}} &
    \multicolumn{1}{r}{\includegraphics[width=0.12\linewidth]{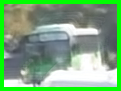}} \\

    \multicolumn{2}{c}{(a) Input}  & \multicolumn{2}{c}{(b) \cite{nah2017deep}}
     & \multicolumn{2}{c}{(c) \cite{tao2018scale}} & \multicolumn{2}{c}{(d) EDVR \cite{Wang2019edvr}}\\

    \multicolumn{2}{c}{\includegraphics[width=0.245\linewidth]{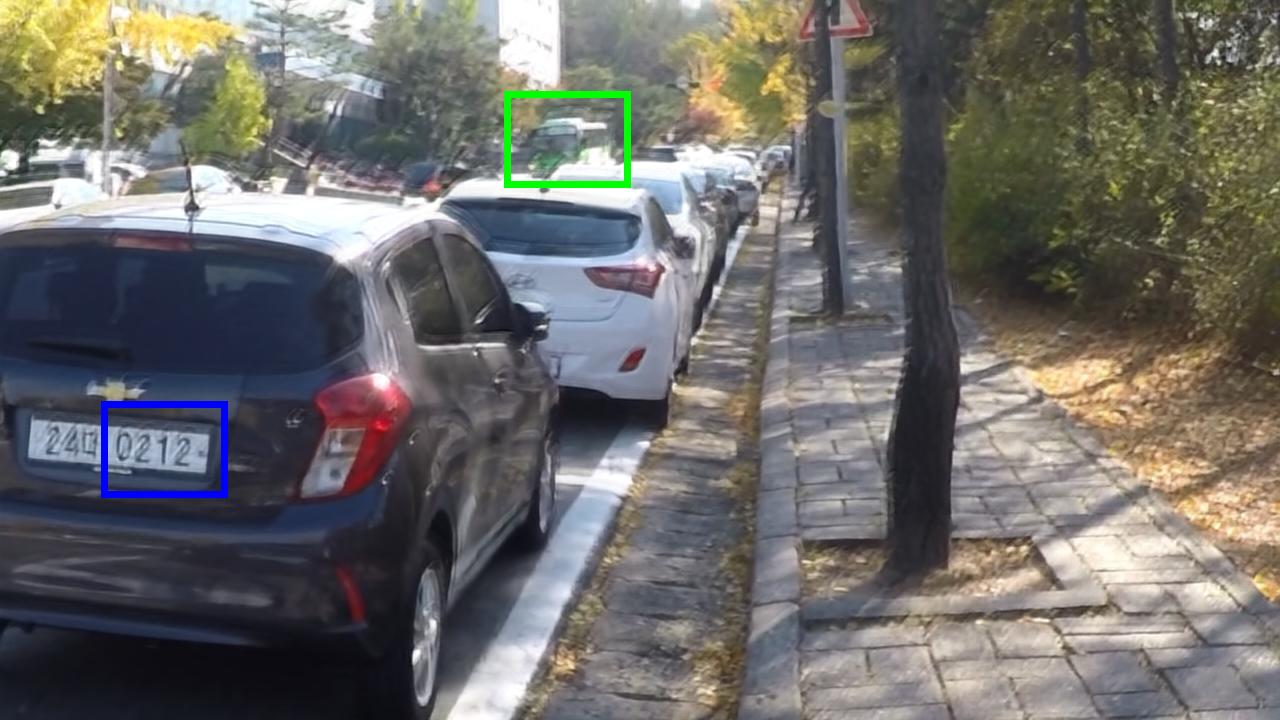}}&
    \multicolumn{2}{c}{\includegraphics[width=0.245\linewidth]{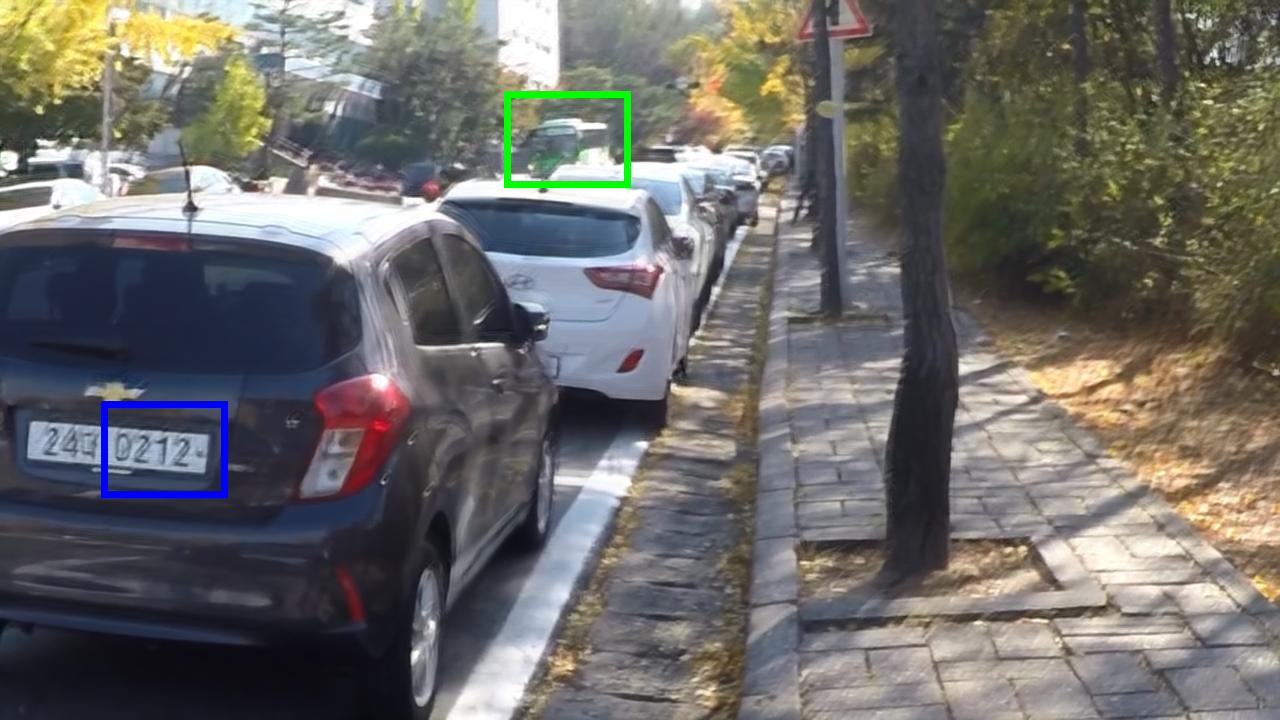}}&
    \multicolumn{2}{c}{\includegraphics[width=0.245\linewidth]{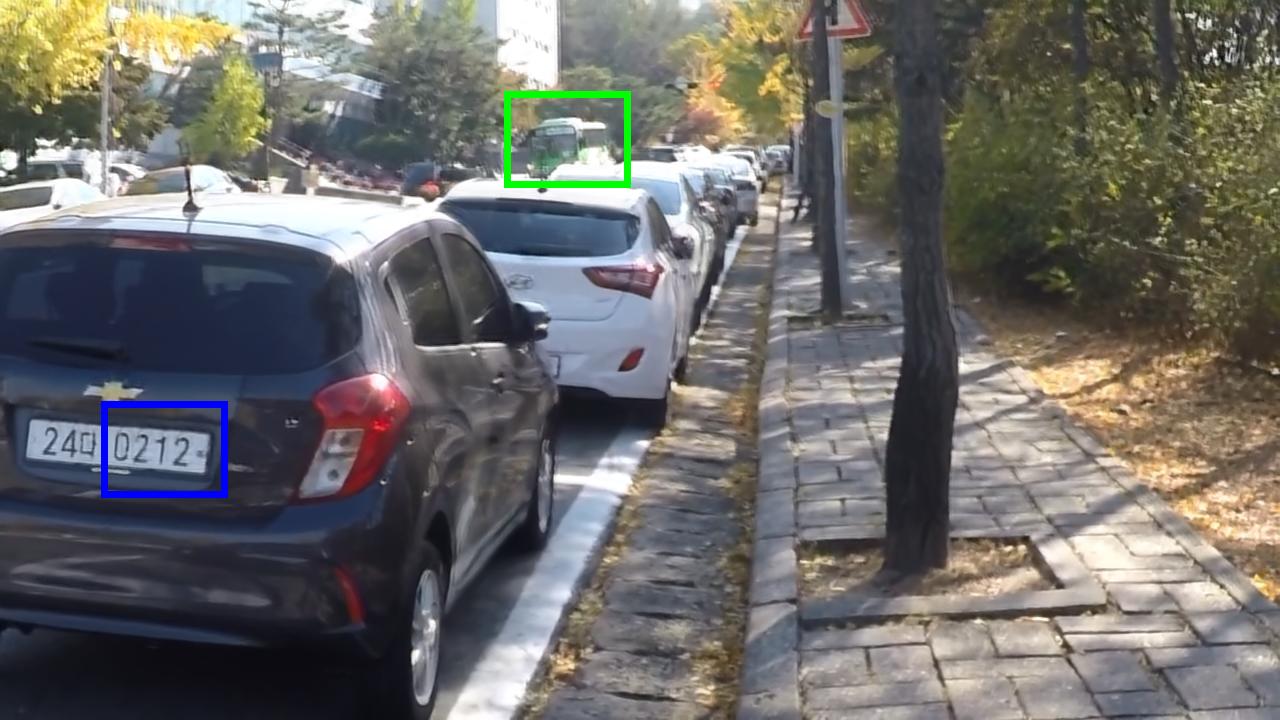}} &
    \multicolumn{2}{c}{\includegraphics[width=0.245\linewidth]{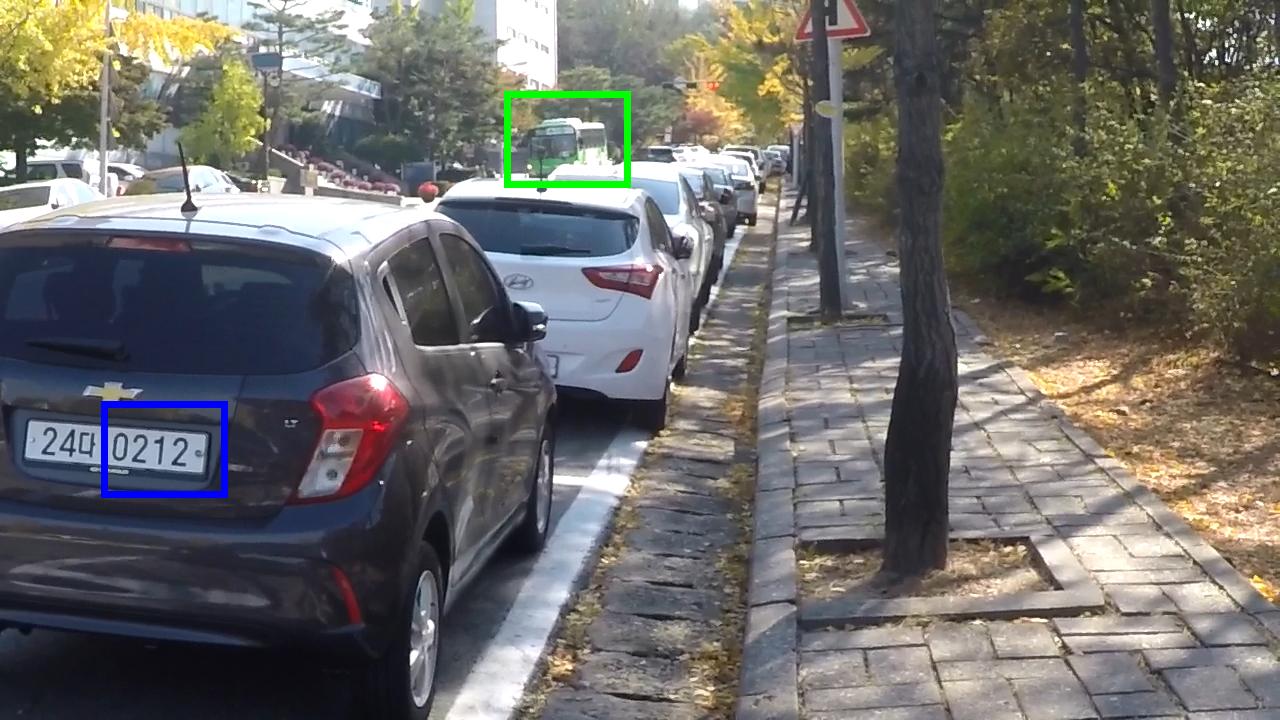}}\\

    \multicolumn{1}{l}{\includegraphics[width=0.12\linewidth]{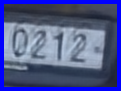}} &
    \multicolumn{1}{r}{\includegraphics[width=0.12\linewidth]{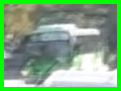}} &
    \multicolumn{1}{l}{\includegraphics[width=0.12\linewidth]{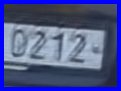}} &
    \multicolumn{1}{r}{\includegraphics[width=0.12\linewidth]{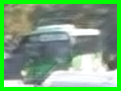}} &
    \multicolumn{1}{l}{\includegraphics[width=0.12\linewidth]{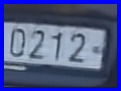}} &
    \multicolumn{1}{r}{\includegraphics[width=0.12\linewidth]{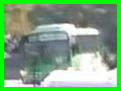}} &
    \multicolumn{1}{l}{\includegraphics[width=0.12\linewidth]{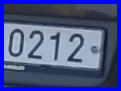}} &
    \multicolumn{1}{r}{\includegraphics[width=0.12\linewidth]{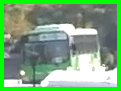}} \\

    \multicolumn{2}{c}{(e) IFI-RNN \cite{Nah2019recurrent}}  & \multicolumn{2}{c}{(f) IFI-RNN-L \cite{Nah2019recurrent}}
     & \multicolumn{2}{c}{(g) Ours} & \multicolumn{2}{c}{(h) GT}\\
  \end{tabular}
  \vspace{-0.2cm}
  \caption{Qualitative comparison on synthetically blurred video frames of Nah \Etal~\shortcite{nah2017deep}'s dataset.}
\label{fig:gopro_nah2}
\end{figure*}

\paragraph{Comparison using Su \Etal~\shortcite{su2017deep}'s dataset}
For fair comparison, we trained all the models except for OVD~\cite{kim2017online} and DVD~\cite{su2017deep} using the dataset of Su \Etal~\shortcite{su2017deep}.
For this training, we used the source code provided by the authors and their own training strategies.
The training code of OVD is not available, so we used its pre-trained model provided by the authors.
We also used the model of DVD provided by the authors; it had already been trained with Su \Etal's dataset.
For quantitative evaluation, we excluded the first frames of test videos, because most video deblurring methods cannot successfully handle them.

In the quantitative comparison (\Tbl{result_table}), 
our model achieved higher PSNR and SSIM than all the other methods, including single image and video deblurring methods.
Compared to both DVD-noMC and DVD-MC, even Ours-small performed much better despite its significantly smaller size, because DVD cannot utilize previous deblurred frames due to the lack of a recurrent structure.
Compared to OVD, Ours-small achieved PSNRs more than 1 dB higher.
Ours-small has more parameters than OVD, but most are used for motion estimation; the deblurring network, \emph{PVDNet}, has fewer parameters than OVD.
This comparison clearly shows the benefit of our motion compensation scheme.
Similarly, compared to IFI-RNN,
Ours-small as well as Ours outperformed it both in PSNR and SSIM, in spite of the smaller deblurring network.
Compared to IFI-RNN-L and EDVR, which have larger numbers of parameters, Ours achieved higher PSNR and SSIM even though the total number of parameters is smaller.
Finally, Ours trained with CA outperformed EDVR by a margin of 0.49 dB, proving the effectiveness of our framework.

The advantage of our approach is emphasized in the qualitative comparison (\Fig{gopro}).
As the input video frame (\Fig{gopro}a) is severely blurred, the single image deblurring method of Tao \Etal~\shortcite{tao2018scale} fails to recover fine-scale structures and produces blurry results.
DVD-MC recovers fine-scale structures better, but produces distorted structures due to motion compensation errors.
IFI-RNN, IFI-RNN-L, and EDVR also fail to restore fine-scale structures due to large motions in the input frames.
In contrast, thanks to our motion compensation scheme, our method effectively restores fine details without distortions even in the presence of large inter-frame motion and blur.

\paragraph{Comparison using Nah \Etal~\shortcite{nah2017deep}'s dataset}
For fair comparison, all the methods except for OVD~\cite{kim2017online} were trained from scratch with the training set of Nah \Etal~\shortcite{nah2017deep}.
In the quantitative comparison (\Tbl{result_table2}), our method using cosine annealing (Ours w/ CA) achieved better or comparable performance to the other state-of-the-art methods including IFI-RNN-L~\cite{Nah2019recurrent} and EDVR~\cite{Wang2019edvr} despite its smaller model size.
In this comparison, we also included our method with a similar model size to that of EDVR, to show the effectiveness of our explicit motion compensation approach.
Specifically, to increase the model size to match that of EDVR, we used a larger \emph{PVDNet} that has 24 residual blocks with 192 channels.
Our larger model (Ours-large w/ CA) clearly outperforms EDVR both in PSNR and SSIM.

In a qualitative comparison, 
the input blurred frame (\Fig{gopro_nah2}a) has severe blur caused by large camera motion.
Consequently, the results of all the other methods have remaining blur.
On the other hand, our result (\Fig{gopro_nah2}g) shows clearly restored details with no remaining blur.

\begin{table}[t]
\centering
\caption{Quantitative comparison of our video deblurring framework with recent state-of-the-art methods on the dataset of \cite{nah2017deep}.
}
\vspace{-8pt}
\begin{tabular}{ |c || c c c | }
\hline
 & PSNR \small{(dB)} & SSIM & Parameters \small{(M)}\\
\hline \hline
\cite{nah2017deep} & 29.47 & 0.876 & 75.7 \\ 
\cite{tao2018scale} &  30.61 & 0.908 & 3.8 \\ 
\hline
OVD \small{\cite{kim2017online}}  & 26.57 & 0.820 & 0.9\\ 
IFI-RNN \small{\cite{Nah2019recurrent}} & 30.15 & 0.895 & 1.6 \\ 
IFI-RNN-L \small{\cite{Nah2019recurrent}} & 31.05 & 0.911 & 12.2 \\ 
EDVR \small{\cite{Wang2019edvr}}& 31.54 & 0.926 & 23.6 \\ 
\hline
Ours     & 31.30 & 0.917 & 10.5 \small{(5.1 + 5.4)} \\ 
Ours w/ CA & 31.52 & 0.921 & 10.5 \small{(5.1 + 5.4)}\\ 
Ours-large w/ CA & \bf{31.98} & \bf{0.928} & 23.4 \small{(18.0 + 5.4)}\\ 
\hline
\end{tabular}
  \vspace{-0.3cm}
\label{tbl:result_table2}
\end{table}

\begin{figure*}[tp]
\centering
\setlength\tabcolsep{1 pt}
  \begin{tabular}{cccccccc}
    \multicolumn{2}{c}{\includegraphics[width=0.24\linewidth]{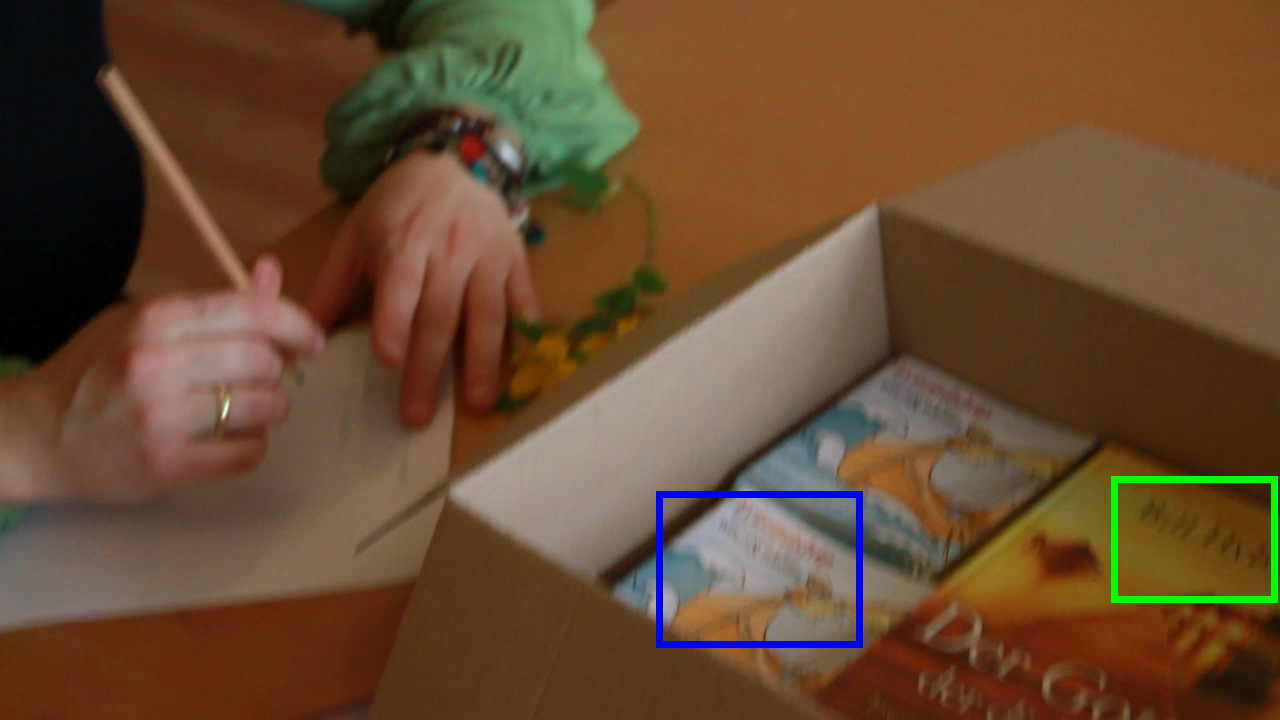}} &
    \multicolumn{2}{c}{\includegraphics[width=0.24\linewidth]{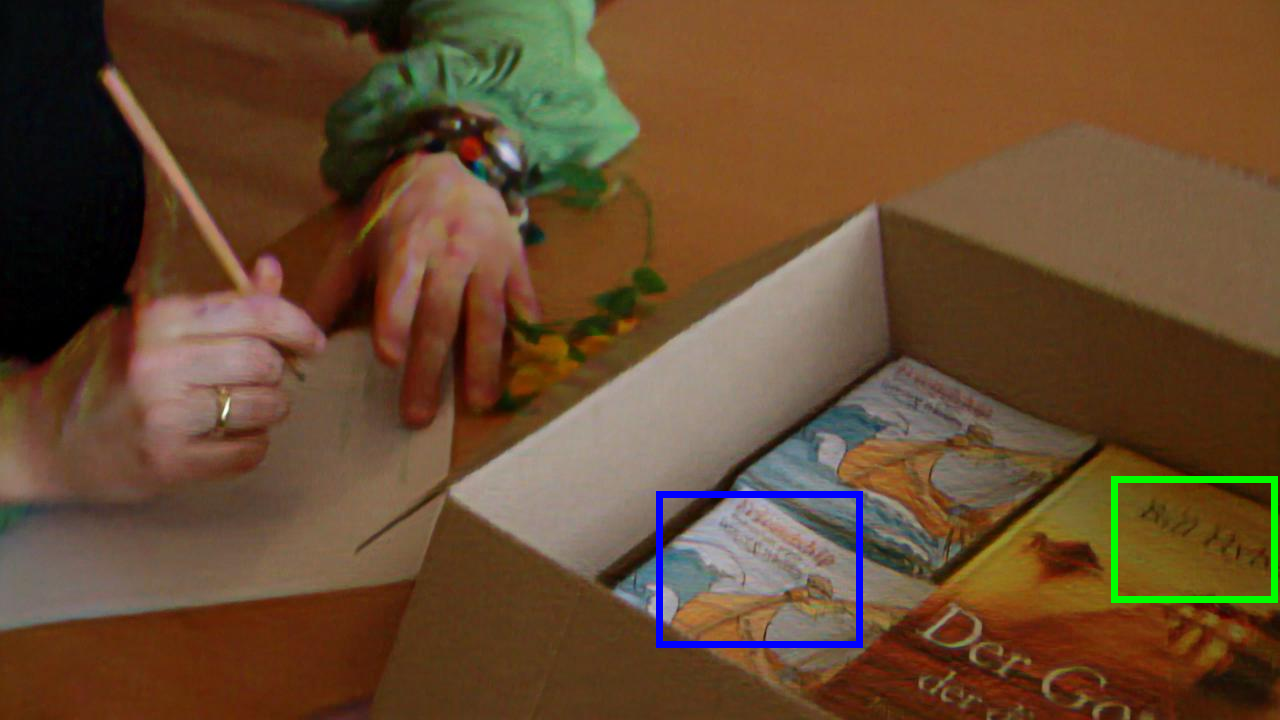}}&
    \multicolumn{2}{c}{\includegraphics[width=0.24\linewidth]{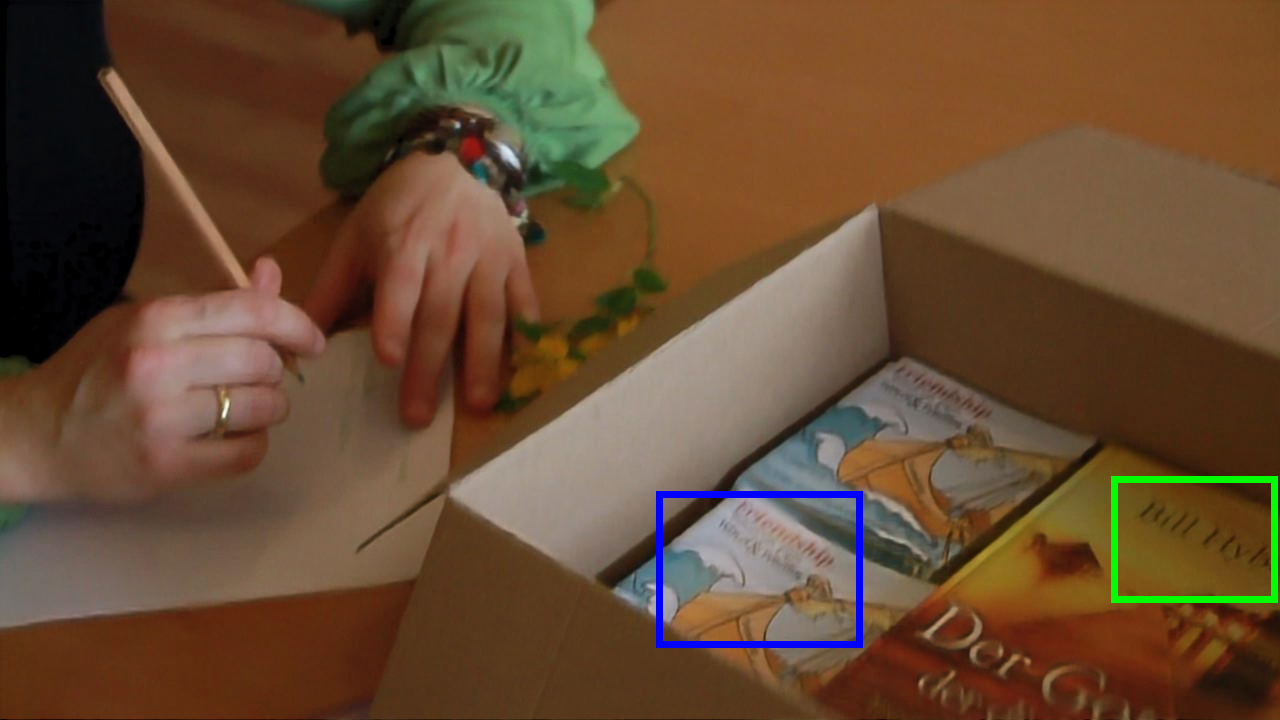}}&
    \multicolumn{2}{c}{\includegraphics[width=0.24\linewidth]{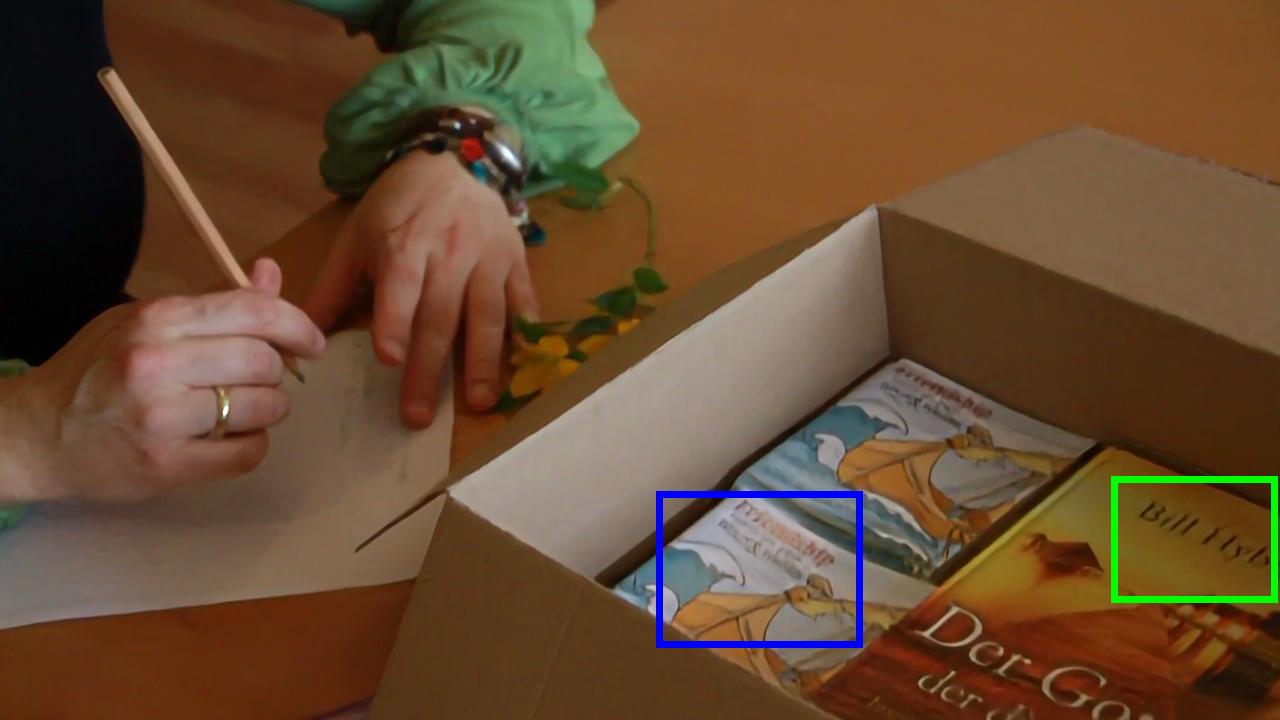}} \\

    \multicolumn{1}{c}{\includegraphics[width=0.118\linewidth]{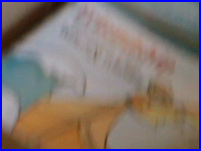}} &
    \multicolumn{1}{c}{\includegraphics[width=0.118\linewidth]{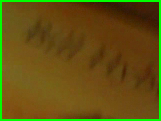}} &
    \multicolumn{1}{c}{\includegraphics[width=0.118\linewidth]{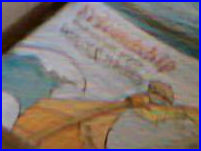}} &
    \multicolumn{1}{c}{\includegraphics[width=0.118\linewidth]{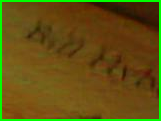}} &
    \multicolumn{1}{c}{\includegraphics[width=0.118\linewidth]{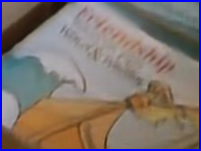}} &
    \multicolumn{1}{c}{\includegraphics[width=0.118\linewidth]{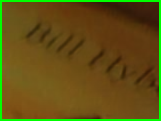}} &
    \multicolumn{1}{c}{\includegraphics[width=0.118\linewidth]{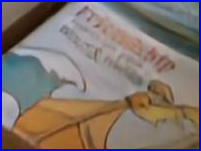}} &
    \multicolumn{1}{c}{\includegraphics[width=0.118\linewidth]{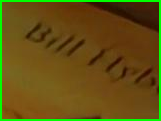}} \\

    \multicolumn{2}{c}{(a) Input}  & \multicolumn{2}{c}{(b) \cite{tao2018scale}}
     & \multicolumn{2}{c}{(c) DVD-MC \cite{su2017deep}} & \multicolumn{2}{c}{(d) EDVR \cite{Wang2019edvr}}\\
     \multicolumn{2}{c}{}  & \multicolumn{2}{c}{3.8M params}
     & \multicolumn{2}{c}{15.3M params} & \multicolumn{2}{c}{23.6M params}\\

    \multicolumn{2}{c}{\includegraphics[width=0.24\linewidth]{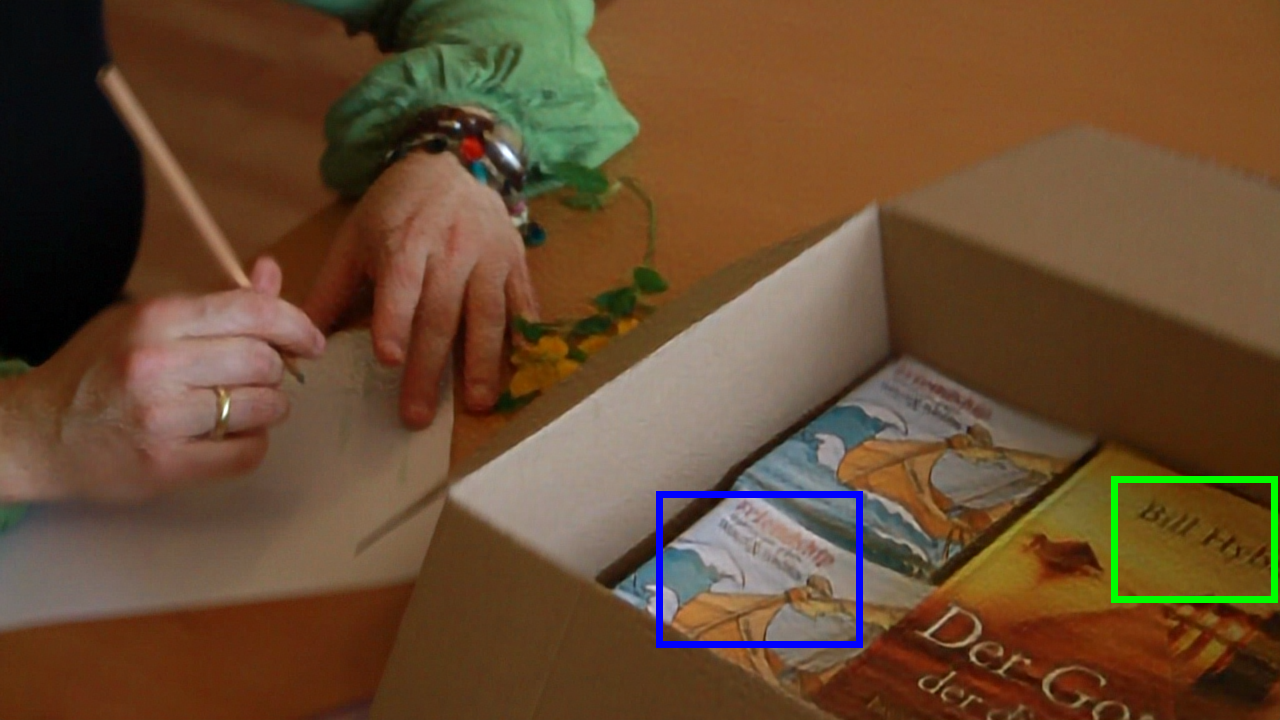}}&
    \multicolumn{2}{c}{\includegraphics[width=0.24\linewidth]{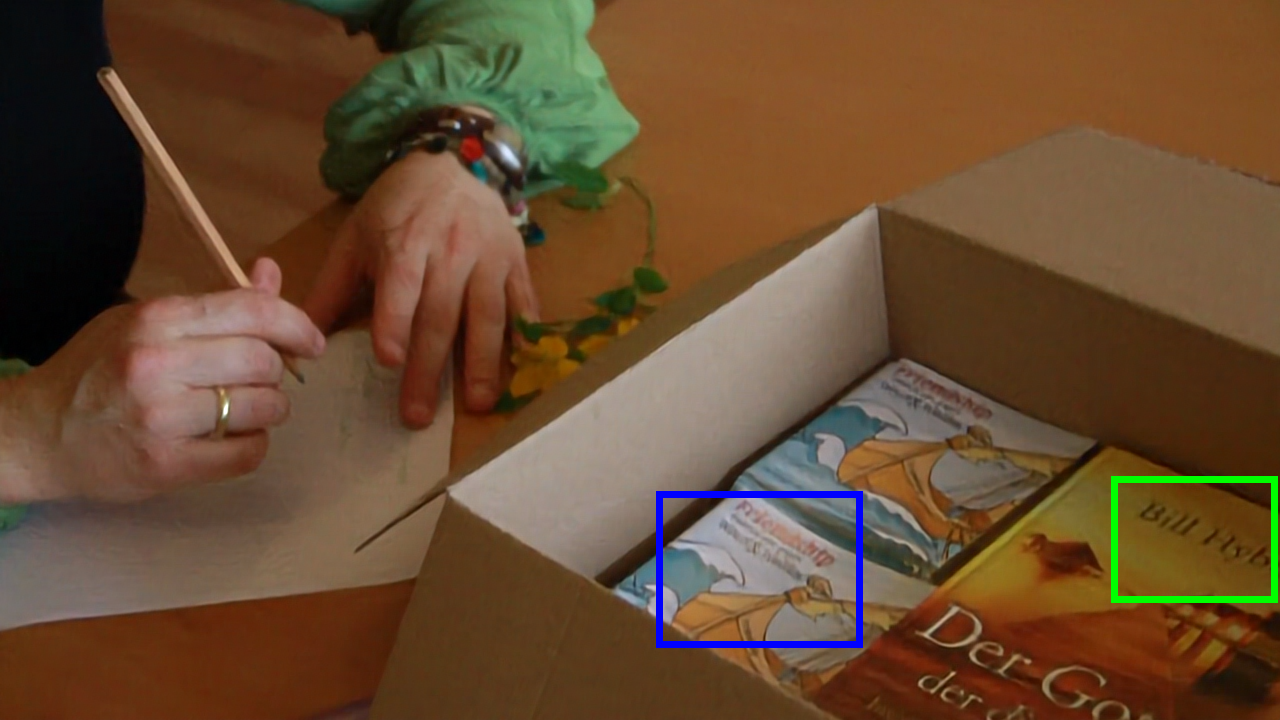}}&
    \multicolumn{2}{c}{\includegraphics[width=0.24\linewidth]{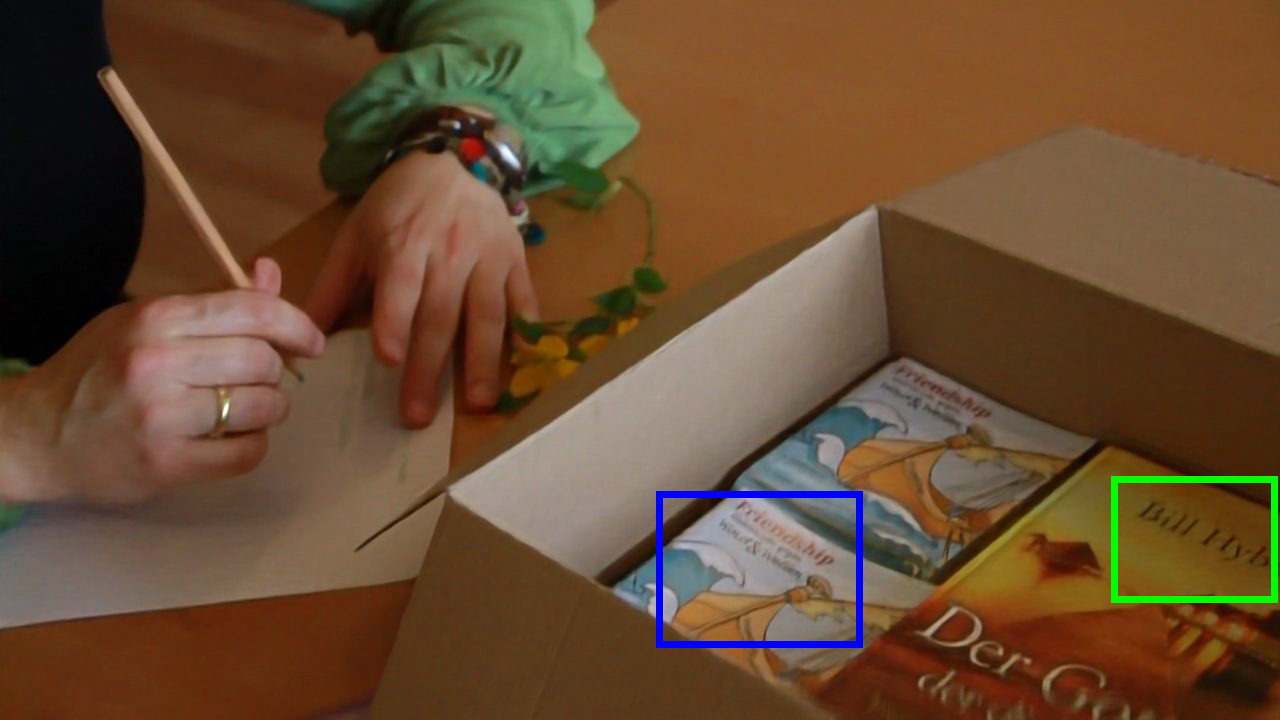}} &
    \multicolumn{2}{c}{\includegraphics[width=0.24\linewidth]{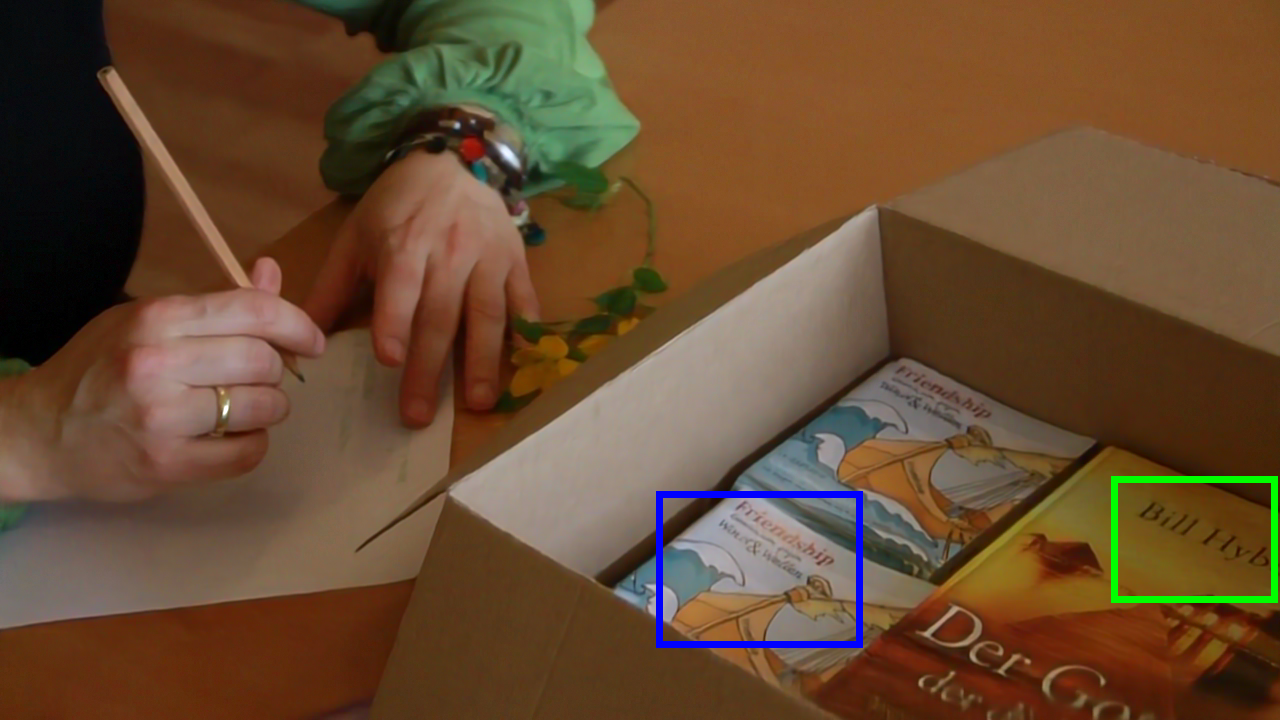}}\\

    \multicolumn{1}{c}{\includegraphics[width=0.118\linewidth]{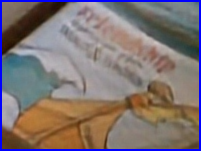}} &
    \multicolumn{1}{c}{\includegraphics[width=0.118\linewidth]{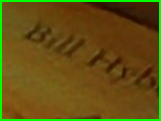}} &
    \multicolumn{1}{c}{\includegraphics[width=0.118\linewidth]{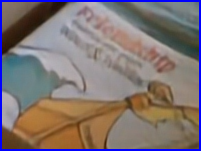}} &
    \multicolumn{1}{c}{\includegraphics[width=0.118\linewidth]{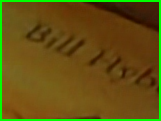}} &
    \multicolumn{1}{c}{\includegraphics[width=0.118\linewidth]{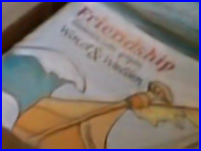}} &
    \multicolumn{1}{c}{\includegraphics[width=0.118\linewidth]{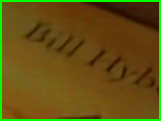}} &
    \multicolumn{1}{c}{\includegraphics[width=0.118\linewidth]{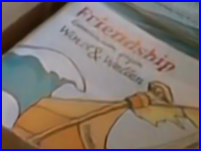}} &
    \multicolumn{1}{c}{\includegraphics[width=0.118\linewidth]{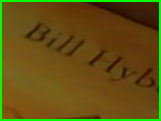}} \\

    \multicolumn{2}{c}{(e) IFI-RNN \cite{Nah2019recurrent}}  & \multicolumn{2}{c}{(f) IFI-RNN-L \cite{Nah2019recurrent}}
     & \multicolumn{2}{c}{(g) Ours-small} & \multicolumn{2}{c}{(h) Ours}\\
     \multicolumn{2}{c}{1.6M params} & \multicolumn{2}{c}{12.2M params} & \multicolumn{2}{c}{6.1M (0.7M + 5.4M) params} & \multicolumn{2}{c}{10.5M (5.1M + 5.4M) params}
  \end{tabular}
  \vspace{-0.3cm}
  \caption{Qualitative comparison on real-world video frames of Su \Etal~\shortcite{su2017deep}'s test set.}
  \vspace{-0.2cm}
\label{fig:real}
\end{figure*}

\paragraph{Comparison using real-world videos}
Finally, we qualitatively compare the performance of our approach to those of the state-of-the-art methods on real-world data (\Fig{real}).
The datasets~\cite{su2017deep,nah2017deep} that were used for training and evaluation are synthetically generated using high-speed cameras, so it may not fully represent the characteristics of real blurry videos.
Thus, evaluation using real-world data is essential for predicting the generalization ability of a method.
For evaluation, we used the real blurry video test set in Su \Etal~\shortcite{su2017deep}.

In this comparison, 
the input video frame (\Fig{real}a) is severely blurred by camera motions, so the characters in the frame are illegible.
The single image deblurring method of Tao \Etal~\shortcite{tao2018scale} (\Fig{real}b) failed to restore the characters properly and introduced noisy artifacts.
The results of the video deblurring methods, DVD-MC (\Fig{real}c) and IFI-RNN (\Fig{real}e) have remaining blurs.
Compared to the results of the other methods, the results of EDVR (\Fig{real}d) and IFI-RNN-L (\Fig{real}f) show relatively better restored details because of their large model sizes, but small characters remain illegible either.
In contrast, both Ours-small (\Fig{real}g) and Ours (\Fig{real}h) more clearly restore characters with less artifacts despite their smaller models than EDVR and IFI-RNN-L.

\begin{figure*}[tp]
\centering
\setlength\tabcolsep{1 pt}
  \begin{tabular}{cccccccc}
    \multicolumn{2}{c}{\includegraphics[width=0.24\linewidth]{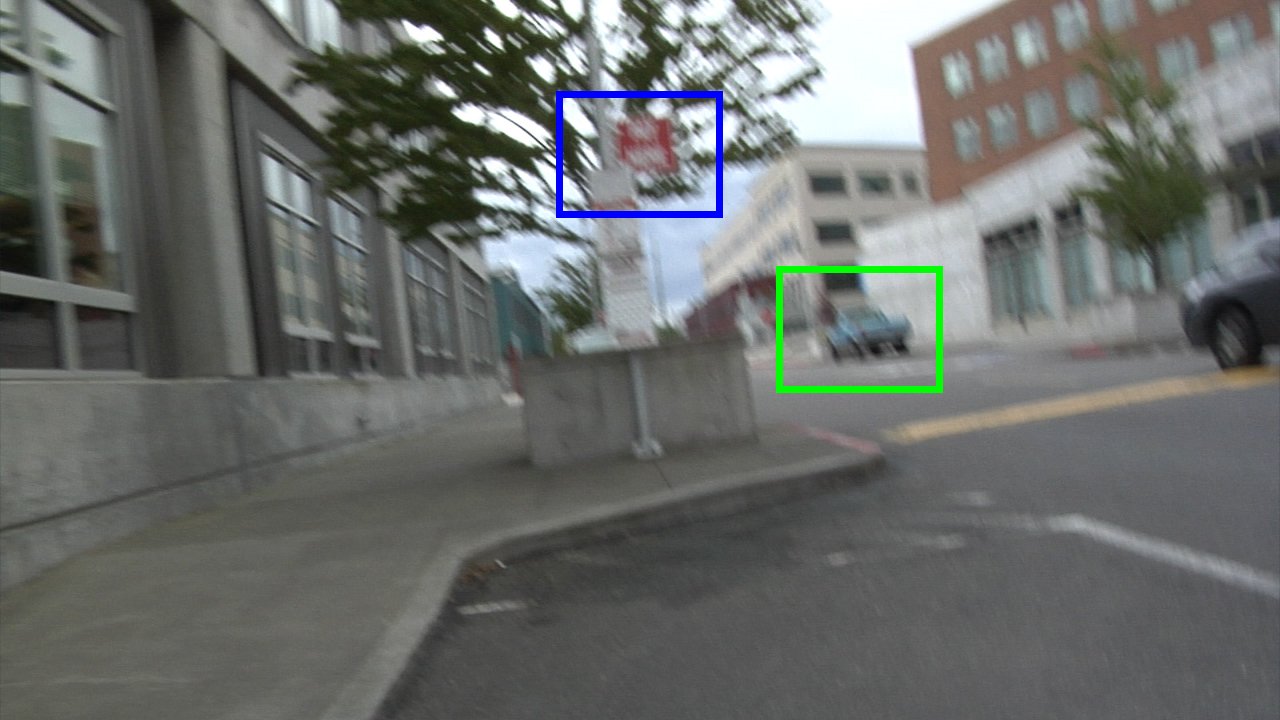}} &
    \multicolumn{2}{c}{\includegraphics[width=0.24\linewidth]{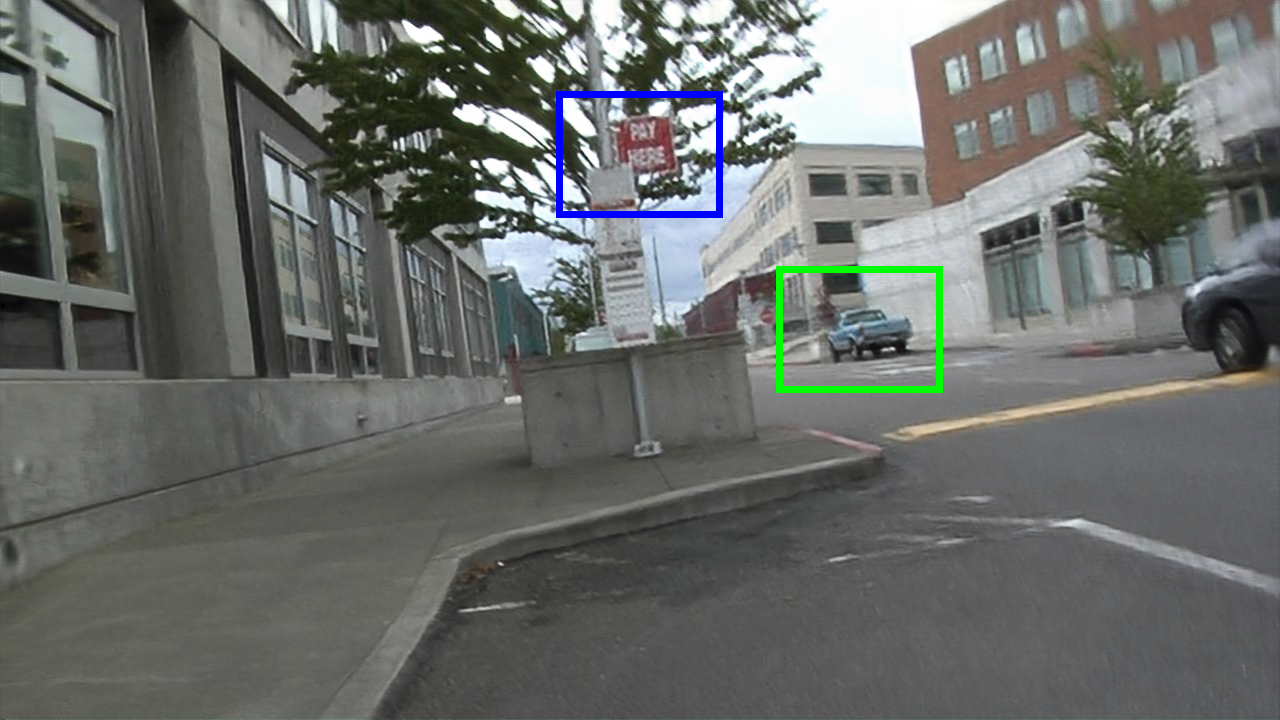}}&
    \multicolumn{2}{c}{\includegraphics[width=0.24\linewidth]{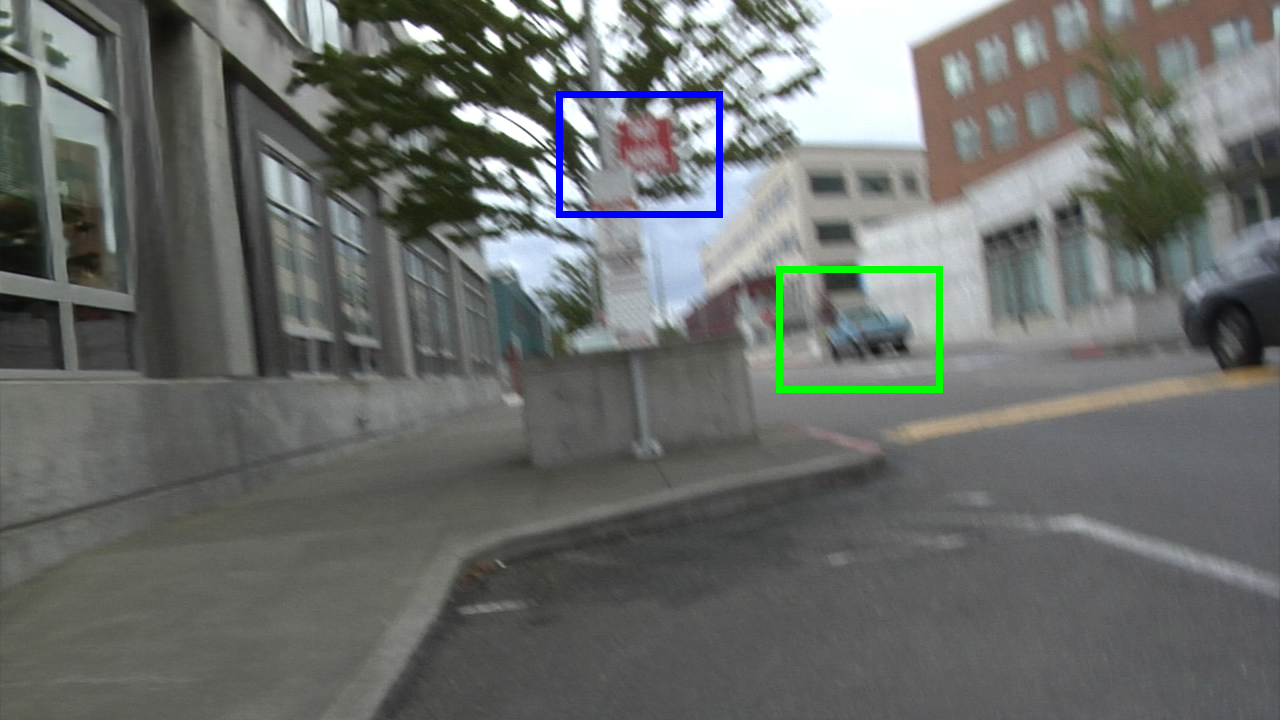}}&
    \multicolumn{2}{c}{\includegraphics[width=0.24\linewidth]{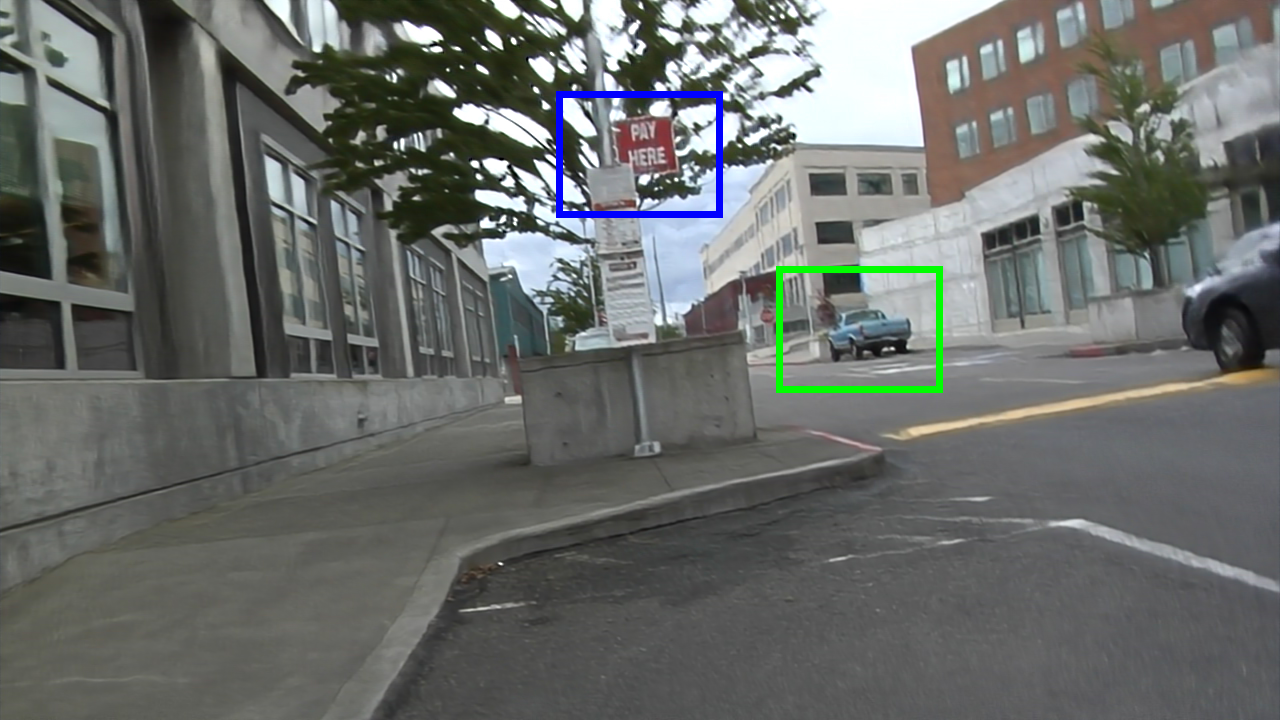}}\\

    \multicolumn{1}{c}{\includegraphics[width=0.118\linewidth]{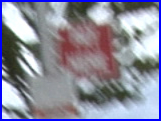}} &
    \multicolumn{1}{c}{\includegraphics[width=0.118\linewidth]{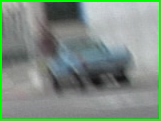}} &
    \multicolumn{1}{c}{\includegraphics[width=0.118\linewidth]{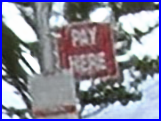}} &
    \multicolumn{1}{c}{\includegraphics[width=0.118\linewidth]{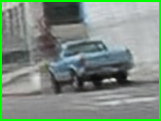}} &
    \multicolumn{1}{c}{\includegraphics[width=0.118\linewidth]{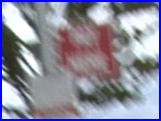}} &
    \multicolumn{1}{c}{\includegraphics[width=0.118\linewidth]{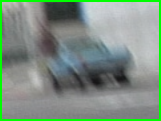}} &
    \multicolumn{1}{c}{\includegraphics[width=0.118\linewidth]{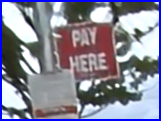}} &
    \multicolumn{1}{c}{\includegraphics[width=0.118\linewidth]{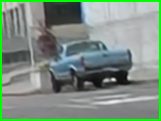}} \\

    \multicolumn{2}{c}{\includegraphics[width=0.24\linewidth]{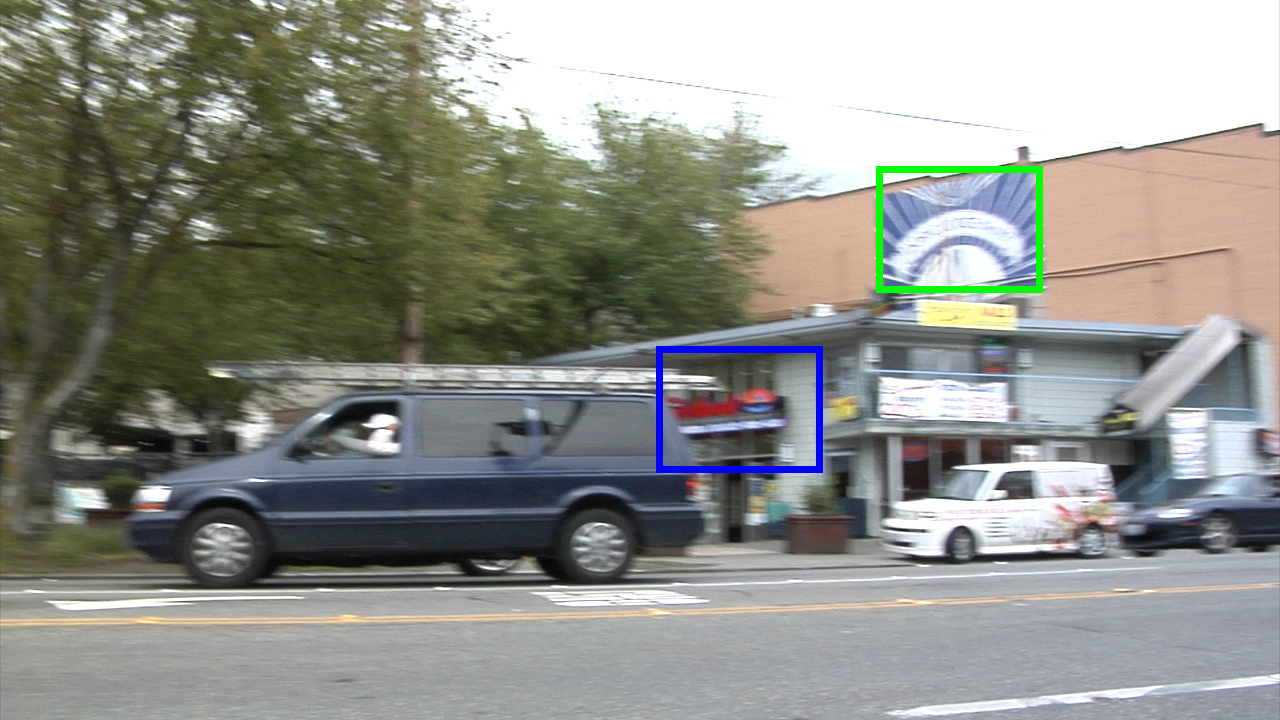}} &
    \multicolumn{2}{c}{\includegraphics[width=0.24\linewidth]{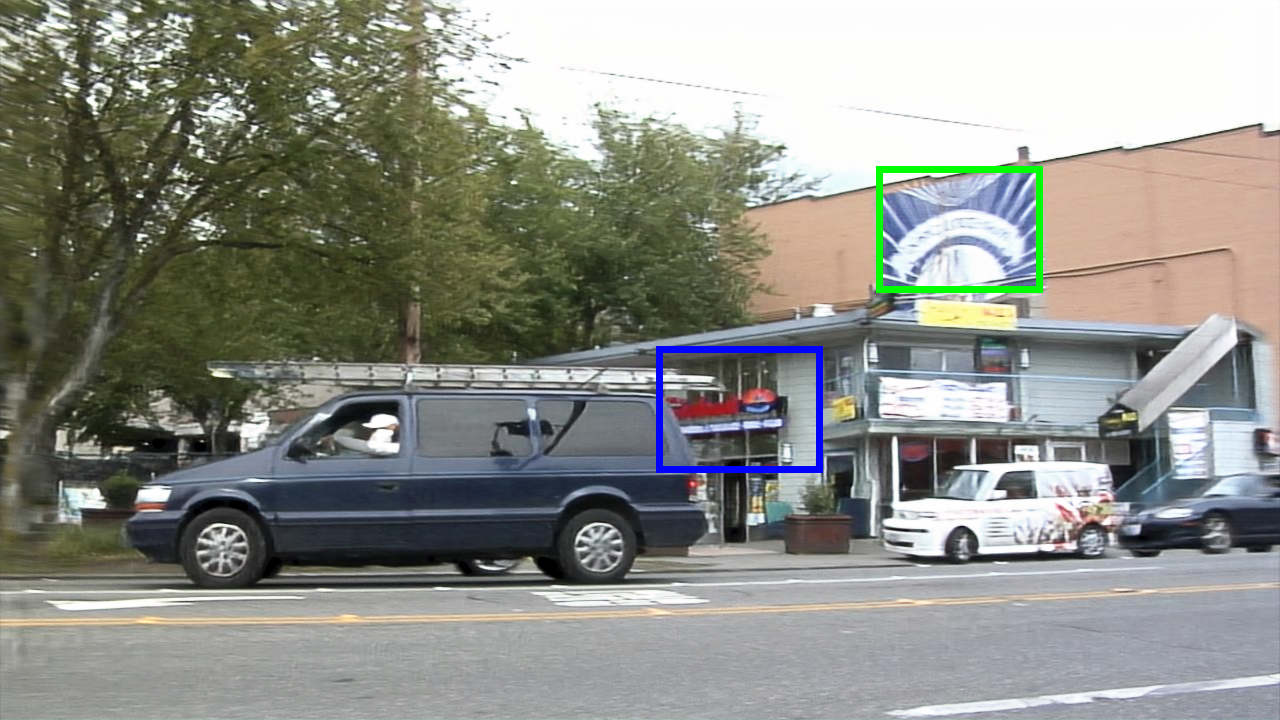}}&
    \multicolumn{2}{c}{\includegraphics[width=0.24\linewidth]{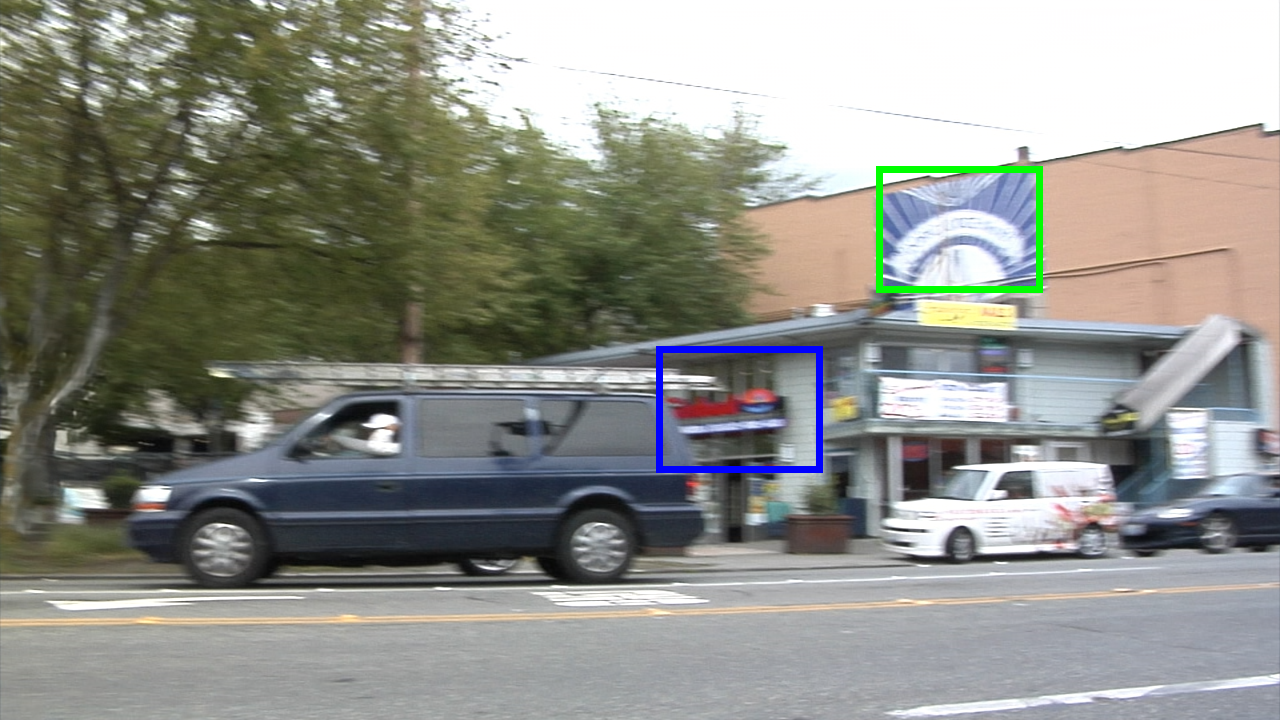}}&
    \multicolumn{2}{c}{\includegraphics[width=0.24\linewidth]{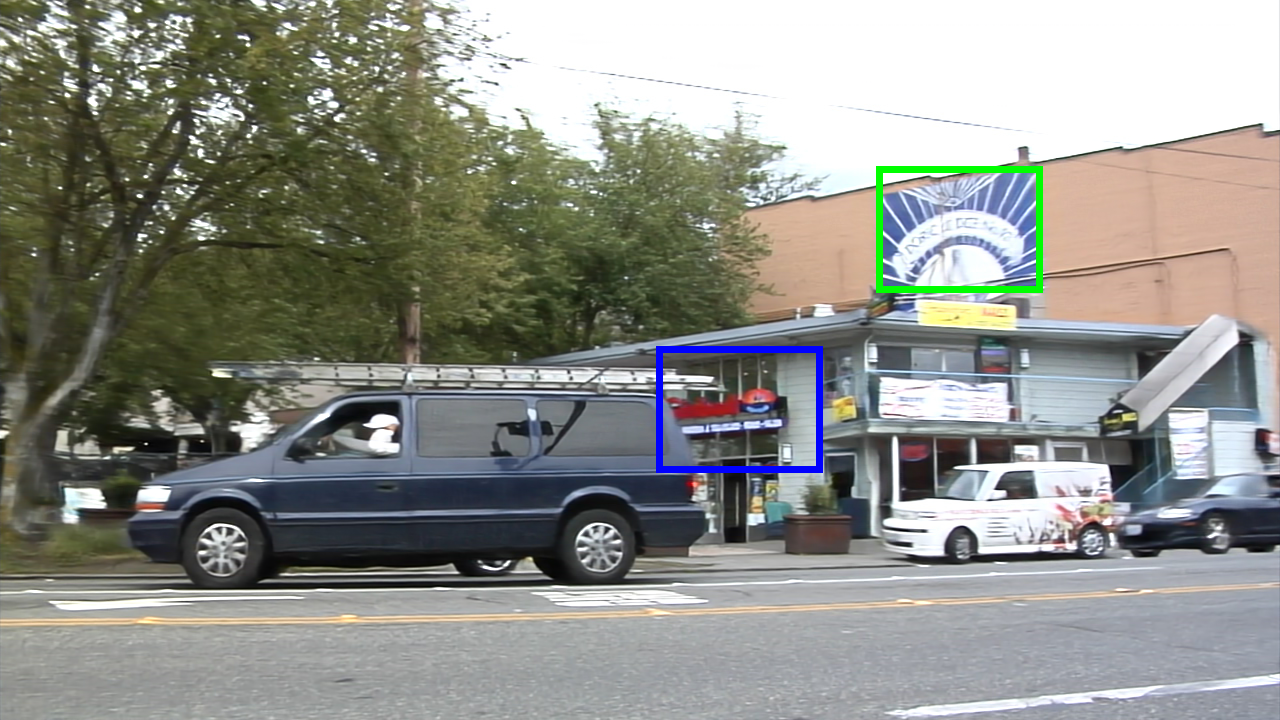}}\\

    \multicolumn{1}{c}{\includegraphics[width=0.118\linewidth]{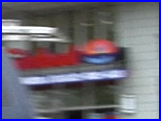}} &
    \multicolumn{1}{c}{\includegraphics[width=0.118\linewidth]{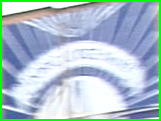}} &
    \multicolumn{1}{c}{\includegraphics[width=0.118\linewidth]{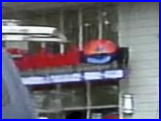}} &
    \multicolumn{1}{c}{\includegraphics[width=0.118\linewidth]{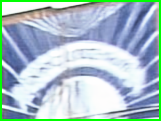}} &
    \multicolumn{1}{c}{\includegraphics[width=0.118\linewidth]{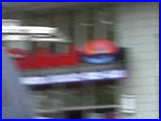}} &
    \multicolumn{1}{c}{\includegraphics[width=0.118\linewidth]{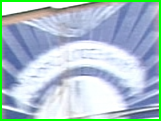}} &
    \multicolumn{1}{c}{\includegraphics[width=0.118\linewidth]{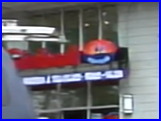}} &
    \multicolumn{1}{c}{\includegraphics[width=0.118\linewidth]{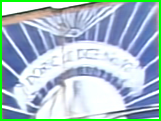}} \\

    \multicolumn{2}{c}{(a) input}  & \multicolumn{2}{c}{(b) DVD-MC \cite{su2017deep}}
     & \multicolumn{2}{c}{(c) IFI-RNN-L \cite{Nah2019recurrent}} & \multicolumn{2}{c}{(d) Ours}\\
  \end{tabular}
  \vspace{-0.3cm}
  \caption{Additional examples on real-world video frames of Su \Etal~\shortcite{su2017deep}'s test set.}
  \vspace{-0.3cm}
\label{fig:real2}
\end{figure*}

In another qualitative comparison (\Fig{real2}), 
the input video frames suffer from large blur.
DVD-MC restored small-scale structures relatively well, but its results still have remaining blur and noisy artifacts.
IFI-RNN-L completely failed to restore sharp frames, possibly due to large motions between input frames that are difficult to handle without explicit motion compensation.
On the contrary, our results clearly restored sharp details.
Additional examples can be found in the supplementary material.

\subsection{Computational cost analysis}
\label{ssec:cost}

To process a single 1280$\times$720 video frame, our framework takes about 0.11s with an NVIDIA Titan Xp GPU, and our framework can process images of arbitrary size, because both \emph{BIMNet} and \emph{PVDNet} consist of convolutional layers.
Specifically, \emph{BIMNet}, \emph{PVDNet}, and pixel volume construction take 0.04s, 0.04s, and 0.03s per frame, respectively.
The pixel volume construction step takes relatively long computation time. However, its code is un-optimized PyTorch code that uses for-loops, which are known to be slow in PyTorch. We believe that the code can be further optimized using CUDA programming.
The computational overhead caused by a pixel volume in \emph{PVDNet} is smaller than that of a single convolution layer in our framework.
In the aspect of memory usage, the size of a pixel volume is $W \times H \times k^2$, and $k$ is set to 5 in our experiments.
In contrast, the first convolution layer of \emph{PVDNet} uses 64 channels so its size is $W \times H \times 64$. Thus, a PV requires only $\sim$39\% memory usage of the first convolution layer.
In the aspect of computation time, once a pixel volume is constructed, it is processed by convolution operations like other layers. Thus, the overhead computation time caused by a pixel volume is also smaller than that of a single convolution layer in our framework. 

We compared the computational costs of our method and others. 
We used the number of Multiply-accumulate (MAC) operations required to process a single 1280$\times$720 video frame as the measurement.
Our method (936 BMACs) has a much smaller computational cost than IFI-RNN-L (1425 BMACs) and EDVR (2739 BMACs) despite achieving higher deblurring accuracy (\Tbl{result_table}).
\section{Conclusion}

In this paper, we presented two novel approaches for motion compensation to effectively deliver previously deblurred information.
We proposed simple but effective blur-invariant learning for accurate estimation of optical flow between blurred video frames.
In addition, we constructed a pixel volume that contains matching candidates for robustly handling motion estimation errors, and fed it to our deblurring network instead of directly feeding a motion compensated warped frame.
Finally, we proposed an effective video deblurring framework based on the proposed motion compensation approaches.

Our framework achieved the state-of-the-art deblurring performance,
as demonstrated by extensive experiments on various challenging videos with large blur. 
Combination of blur-invariant motion estimation learning and a pixel volume can improve motion compensation without significant computational overhead or changes in network design. We expect that these two modules would be applicable to other video processing tasks, such as video super-resolution that requires sub-pixel-level motion compensation.

Similarly to previous deblurring methods~\cite{nah2017deep,tao2018scale,su2017deep,kim2017online,kim2018spatio,Wang2019edvr}, in the synthetic datasets \cite{su2017deep,nah2017deep}, we use temporal center frames as ground truth sharp frames for training both \emph{BIMNet} and \emph{PVDNet}.
However, complex blur such as circular blur may introduce ambiguity, because more than one camera motions with different temporal centers can produce the same shape of blur; this ambiguity may confuse the networks and degrade their performances.
This problem would not be severe in practice given the success of recent deblurring methods that use temporal center frames for ground truths, 
but analyzing its effect and resolving it can be a good future direction.

\section*{Acknowledgements}
We would like to thank the anonymous reviewers for their constructive comments.
This work was supported by the Ministry of Science and ICT, Korea, through
IITP grants (SW Star Lab, IITP-2015-0-00174; Artificial Intelligence Graduate School Program (POSTECH), IITP-2019-0-01906) and
NRF grants (NRF-2018R1A5A1060031; NRF-2020R1C1C1014863).

\appendix
\section*{Appendix}
\section{\emph{BIMNet}}

\subsection{Effect of Siamese structure on blur-invariant feature learning}
\label{ssec:siamese_structure}
As described in \SSec{BIMNet}, we adopt \emph{LiteFlowNet}~\cite{LiteFlowNet} for our \emph{BIMNet} as \emph{LiteFlowNet} has a Siamese network architecture that is effective to learn blur-invariant features.
In this section, we conduct an experiment to verify the effectiveness of the Siamese architecture on blur-invariant feature learning.

To this end, we compare two different networks for optical flow estimation: \emph{FlowNetS} and \emph{FlowNetC}~\cite{FlowNet}.
\emph{FlowNetS} and \emph{FlowNetC} share almost the same network architecture except that \emph{FlowNetC} has a Siamese network at the beginning of the network, while \emph{FlowNetS} has a plain CNN architecture.
Using the Siamese network architecture, \emph{FlowNetC} first extracts a feature for each pixel and then match the extracted features to estimate optical flow.
While both \emph{FlowNetS} and \emph{FlowNetC} can run as stand-alone networks for optical flow estimation, \emph{LiteFlowNet} combines them together for obtaining high-quality optical flow.

To investigate the effect of the Siamese network architecture,
we evaluated the performance of \emph{FlowNetS} and \emph{FlowNetC} on blur-invariant feature learning.
Specifically, we fine-tuned both pretrained networks with our {\em blur-invariant loss} $L_{BIM}^{\alpha\beta}$.
The networks were originally pre-trained with datasets of \cite{Geiger2012CVPR:KITTI,Scharstein:2014:Middlebury,Butler:ECCV:2012:sintel}
and we used the dataset of Su~\Etal~\shortcite{su2017deep} for fine-tuning them.
\Tbl{bi_table_sc} compares \emph{FlowNetS} and \emph{FlowNetC} with their fine-tuned versions.
While $L_{BIM}^{\alpha\beta}$ improves the accuracy of both networks for blurry videos,
the improvement of \emph{FlowNetC} is much larger than that of \emph{FlowNetS},
showing that the Siamese network architecture of \emph{FlowNetC} learns to extract blur-invariant features more effectively than a plain CNN architecture.

\begin{table}[t]
\centering
\caption{Accuracy gain of a Siamese structure in blur-invariant feature learning. The networks with subscript $L_{BIM}^{\alpha\beta}$ are fine-tuned ones with blur-invariant loss $L_{BIM}^{\alpha\beta}$ on the training set of Su~\Etal~\shortcite{su2017deep}. We computed the errors between $I_{t}^s$ and $Warp(I_{t-1}^s, W^{bb})$ using PSNR and SSIM for each network, where $W^{bb}$ is the optical flow estimated from $I_{t}^b$ to $I_{t-1}^b$. Errors were measured on the test set of Su~\Etal~\shortcite{su2017deep}. Accuracy gain is higher for \emph{FlowNetC}$_{L_{BIM}^{\alpha\beta}}$ than \emph{FlowNetS}$_{L_{BIM}^{\alpha\beta}}$.}
\vspace{-6pt}
\begin{tabular}{ |c||cc| }
\hline
Model & PSNR (dB) & SSIM\\

\hline \hline
\emph{FlowNetS} &25.8619&0.8653\\
\emph{FlowNetS}$_{L_{BIM}^{\alpha\beta}}$ &26.7387&0.8792\\
\hline
accuracy gain &3\%&1\%\\
\hline \hline
\emph{FlowNetC} & 26.0403 & 0.8659 \\
\emph{FlowNetC}$_{L_{BIM}^{\alpha\beta}}$ & 27.7182 & 0.9179 \\
\hline
accuracy gain &6\%&6\%\\
\hline
\end{tabular}
\vspace{-10pt}
\label{tbl:bi_table_sc}
\end{table}

\begin{table*}[t]
\centering
\caption{Quantitative comparison between \emph{BIMNet} and other optical flow estimation methods on different types of image pairs $(I_{t-1}^\alpha,I_{t}^\beta)$. We computed average errors between $I_{t}^s$ and $Warp(I_{t-1}^s, W^{\alpha\beta})$ in PSNR and SSIM, where $W^{\alpha\beta}$ is the optical flow estimated from $I_{t}^\alpha$ to $I_{t-1}^\beta$. \emph{BS}, \emph{SB}, and \emph{SS} denote $(\alpha,\beta)$: $(b, s)$, $(s, b)$, and $(s, s)$, respectively.}
\vspace{-8pt}
\begin{tabular}{ | c || c c | c c | c c | }
\hline
 \multirow{2}{*}{Model} & \multicolumn{2}{|c|}{\emph{BS}} & \multicolumn{2}{|c|}{\emph{SB}} & \multicolumn{2}{|c|}{\emph{SS}}\\
 & PSNR (dB) & SSIM & PSNR (dB) & SSIM & PSNR (dB) & SSIM\\
\hline \hline
\emph{TVL1}~\cite{perez2013}           & 27.34          & 0.869          & 28.08          & 0.875          & 30.78          & 0.904\\
\emph{FlowNet2}~\cite{ilg17flow}       & 28.22          & 0.903          & 28.04          & 0.898          & 30.03          & 0.928\\
\emph{LiteFlowNet}~\cite{LiteFlowNet}  & 27.47          & 0.882          & 26.51          & 0.858          & 29.49          & 0.923\\
\emph{FlowNet2$_{SS}$}                 & 27.41          & 0.869          & 27.76          & 0.864          & \textbf{30.95} & 0.910\\
\emph{LiteFlowNet$_{SS}$}              & 27.90          & 0.885          & 27.57          & 0.874          & 30.64          & 0.927\\
\emph{FlowNet2$_{**}$}                 & 27.86          & 0.824          & 27.33          & 0.837          & 30.64          & 0.868\\
\emph{LiteFlowNet$_{**}$}              & 27.02          & 0.811          & 26.83          & 0.789          & 29.76          & 0.887\\
\emph{BIMNet$_{FN2}$}                  & \textbf{28.98} & \textbf{0.909} & \textbf{29.29} & \textbf{0.910} & 30.68          & \textbf{0.928}\\
\emph{BIMNet$_{LFN}$}                  & 28.85          & 0.901          & 28.86          & 0.901          & 30.64          & \textbf{0.928}\\

\hline
\end{tabular}
\label{tbl:additional_mc_table}
\end{table*}

\subsection{Performance of \emph{BIMNet} on different types of image pairs $(I_{t-1}^\alpha,I_{t}^\beta)$}

\SSec{mc_performance} verified the effectiveness of our blur-invariant motion estimation learning
by comparing \emph{BIMNet} with state-of-the-art optical flow estimation methods, by measuring PSNR and SSIM between $Warp(I_{t-1}^s, W^{bb})$ and $I_{t}^s$.
In this section, we provide an additional evaluation of the performance of \emph{BIMNet} on images with different types of blur.
Specifically, we evaluated the performance of \emph{BIMNet} for the other three cases: \emph{BS} $(Warp(I_{t-1}^s, W^{bs}),I_{t}^{s})$, \emph{SB} $(Warp(I_{t-1}^s, W^{sb}),I_{t}^{s})$, and \emph{SS} $(Warp(I_{t-1}^s, W^{ss}),I_{t}^{s})$.

Table~\ref{tbl:additional_mc_table} shows the evaluation results of \emph{BS}, \emph{SB}, and \emph{SS}.
In the cases of \emph{BS} and \emph{SB},
\emph{BIMNet$_{FN2}$} and \emph{BIMNet$_{LFN}$} achieved the best and second-best performances, respectively, in both PSNR and SSIM, clearly outperforming all the other methods.
In the case of \emph{SS}, while \emph{FlowNet2$_{SS}$}, which is trained for sharp images, performs the best in PSNR,
\emph{BIMNet$_{FN2}$} and \emph{BIMNet$_{LFN}$} perform comparably in PSNR and better than the other methods in SSIM, showing that both of them perform reasonably well even for sharp images.


\section{\emph{PVDNet}}

\subsection{Network architecture}

A detailed architecture of \emph{PVDNet} is given in \Tbl{network_architecture}.

\subsection{Preliminary tests for \emph{PVDNet}}
\label{ssec:preliminary_test}

As described in \Sec{overall}, our framework takes neighboring frames $I_{t-1}^b$ and $I_{t+1}^b$ {\em without} motion compensation.
We also construct a pixel volume of $I_{t-1}^{est}$ using its {\em grayscale} pixel values.
While this combination may seem less optimal, we found that it actually achieves the highest quality in our preliminary tests.
In this section, we summarize the results of our preliminary tests.

In the tests, we considered the following options for \emph{PVDNet}:
\begin{itemize}
\item warping {\em vs}.\ pixel volume for motion compensation,
\item grayscale pixel volume {\em vs}.\ color pixel volume,
\item applying motion compensation only to $I_{t-1}^{est}$ {\em vs}.\ to all images of $I_{t-1}^{est}$, $I_{t-1}^b$, and $I_{t+1}^b$, and
\item using only $I_{t-1}^{est}$ {\em vs}.\ all images of $I_{t-1}^{est}$, $I_{t-1}^b$, and $I_{t+1}^b$ for the neighboring frame information. Here, motion compensation is applied only to $I_{t-1}^{est}$.
\end{itemize}
To compare the different options, we prepared various variants of \emph{PVDNet}.

\begin{table}%
\centering
\caption{Detailed architecture of \emph{PVDNet}.}
\vspace{-8pt}
\begin{tabular}{cccccc}
  \hline
  \multicolumn{2}{c}{\emph{layer type}(\#)}&size & stride & out & act.\\
  \hline
  \hline
  \multicolumn{6}{c}{\textbf{Feature Transform Layers}}\\
  \hline
  \hline
  \multicolumn{2}{c}{\emph{Input}} & \multicolumn{4}{c}{$V_{t-1}$}\\
  \hline
  \multicolumn{2}{c}{\emph{Conv}} & $3\times3$  & (2, 2) & 64  & \emph{relu} \\
  \multicolumn{2}{c}{\emph{Conv}} & $3\times3$  & (2, 2) & 64  & \emph{relu} \\
  \multicolumn{2}{c}{\emph{add1}} & \multicolumn{4}{c}{\emph{Conv}, \emph{Input}}\\
  \hline
  \multicolumn{2}{c}{\emph{Concat}} & \multicolumn{4}{c}{$I_{t-1}^b$, $I_{t}^b$, $I_{t+1}^b$}\\
  \hline
  \multicolumn{2}{c}{\emph{Conv}} & $3\times3$  & (2, 2) & 32  & \emph{relu} \\
  \multicolumn{2}{c}{\emph{Conv}} & $3\times3$  & (2, 2) & 32  & \emph{relu} \\
  \multicolumn{2}{c}{\emph{add2}} & \multicolumn{4}{c}{\emph{Conv}, \emph{Concat}}\\
  \hline
  \multicolumn{2}{c}{\emph{Concat}} & \multicolumn{4}{c}{\emph{add1}, \emph{add2}}\\
  \hline
  \hline
  \multicolumn{6}{c}{\textbf{Encoder}}\\
  \hline
  \hline
  \multicolumn{2}{c}{\emph{Conv}1\_1} & $3\times3$  & (1, 1) & 64 & - \\
  \multicolumn{2}{c}{\emph{Conv}2\_1} & $3\times3$  & (2, 2) & 64  & \emph{relu} \\
  \hline
  \multicolumn{2}{c}{\emph{Conv}2\_2} & $3\times3$  & (1, 1) & 64  & - \\
  \multicolumn{2}{c}{\emph{Conv}3\_1} & $3\times3$ & (2, 2) & 128  & \emph{relu} \\
  \hline
  \multicolumn{2}{c}{\emph{Conv}3\_2} & $3\times3$ & (1, 1) & 128  & \emph{relu} \\

  \hline
  \hline
  \multicolumn{6}{c}{\textbf{ResBlocks $\times 12$}}\\
  \hline
  \hline
  \multicolumn{2}{c}{\emph{Conv}} & $3\times3$ & (1, 1) & 128  & \emph{relu} \\
  \multicolumn{2}{c}{\emph{Conv}} & $3\times3$ & (1, 1) & 128  & \emph{relu} \\

  \hline
  \hline
  \multicolumn{6}{c}{\textbf{Decoder}}\\
  \hline
  \hline
  \multicolumn{2}{c}{\emph{Conv}4\_1} & $3\times3$ & (1, 1) & 128  &  - \\
  \multicolumn{2}{c}{\emph{add}} & \multicolumn{3}{c}{\emph{Conv}4\_1, \emph{Conv}3\_2}  & - \\
  \hline
  \multicolumn{2}{c}{\emph{DeConv1}} & $4\times4$ & (2, 2) & 64  &  - \\
  \multicolumn{2}{c}{\emph{add}} & \multicolumn{3}{c}{\emph{DeConv}1, \emph{Conv}2\_2} &  \emph{relu} \\
  \multicolumn{2}{c}{\emph{Conv5\_2}} & $3\times3$  & (1, 1) & 64  & \emph{relu} \\
  \hline
  \multicolumn{2}{c}{\emph{DeConv2}} & $4\times4$ & (2, 2) & 64  &  - \\
  \multicolumn{2}{c}{\emph{add}} & \multicolumn{3}{c}{\emph{DeConv}2, \emph{Conv}1\_1} & \emph{relu} \\
  \multicolumn{2}{c}{\emph{Conv6\_2}} & $3\times3$  & (1, 1) & 64 & \emph{relu} \\
  \multicolumn{2}{c}{\emph{Conv6\_3}} & $3\times3$  & (1, 1) & 3 & - \\
  \hline
  \multicolumn{2}{c}{\emph{add}} & \multicolumn{3}{c}{\emph{Conv}6\_3, \emph{Input}} & - \\
  \hline

\end{tabular}
\vspace{-0.3cm}
\label{tbl:network_architecture}
\end{table}

\begin{table}[t]
\centering
\caption{Preliminary test results of various variants of \emph{PVDNet}.
The first column (`MC target') denotes the target frames among $I_{t-1}^{est}$, $I_{t-1}^b$ and $I_{t+1}^b$ for which motion compensation is applied.
The second column (`MC type') denotes the type of motion compensation applied for the frames in `MC target'.
The third column (`use $I_{t-1}^b$ and $I_{t+1}^b$') denotes whether \emph{PVDNet} takes $I_{t-1}^b$ and $I_{t+1}^b$ as input or not.
We computed the average errors between deblurred results $I_{t}^{est}$ and ground truths $I_{t}^{s}$ in PSNR and SSIM.}
\vspace{-8pt}
\begin{tabular}{ |c|c|c||c c| }
\hline
MC target & MC type & use $I_{t-1}^{b}$ and $I_{t+1}^{b}$ & PSNR (dB) & SSIM\\
\hline \hline
all &  & yes & 31.30 & 0.912\\
$I_{t-1}^{est}$ only & gray PV & yes & \bf{31.63} & \bf{0.915} \\
$I_{t-1}^{est}$ only &  & no & 31.10 & 0.912\\
\hline
all& & yes & 30.65 & 0.901 \\
$I_{t-1}^{est}$ only & color PV & yes & \bf{31.67} & \bf{0.916} \\
$I_{t-1}^{est}$ only & & no & 31.11 & 0.912\\
\hline
all & & yes & 31.11 & 0.906\\
$I_{t-1}^{est}$ only & warp & yes & 31.42 & 0.909\\
$I_{t-1}^{est}$ only & & no & 30.89 & 0.904 \\
\hline
\end{tabular}
\vspace{-0.3cm}
\label{tbl:pre_table}
\end{table}

\Tbl{pre_table} shows the variants of \emph{PVDNet} and their performances in PSNR and SSIM.
In the table, `all' in the first column (`MC target') means that all three images, $I_{t-1}^{est}$, $I_{t-1}^b$, and $I_{t+1}^b$, are motion-compensated.
On the other hand, `$I_{t-1}^{est}$ only' means that only $I_{t-1}^{est}$ is motion-compensated while $I_{t-1}^b$ and $I_{t+1}^b$ are used as they are.
In the second column (`MC type'), `gray PV' means that a pixel volume is constructed using grayscale pixel values, while `color PV' means that it is constructed using RGB values.
`warp' means that simple warping is used instead of a pixel volume for motion compensation.
In the third column (`use $I_{t-1}^b$ and $I_{t+1}^b$'), `no' means that a \emph{PVDNet} does not take $I_{t-1}^b$ and $I_{t+1}^b$ as input, and uses only $I_{t-1}^{est}$.
Our final model corresponds to \{`$I_{t-1}^{est}$', `gray PV', `yes'\} in the table.

From \Tbl{pre_table}, we can make the following observations:
1) using a pixel volume consistently outperforms using warping, showing the effectiveness of a pixel volume,
2) using a grayscale pixel volume achieves comparable performance to using a color pixel volume, while it requires much less memory and computation,
3) interestingly, applying motion compensation only to $I_{t-1}^{est}$ leads to higher quality results than applying motion compensation to all images, and
4) using the neighboring input blurry frames $I_{t-1}^b$ and $I_{t+1}^b$ helps increase the performance, compared to using $I_{t-1}^{est}$ only.
Based on these observations, we chose \{`$I_{t-1}^{est}$', `gray PV', `yes'\} as our final model.

It may seem counter-intuitive that applying motion compensation to all images decreases the deblurring performance.
A possible reason is that the neighboring input blurry frames $I_{t-1}^b$ and $I_{t+1}^b$ would help deblurring only in limited cases.
Specifically, if there is a large motion between $I_{t-1}^b$ and $I_{t}^b$, it is likely that $I_{t-1}^b$ is blurry, and motion-compensating the blurry neighboring frame would not help deblurring much.
In other words, $I_{t-1}^b$ helps deblurring only when there is slight motion, and in that case, keeping the original relatively sharp pixels could be more beneficial than distorting the image for small motion compensation.
The same arguments hold for $I_{t+1}^b$.
Our \emph{PVDNet} is trained to learn to selectively use the information in motion-compensated $I_{t-1}^{est}$ and input neighboring frames, $I_{t-1}^b$ and $I_{t+1}^b$, and consequently leads to higher performance.

\subsection{Non-recurrent \emph{PVDNet}}

While our PVDNet adopts a recurrent framework, non-recurrent approaches like DVD~\cite{su2017deep} have also been widely adopted so far.
Thus, we study the performance of non-recurrent variants of our framework and the effect of our motion compensation on them.
In this study, we prepared four different non-recurrent versions of our framework: baseline$_{nr}$, baseline$_{nr}$ + MC, baseline$_{nr}$ + BIMC, and baseline$_{nr}$ + BIMC + PV, similarly to the experiment in \Tbl{ablation_table}.
For baseline$_{nr}$, we modified the architecture of \emph{PVDNet} so that it receives three consecutive input blurry frames without a previous deblurred frame.
For all the variants, motion compensation is applied to the previous and next input blurry frames instead of a deblurred previous frame.
Specifically, baseline$_{nr}$ takes input blurry frames without any motion compensation, while baseline$_{nr}$ + MC and baseline$_{nr}$ + BIMC take warped input blurry frames whose motions are estimated by \emph{LiteFlowNet$_{SS}$} and \emph{BIMNet}, respectively.
For baseline$_{nr}$ + BIMC + PV, we modified the network architecture to take two pixel volumes of the previous and next input blurry frames, and the current target frame.
Specifically, the network has two input branches similarly to the network in \Fig{framework}.
One branch takes the current target blurry frame $I_t^b$ instead of three blurry frames.
On the other hand, the other branch takes two pixel volumes of $I_{t-1}^b$ and $I_{t+1}^b$ and concatenates them before the feature transform layer.

\Tbl{dvd_like} shows the performance evaluation result of the non-recurrent models.
Both baseline models of the non-recurrent and recurrent networks show similar accuracy (30.65 dB and 30.66 dB) as shown in \Tbl{dvd_like} and \Tbl{ablation_table}.
Although BIMC and a pixel volume improve the deblurring quality the most, all kinds of motion compensation in \Tbl{dvd_like} do not lead to significant improvement of the deblurring quality.
Su \Etal~\shortcite{su2017deep} also reported a similar result where motion compensation does not improve deblurring quality much.
We guess that this is because of the higher training complexity of non-recurrent deblurring.
In non-recurrent deblurring, there would be no sharp information that can help deblur the current target frame, in contrast to recurrent deblurring that uses a deblurred previous frame.
This can induce the network trained to less utilize information from neighboring frames.

\begin{table}[t]
\centering
\caption{Effectiveness of blur invariant motion estimation and pixel volumes with the non-recurrent \emph{PVDNet}.}
\vspace{-8pt}
\begin{tabular}{ |c || c | c | }
\hline

Model & PSNR (dB) & SSIM  \\
\hline \hline
baseline$_{nr}$ & 30.65 & 0.889 \\
baseline$_{nr}$ + MC & 30.79 & 0.896  \\
baseline$_{nr}$ + BIMC & 30.84 & 0.899\\
baseline$_{nr}$ + BIMC + PV & 30.87 & 0.900\\
\hline

\hline
\end{tabular}
  \vspace{-0.3cm}
\label{tbl:dvd_like}
\end{table}

\begin{table}[t]
\centering
\caption{Deblurring performance of our \emph{PVDNet} according to different motion compensation approaches.}
\vspace{-8pt}
\begin{tabular}{ | c || c | }
\hline
Method & PSNR (dB) \\
\hline \hline
Simple warping & 31.46 \\
Majority-based warping & 31.47 \\
Naïve PV & 31.48 \\
Our PV & 31.63 \\
\hline
\end{tabular}
\vspace{-0.3cm}
\label{tbl:PV_analysis}
\end{table}

\begin{figure*}[tp]
\centering
\setlength\tabcolsep{1 pt}
  \begin{tabular}{cccccc}
    \rotatebox[origin=l]{90}{  \,    \,    \,  \,  Input}  &
    \multicolumn{1}{c}{\includegraphics[width=0.182\linewidth]{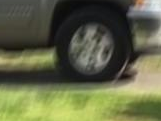}} &
    \multicolumn{1}{c}{\includegraphics[width=0.182\linewidth]{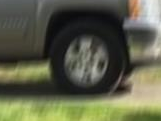}} &
    \multicolumn{1}{c}{\includegraphics[width=0.182\linewidth]{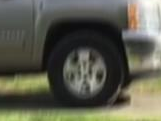}} &
    \multicolumn{1}{c}{\includegraphics[width=0.182\linewidth]{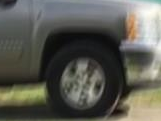}} &
    \multicolumn{1}{c}{\includegraphics[width=0.182\linewidth]{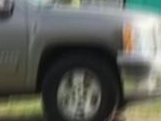}} \\

    \rotatebox[origin=l]{90}{  \,    \,    \,  \, \, Ours} &
    \multicolumn{1}{c}{\includegraphics[width=0.182\linewidth]{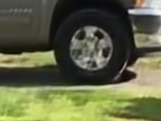}} &
    \multicolumn{1}{c}{\includegraphics[width=0.182\linewidth]{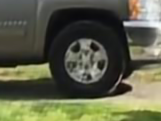}} &
    \multicolumn{1}{c}{\includegraphics[width=0.182\linewidth]{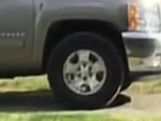}} &
    \multicolumn{1}{c}{\includegraphics[width=0.182\linewidth]{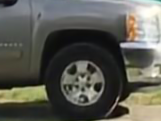}} &
    \multicolumn{1}{c}{\includegraphics[width=0.182\linewidth]{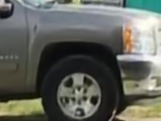}} \\

    & frame $\# 1$ & frame $\# 2$ & frame $\# 3$ & frame $\# 4$ & frame $\# 5$

  \end{tabular}
  \vspace{-0.4cm}
  \caption{Deblurring examples of a video sequence from Su \Etal~\shortcite{su2017deep}'s test set. Frame $\# 1$ is the first frame of the input video.}
  \vspace{-0.2cm}
\label{fig:deblurred_previous}
\end{figure*}

\subsection{Additional comparisons of pixel volume with other motion compensation approaches}
\label{ssec:pv_analysis}

In this section, we compare our pixel volume approach to two additional motion compensation approaches to analyze the effect of a pixel volume.
As described in \SSec{PV}, our pixel volume approach leads to the performance improvement in video deblurring by utilizing multiple candidates with majority cues in a pixel volume.
One natural question would be whether simply using the majority values in the pixel volume or multiple candidates without majority cues also leads the performance improvement.
Therefore, we experimented with two other motion compensation approaches. 

First, instead of our PV, we simply fed the deblurring result of the previous sharp frame warped by the majority-based warping to \emph{PVDNet}.
In majority-based warping, we built a PV based on the estimated flow, then chose the most frequent pixel value (the mode) among the $k^2$ candidates for each pixel.
Second, instead of constructing our PV, we constructed a na\"ive PV in a way that it can provide multiple candidates without spatial coherency, corresponding to simply increase the number of candidates.
Specifically, for each spatial position in a PV, we do not exploit the flow vectors of the neighbor pixels, but we simply collect pixels in the $k \times k$ window centered at the estimated position by \emph{BIMNet}.
This approach is similar to the neighbor search of \cite{cho2012video} and the cost volume construction of conventional stereo matching algorithms.
For the experiment and comparison, we conducted the same procedure used in \SSec{ablation}.

\Tbl{PV_analysis} summarizes the performance of different approaches.
The performance improvement of majority-based warping over simple warping is marginal, as the warped images from simple warping and majority-based warping would be similar to each other, as shown in \Fig{pv_work}c. 
The performance improvement of the na\"ive PV over simple warping is also marginal. Although the na\"ive PV contains multiple candidates, those multiple candidates do not have majority values due to the way of construction.
In contrast, the multiple candidates in our PV are obtained by exploiting flow information of neighboring pixels and provide the majority cue. 

\begin{table}[t]
\centering
\caption{Performances with different patch sizes for a pixel volume.}
\vspace{-8pt}
\begin{tabular}{ |c || c | c | c | c | c | }
\hline

Patch size & 1 & 3 & 5 & 7 & 9 \\
\hline \hline
PSNR (dB) & 31.42 & 31.54 & 31.63 & 31.55 & 31.11\\
SSIM & 0.912 & 0.912 & 0.915 & 0.913 & 0.905\\

\hline
\end{tabular}
\vspace{-0.2cm}
\label{tbl:patch_size}
\end{table}

\subsection{Patch size of pixel volume}
In this experiment, we analyze the deblurring performance change by a different patch size of a pixel volume.
We tested five patch sizes (1, 3, 5, 7, 9) and
found that $5\times 5$ performs the best (\Tbl{patch_size}).
Ideally, it can be expected that a larger patch size can help the network to handle more spacious flow errors, but a large patch size rather decreases the performance in practice.
This may be because a pixel volume generated by large patches contains too many candidates to consider, which exceeds the capability of \emph{PVDNet}.

\subsection{Pixel volume vs flow aggregation using conv layers}

Our framework explicitly constructs a pixel volume for motion compensation.
One natural question would be whether it would be possible to learn to aggregate flow information of neighboring pixels using additional convolution layers instead of explicit construction of a pixel volume.
To verify this, we may consider the following two options:
\begin{enumerate}
    \item Additional convolution layers at the end of BIMNet to refine the flow result
    \item Additional convolution layers at the beginning of PVDNet to aggregate flow results
\end{enumerate}
Specifically, the first option refines the flow result through additional convolution layers. This option is equivalent to adopt a larger network architecture for BIMNet. To verify whether this option brings a significant improvement, we trained a larger network with additional five convolution layers for BIMNet. We then measured the deblurring performance, but found no noticeable improvement.

The second option is not trivial. To deblur the current frame, video deblurring requires pixel value information of the previous frame, not its flow information \cite{cho2012video,su2017deep,kim2017online,kim2018spatio,Wang2019edvr}. Thus, even if we somehow aggregate flow information using additional convolution layers, we still need to warp the previous deblurred frame. Moreover, aggregating and refining flow information is not trivial as well. Instead, our pixel-volume approach provides an effective way that allows the network to aggregate pixel value information from multiple candidate pixels. In addition, a pixel volume takes smaller memory than a single convolution layer (\SSec{cost}). 

\begin{figure}[tp]
\centering
	\begin{tabular}{c}
		\includegraphics[width=0.9\columnwidth]{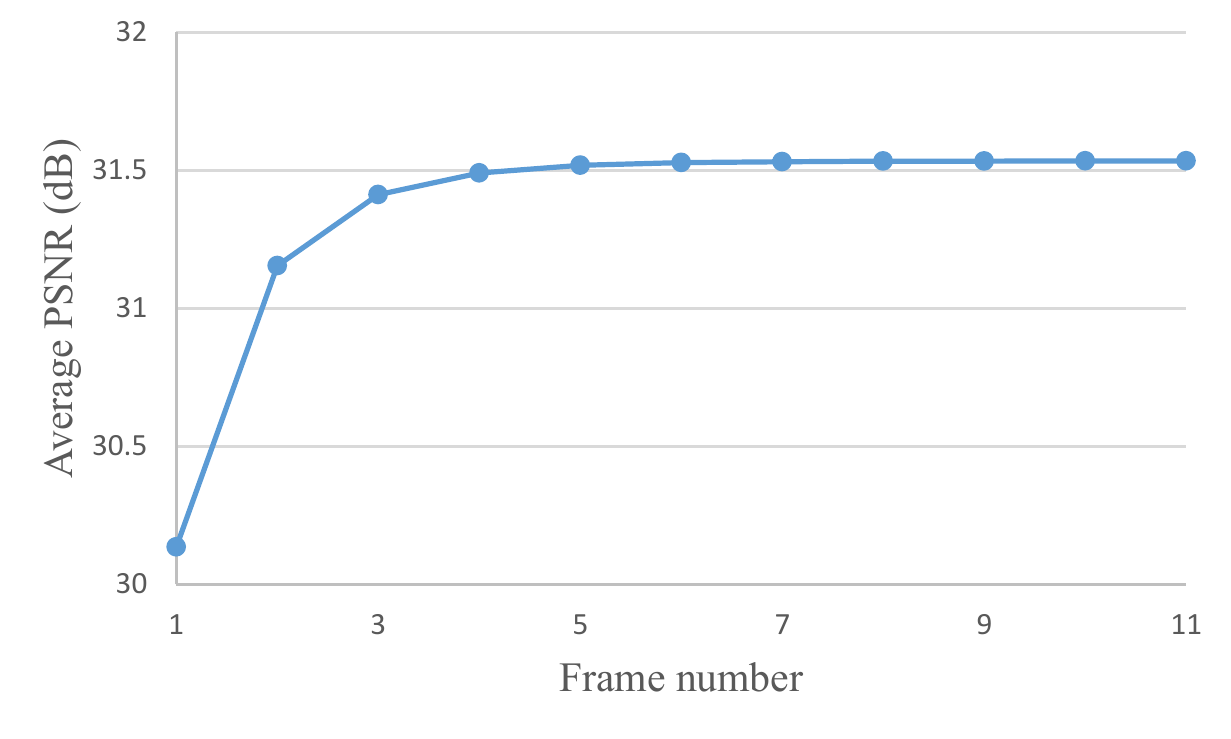}
	\end{tabular}
	\vspace{-0.6cm}
	\caption{Performance graph of our video deblurring framework on Su \Etal's dataset~\cite{su2017deep} 
	at different time steps.
	The average PSNR is low at the first frame, but it quickly raises and gets stabilized as our method proceeds.}
	\vspace{-0.4cm}
\label{fig:number_of_frames}
\end{figure}
 
\subsection{Deblurring of first few frames}

\Fig{deblurred_previous} shows the deblurring results of the first few frames of a video. 
The top row shows input consecutive blurry frames, while the bottom row shows their corresponding results.
When our method starts to deblur the video from the first frame, the first frame has no deblurred previous frame, and its deblurring result still has a small amount of remaining blur, e.g., the details of the wheel are slightly blurry.
Nevertheless, from the second frame, our method can utilize the deblurred result of the previous frame, so it produces a clearer result with sharper details. 


\Fig{number_of_frames} visualizes the average PSNRs of the deblurring results at different time steps.
The average PSNRs are measured from the test set of Su \Etal's dataset~\cite{su2017deep}.
The average PSNR of the first frames is low as they have no deblurring results of the previous frames.
However, as the method proceeds, the average PSNR quickly raises and gets stabilized.
This proves the practicality and effectiveness of our approach despite its recurrent structure that requires deblurring results of previous frames.

\bibliographystyle{ACM-Reference-Format}
\bibliography{egbib}

\end{document}